\documentclass{article}

\usepackage{arxiv}

\usepackage[utf8]{inputenc} 
\usepackage[T1]{fontenc}    
\usepackage{hyperref}       
\usepackage{url}            
\usepackage{booktabs}       
\usepackage{amsfonts}       
\usepackage{nicefrac}       
\usepackage{microtype}      
\usepackage{lipsum}		
\usepackage{graphicx}
\usepackage{doi}

\usepackage{amsmath}
\usepackage{amsthm}

\usepackage{overpic}
\usepackage{color}

\title{First qualitative observations on deep learning vision model YOLO and DETR for automated driving in Austria}

\author{ \href{https://orcid.org/0000-0002-2148-6703}{\includegraphics[scale=0.06]{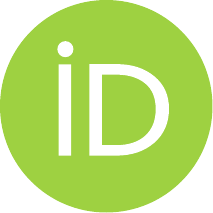}\hspace{1mm}Stefan~Schoder}\\
	TU Graz\\
	8010 Graz, Austria \\
	\texttt{stefan.schoder@tugraz.at} \\
}

\usepackage{caption}
\usepackage{subcaption}

\begin{document}
\maketitle

\begin{abstract}
This study investigates the application of single and two-stage 2D-object detection algorithms like You Only Look Once (YOLO), Real-Time DEtection TRansformer (RT-DETR) algorithm for automated object detection to enhance road safety for autonomous driving on Austrian roads. The YOLO algorithm is a state-of-the-art real-time object detection system known for its efficiency and accuracy. In the context of driving, its potential to rapidly identify and track objects is crucial for advanced driver assistance systems (ADAS) and autonomous vehicles. The research focuses on the unique challenges posed by the road conditions and traffic scenarios in Austria. The country's diverse landscape, varying weather conditions, and specific traffic regulations necessitate a tailored approach for reliable object detection. The study utilizes a selective dataset comprising images and videos captured on Austrian roads, encompassing urban, rural, and alpine environments.
\end{abstract}


\section{Introduction}

In autonomous driving or driving safety systems, a rapid 2D-object detection algorithm is key for various assisting technologies like the advanced driver assistance systems (ADAS). For this purpose, several fast 2D-object detection libraries based on pre-trained models are available, like the single object detection algorithms like You Only Look Once (YOLO) and the more advanced object detection algorithm Real-Time DEtection TRansformer (RT-DETR) and are selected for the study in hand based on performance on various datasets \cite{lv2023detrs,yolov8ultralytics,redmon2018yolov3,zhao2023fast}. The YOLO family algorithms exists in different variants and sizes. Within this qualitative comparative study the YOLO version 2 (YOLOv2) \cite{redmon2017yolo9000}, version 3 (YOLOv3) \cite{redmon2018yolov3}, version 5 (nano, small, medium, large and very large flavor) (YOLOv5-n/s/m/l/x) \cite{Jocher_YOLOv5_by_Ultralytics_2020} and version 8 (nano, small, medium, large and very large flavor) (YOLOv8-n/s/m/l/x) \cite{yolov8ultralytics} are compared to each other and to the transformer based RT-DETR (small and large flavor) (RT-DETR-s/l) algorithm \cite{lv2023detrs} in the context of road traffic. In a first evaluation, the performance (of the pre-trained models) on well-known datasets and computer architectures is reported as literature review for the selected models. The aim of the literature study is to detect potential weak points concerning driving systems, improvements of the models historically to overcome detected weaknesses. As a result, a potential list of challenges is identified and used for further evaluation on the traffic datasets obtained. In a first qualitative study, characteristics of the investigated models (YOLOv2, YOLOv3, YOLOv5-n/s/m/l/x, YOLOv8-n/s/m/l/x, RT-DETR-s/l) are discussed on a freely available road traffic dataset from drive.ai\footnote{Drive.ai Sample Dataset, 100 frames from drive.ai, Creative Commons Attribution 4.0 International License, accessed on 29/11/2023 and redistributed by \href{https://github.com/jungsoh/yolo-car-detection-for-autonomous-driving/tree/main/images}{the GitHub repository}} recorded (100 frames) in the United States. In these scenes, the characteristics of the individual frames are compared and reported defined on detected obstacles, (below threshold risky) non-detected obstacles, (risky artifacts) wrong-detected obstacles, (risky) wrong-classified obstacles. With this first evaluation and from a cautious driver perspective, risky is defined as an object being near the car and may cause a legal hazard (car crash, harming other people, car damage, ...) if the car drove on in the scene. For instance, a risky artifact would be wrongly detected fire hydrant (indicating no road) in the middle of the road by falsely classifying a traffic cone. As a second step, urban, rural, alpine road and alpine tunnel entry with construction of Austrian traffic scenes (a snapshot and video for each scene) are evaluated. Specific characteristics of the road features are investigated and qualitatively compared to the US dataset. The qualitative Austrian road dataset consists of specific characteristics and known algorithm breakdowns are tried to build into the videos (like rotation of the video).



The outcomes of this study aim to provide first valuable insights into the effectiveness of real-time deep learning vision algorithm for automated 2D-object detection on Austrian roads. The findings are a starting point for contributing to the development of more robust and region-specific ADAS and autonomous driving systems, ultimately promoting road safety in Austria and serving as a foundation for similar studies in a diverse geographical context.








\section{Theory on object detection}
For autonomous driving control and safety, various sensor data is used to determine a possible action space for future movements. Within this first study, the focus is lying on 2D image processing from simple cameras. During the survey of possible algorithm tasks, image classification, 2D object detection and image segmentation (e.g. segmenting anything \cite{zhao2023fast}) are important tasks to be considered. Where object classification is usually one part of the task of object detection or post-processing the detected object into a class. The state of the art of image classification is now shifting from convolutional neural network (CNN) architecture to transformer \cite{dosovitskiy2020image,lv2023detrs} also suitable for multimodal tasks \cite{srivastava2023omnivec}. CNN processes pixel data and is a basic building block for the computer vision task of fast object detection algorithms and image segmentation \cite{yolov8ultralytics}. 
The CNN architecture has typically three main layers, a convolutional layer (filters the input image as a feature map using kernels), a pooling layer (downsamples feature maps by summarizing the features in patches) and a fully connected layer. A simple CNN structure can perform image classification (a single object in a single image) into proposed classes (supervised training based on labeled image data for all classes). From this early classification task, object detection was the next step by a process named region-based CNN (R-CNN) \cite{girshick2015fast}. A R-CNN uses an input image and then distributes possible bounding boxes across the image (object regions, typically named Regions of Interest (ROI)). Each ROI mapped to a standard size and is further classified by a CNN  architecture into a proposed class. So objects in a picture can be detected. Building upon the R-CNN idea, two-stage object detection algorithm like faster R-CNN emerged (with a nice explaination found here\footnote{https://towardsdatascience.com/understanding-and-implementing-faster-r-cnn-a-step-by-step-guide-11acfff216b0}). Illustratively, the two stages are explained here and to enhance the understanding of the YOLO algorithm evolution later in this manuscript. The first stage consists of a region proposal network (RPN) (limiting the number of possible bounding boxes in the image) and a second stage classifying the ROI based on the RPN. During the training (e.g. ground truth labeled with CVAT \cite{CVAT_ai_Corporation_Computer_Vision_Annotation_2023}),
the RPN learns what are important bounding box proposals to look at in an image and the second step aims to classify and predict the offsets of the bounding boxes. Stage 1, the image is processed through the RPN backbone (a CNN e.g. ResNet, VGG16, ...) learning the features in the image being located at a feature point (anchor). At each anchor (1x1 convolutional network), multiple (different sized) bounding boxes are seeded. From the ground truth, anchor boxes, mostly overlapping with projected ground truth boxes, are assigned the category object. Other boxes are assigned to background. Binary cross-entropy loss tailors the network learns to classify the seeded anchor boxes. Predicted (object) anchor boxes may not exactly align with the projected ground truth boxes (true object). Similarly, a 1x1 convolutional network learns the predicted offsets from ground truth boxes (true object). Finally, at stage one, the predicted (object) anchor boxes are corrected by the predicted offset and are the input region proposals (RP) for the second stage of the faster R-CNN algorithm.
In stage 2, the object is classified in the resized RP by a CNN architecture into a proposed class. A second regression network aims to determine the RP offsets from ground truth boxes. The total loss is a linear combination of the individual task losses. During inference, a crucial step is that only the anchor boxes with the best qualification score are processed further to stage 2 for the classification into proposed classes. After stage 2, in a post-processing step - duplicates are removed by a method called non-max suppression (removing bounding boxes with large overlapping regions). Built upon faster R-CNN, the mask R-CNN \cite{he2017mask} method performs can perform both an instance segmentation (a single object from a class from the background) and a semantic segmentation (classes from the background). The mask R-CNN is the pixel-to-pixel alignment to generate at the second stage a binary mask for each RP (segmenting the object from the background). For instance, this is valuable for detecting roads and admissible movements in autonomous driving. 

\subsection{YOLO}

In this section, the basics of You Only Look Once (YOLO) model are described being a single level object detection algorithm with the aim of real-time processing of images and video streams. The speed of Yolo makes it an interesting candidate for algorithms to look at, the Yolo base network runs at 45 frames per second with no batch processing on a Titan X GPU and a fast version runs at more than 150 fps \cite{redmon2016you}.

\begin{figure}[ht!]
    \centering
    \includegraphics[width=0.9\textwidth]{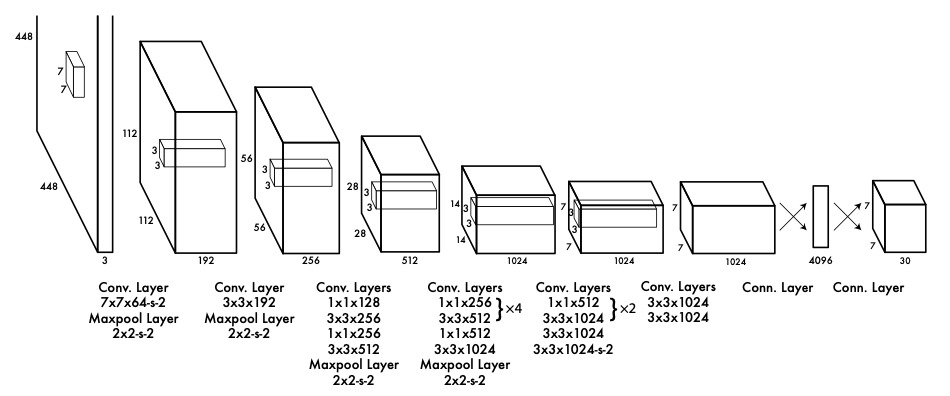}
    \caption{The first generation YOLO architecture \cite{redmon2016you}.}
    \label{fig:networkYOLO}
\end{figure}

YOLO divides the image into a grid $S\times S$ and predicts bounding boxes and class probabilities directly, allowing simultaneous detection across the entire image (see original publication \cite{redmon2016you}). First, the image is divided into a grid, consisting of grid cells. Whenever a center of an object is located in a grid cell this grid cell is responsible for detecting this object. For each grid cell, a number of $B$ bounding boxes (with box location $(x,y)$, height $h$ and width $w$) and $P(Obj\mid B) IuO^\mathrm{truth}_\mathrm{pred}$ confidence scores are predicted (prediction tuple of 5 entries $\{x,y,h,w,P(Obj\mid B) IuO^\mathrm{truth}_\mathrm{pred}\}$. The confidence score describes the confidence (probability) $P(Obj\mid B)$ of a bounding box $B$ containing an object $Obj$ and reflects how accurate the bounding box is predicted by intersection over union $IuO^\mathrm{truth}_\mathrm{pred}$ between the predicted box and the ground truth. For each grid cell, the proposed classes $C_i$ with $i = 1,\dots, C$ are predicted based on conditional class probabilities $P(C_i\mid Obj)$. $P(C_i\mid Obj) P(Obj\mid B) IuO^\mathrm{truth}_\mathrm{pred}$ encode both the probability of that class appearing in the box and how well the predicted box fits the object \cite{redmon2016you}. In total, for each image predictions are encoded as an $S \times S \times (5B + C)$ tensor. E.g. with $S = 7$, $B = 2$ and $C = 20$ the following network architecture was used (see Fig.~\ref{fig:networkYOLO}). Further details on the architecture can be found in \cite{redmon2016you}. The following loss function $\mathcal{L}$ is used during training 
\begin{equation}
    \begin{gathered}
\mathcal{L} = \lambda_{\text {coord }} \sum_{i=0}^{S\times S} \sum_{j=0}^B \mathbf{1}_{i j}^{\text {obj }}\left[\left(x_i-\hat{x}_i\right)^2+\left(y_i-\hat{y}_i\right)^2\right] \\
+\lambda_{\text {coord }} \sum_{i=0}^{S\times S} \sum_{j=0}^B \mathbf{1}_{i j}^{\text {obj }}\left[\left(\sqrt{w_i}-\sqrt{\hat{w}_i}\right)^2+\left(\sqrt{h_i}-\sqrt{\hat{h}_i}\right)^2\right] \\
+\sum_{i=0}^{S\times S} \sum_{j=0}^B \mathbf{1}_{i j}^{\text {obj }}\left(C_i-\hat{C}_i\right)^2 \\
+\lambda_{\text {noobj }} \sum_{i=0}^{S\times S} \sum_{j=0}^B \mathbf{1}_{i j}^{\text {noobj }}\left(C_i-\hat{C}_i\right)^2 \\
+\sum_{i=0}^{S\times S} \mathbf{1}_i^{\text {obj }} \sum_{c \in \text { classes }}\left(p_i(c)-\hat{p}_i(c)\right)^2
\end{gathered}
\end{equation}
where $\mathbf{1}_i^{\mathrm{obj}}$ denotes if an object appears in a cell $i$ and $\mathbf{1}_{i j}^{\mathrm{obj}}$ denotes that the $j$-th bounding box predictor in the cell $i$ takes responsibility for that prediction \cite{redmon2016you}. 
The first two terms are localization loss of the predicted (hat) bounding boxes compared to the true ones. The second two terms are the confidence losses that there is or there is not an object in the cell. The last term is the bounding box object classification loss for the class prediction.
During inference, the network is evaluated once predicting bounding boxes, for the objects and object classes. In cases of large objects or near objects multiple cells may predict the object and bounding boxes, non-maximal suppression is used to fix multiple detections. The limitation of the YOLO base model can be summarized by the following statements \cite{redmon2016you}: \begin{itemize}
    \item For each grid cell, two bounding box are predicted that can only have one class. This limits the near object recognition abilities.
    \item It struggles with small objects and small objects that appear in groups, like a swarm of birds.
    \item Generalization of objects in unusual aspect ratios or configurations is weak.
    \item The model relies on coarse features to predicting bounding boxes (multiple downsampling layers from the input image, see Fig.~\ref{fig:networkYOLO}).
    \item During training, the loss function of small bounding boxes versus large bounding boxes are treated by an absolute measure of distance. Leading to relatively large possible errors for small bounding boxes and incorrect localizations.
\end{itemize}
Comparisons on the performance on various datasets and against other architectures can be found in \cite{redmon2016you}. The YOLO architecture was improved in the last few years, with version 2 (YOLOv2 \cite{redmon2017yolo9000}) batch normalization was added to the convolutional layers and dropout was removed (adding an improvement of about mAP of about 2\%), multi-resolution classification was fine-tuned on 448x448 image resolutions ImageNet before tuning it to detection (adding an improvement of about mAP of about 4\%). YOLO was using arbitrary boundary boxes, with v2 bounding boxes based on anchor box types were proposed with defined offsets to these anchor boxes to maintain the generic capabilities (improvement of the recall). Typically, objects like a standing person or a car have a defined box ratio. In contrast to YOLO, YOLOv2 now predicts for each bounding box the classes and not for each cell\footnote{Some grafical explaination can be found  \href{https://jonathan-hui.medium.com/real-time-object-detection-with-yolo-yolov2-28b1b93e2088}{here.}}. By k-means anchor box dimension clusters, the data-driven selection (compared to hand selection) of anchor boxes was achieved based on $IoU$. Furthermore, the offsets of the anchor boxes were constrained with distant from the cell centroid. Furthermore, capabilities for fine-grained features and multi-scale training added to YOLOv2 performance. A detailed discussion can be found in \cite{redmon2017yolo9000}. 
In YOLOv3 \cite{redmon2018yolov3}, multi-label classification was used, since some classes are not mutual exclusive (person, pedestrian, child, ...). In doing so, the soft max operation is avoided and the classification loss is now based on binary cross-entropy. It makes 3 predictions per location at different resolution levels. One prediction is carried out at the last feature map layer, one that upsamples features from two layers back by two. And a third, by going back another two and upsample it again by two. YOLOv3 gained significant capabilities of detecting small objects \cite{redmon2018yolov3}. Additional improvments on the usability and functionality were added with version 5 \cite{Jocher_YOLOv5_by_Ultralytics_2020} 
(integrates the anchor-free and objectness-free split head) and version 8 \cite{yolov8ultralytics}. Version 8 can also be used for instance segmentation, skeleton prediction of a human pose and classification. To conclude, YOLO's real-time capabilities and easy to handle model architecture are crucial for rapid object detection in autonomous driving scenarios. 

\subsection{RT-DETR}
\begin{figure}[ht!]
    \centering
    \includegraphics[width=0.9\textwidth]{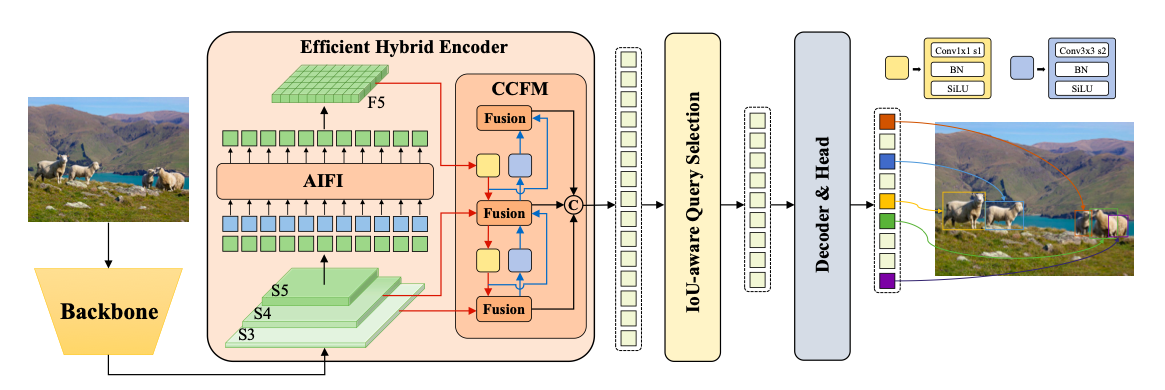}
    \caption{\textit{"Overview of RT-DETR. We first leverage features of the last three stages of the backbone $\{S3 , S4 , S5 \}$ as the input to the encoder. The efficient hybrid encoder transforms multi-scale features into a sequence of image features through intra-scale feature interaction (AIFI) and cross-scale feature-fusion module (CCFM). The $IoU$-aware query selection is employed to select a fixed number of image features to serve as initial object queries for the decoder. Finally, the decoder with auxiliary prediction heads iteratively optimizes object queries to generate boxes and confidence scores."} \cite{lv2023detrs}.}
    \label{fig:rtdetrArch}
\end{figure}
DETRs have achieved remarkable performance in object detection tasks. Initially, the high computational cost limits their practical usage. Especially, the post-processing with non-maximum suppression is beneficial with the computational cost, preventing original DETRs from being a new state-of-the-art (SOTA) for real-time object detection. The RT-DETR was developed to solve the problem of high computational cost, above-mentioned \cite{lv2023detrs}. In \cite{lv2023detrs},  it was shown how the $IoU$-threshold for admissible bounding boxes varies remaining prediction bounding boxes for YOLOv5 and YOLOv8. Based on the number of remaining prediction bounding boxes, the non-maximum suppression takes a significant execution time (depending on the $IoU$-threshold hyperparameters) and motivates the use of DETRs, with an overview of the architecture in Fig.~\ref{fig:rtdetrArch}. Firstly, the big picture of RT-DETR\footnote{Implementation of RT-DETR is found on \href{https://github.com/lyuwenyu/RT-DETR}{github.com/lyuwenyu/RT-DETR}} is discussed. As described in \cite{lv2023detrs}, RT-DETR consists of a backbone, a hybrid encoder and a transformer decoder with auxiliary prediction heads. The last three stages of the backbone $\{S3 , S4 , S5 \}$ are fed as input into the encoder. The efficient hybrid encoder processes multiscale features by a process decoupling intra-scale feature interaction (AIFI) and cross-scale feature-fusion module (CCFM). The details of the hybrid encoder (removing redundant operations of existing encoders) can be found in \cite{lv2023detrs}. After the encoder, the results are processed $IoU$-aware query selection. This is important to have the focus on the most relevant objects in the scene by avoiding non-relevant parts and therefore enhancing the detection accuracy. The IoU-aware query selection constraints the model to produce high classification scores for features with high $IoU$ scores and low classification scores for features with low $IoU$ scores during training \cite{lv2023detrs}. 
\begin{figure}[ht!]
    \centering
    \includegraphics[width=0.5\textwidth]{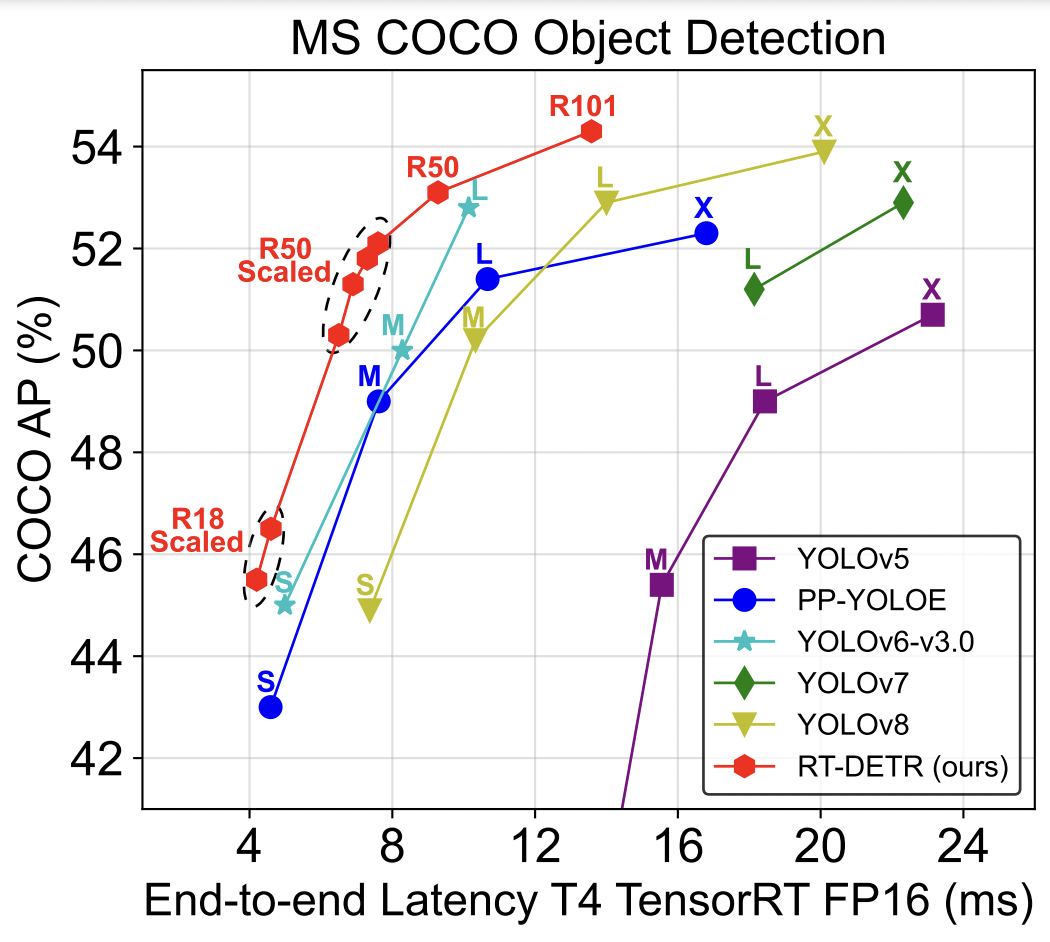}
    \caption{Compared to other real-time object detectors, RT-DETR achieves state-of-the-art performance in both speed and accuracy \cite{lv2023detrs}.}
    \label{fig:rtdetr}
\end{figure}
Finally, the decoder predicts outputs to generate boxes and confidence scores. This design reduces computational costs and allows for real-time object detection on accelerated backends, outperforming other real-time object detectors (see Fig.~\ref{fig:rtdetr}.


\section{Testing - USA-CA}
A first qualitative comparison for all models (YOLOv2, YOLOv3, YOLOv5, YOLOv8, RT-DETR) will be performed with the pre-trained models and applied to a dataset provided during the Coursera specialization "Deep Learning" by DeepLearning.ai and hosted by drive.ai. During this study, a visual interpretation of the real-time object detection algorithms is executed for different scenes and common characteristics are described. This will be a starting point for detecting common characteristics of the model and possible pitfalls when applying it to road traffic scenes. Special attention will be put on close-up objects and traffic signs that are not detected by the algorithm but are of potential interest for road safety.

\subsection{Dataset}
In the dataset, 120 images collected by a camera mounted to the hood, simply speaking in front, of a car. While driving, it takes pictures of the road ahead every few seconds. The scenes analyzed exemplarily below shows images taken from a car-mounted camera while driving around Silicon Valley. Only the most significant representations of the dataset provided by drive.ai are reported visually. 

\subsection{Results}
\begin{figure}[ht!] 
     \centering
     \begin{overpic}[width=0.49\textwidth]{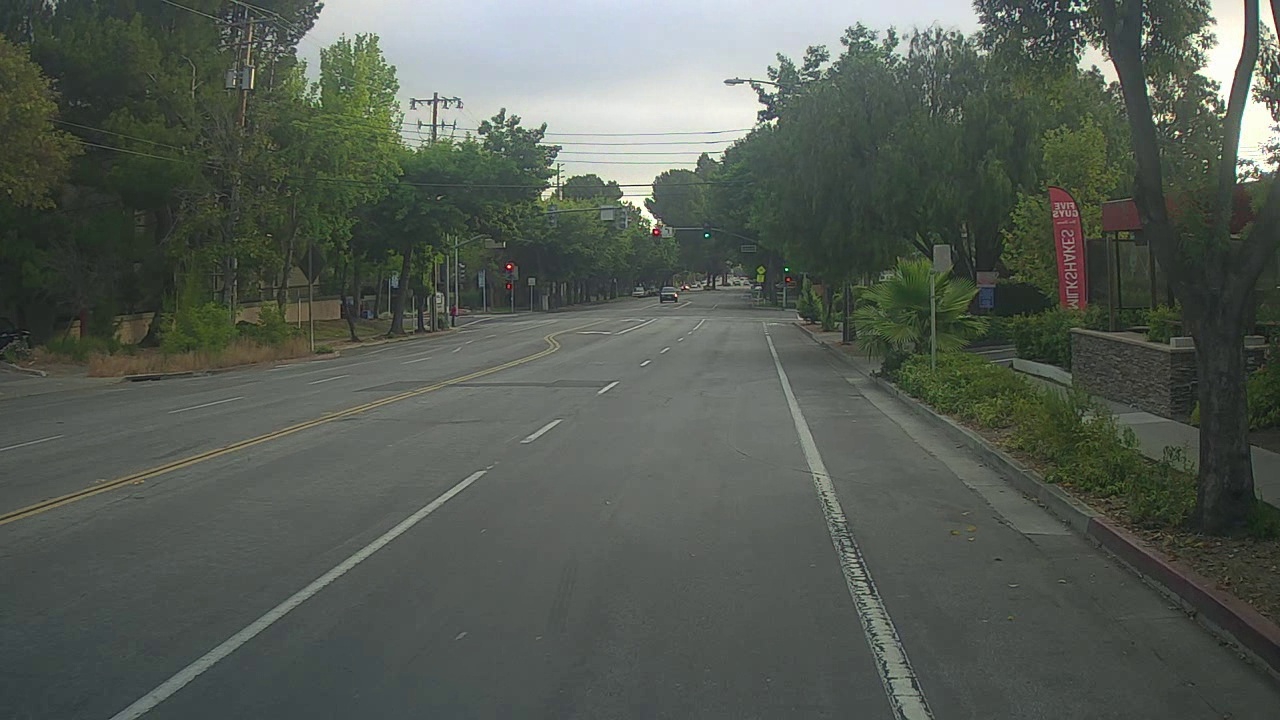}
     \put(5,5){\textcolor{white}{Frame 60, YOLOv2}}
     \end{overpic}
     \begin{overpic}[width=0.49\textwidth]{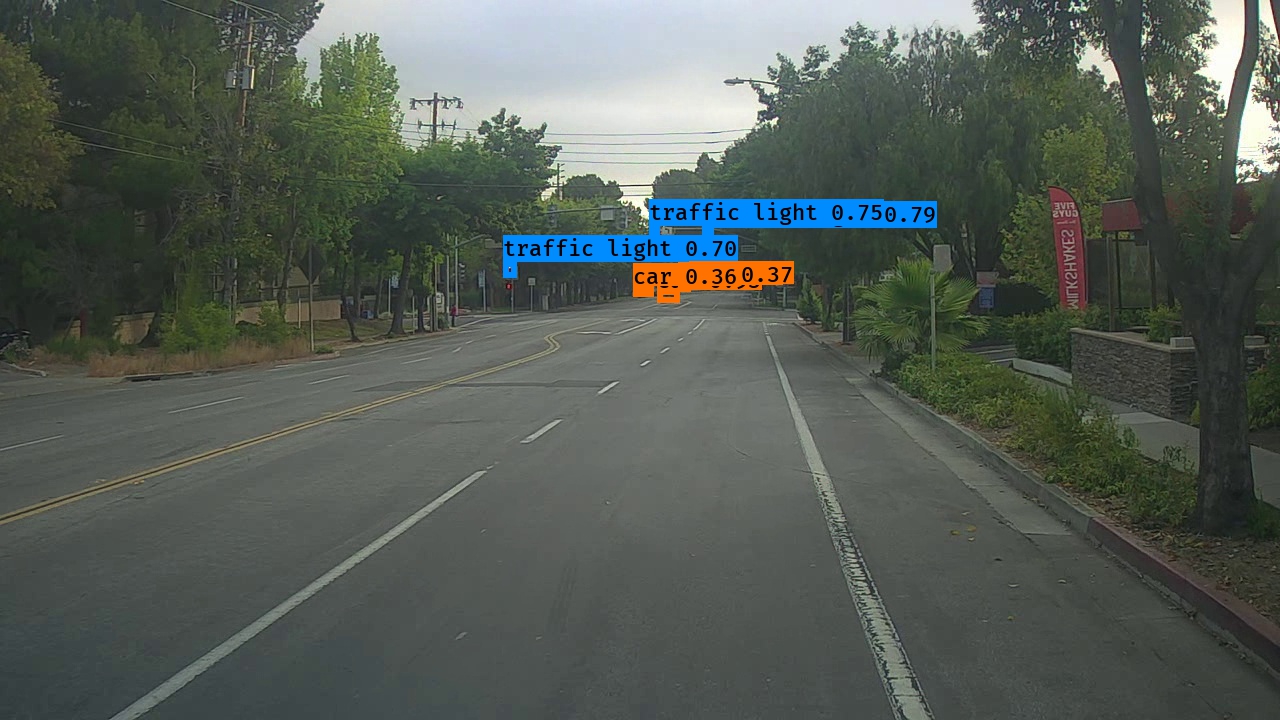}
     \put(5,5){\textcolor{white}{Frame 60, YOLOv3}}
     \end{overpic}
     \begin{overpic}[width=0.49\textwidth]{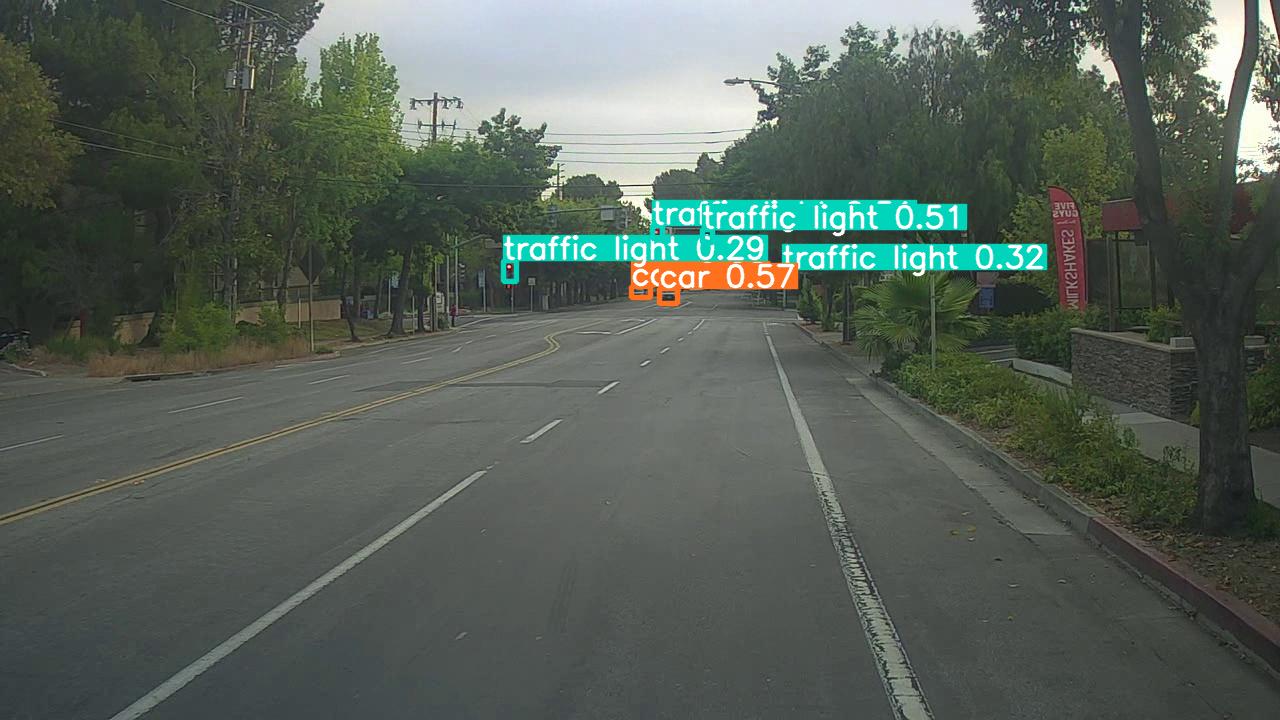}
     \put(5,5){\textcolor{white}{Frame 60, YOLOv5 (l)}}
     \end{overpic}
     \begin{overpic}[width=0.49\textwidth]{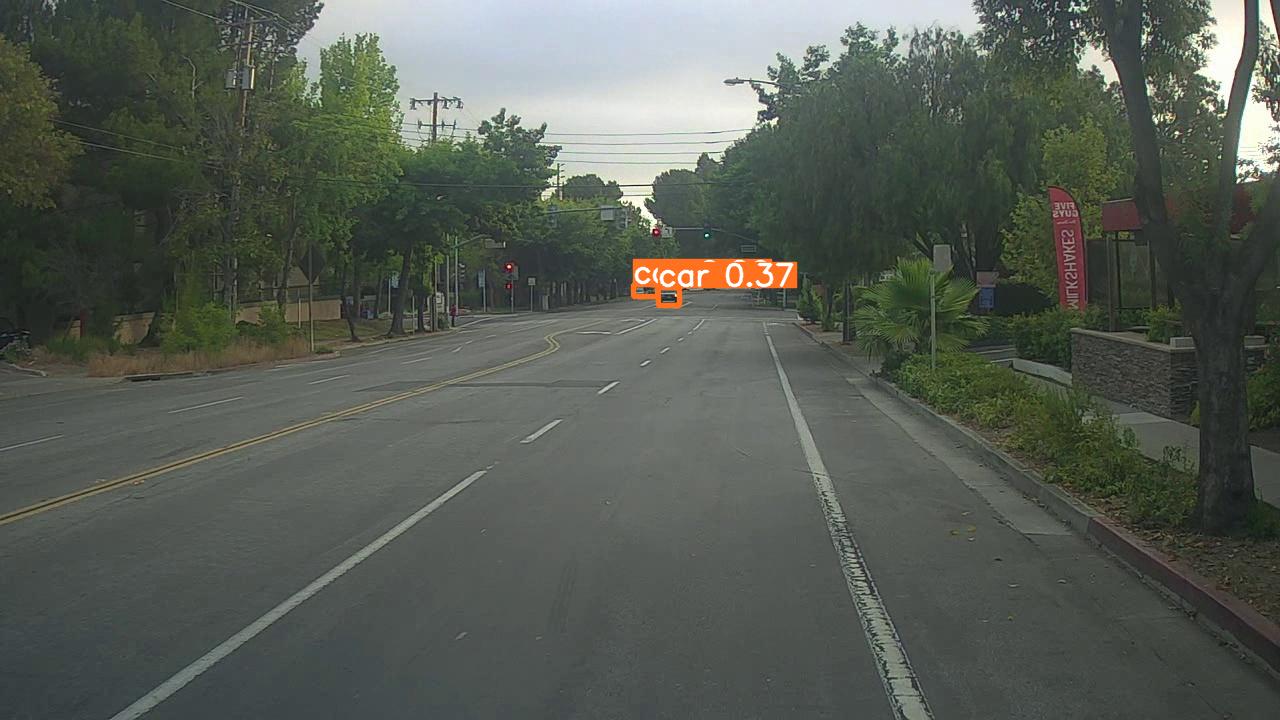}
     \put(5,5){\textcolor{white}{Frame 60, YOLOv8 (l)}}
     \end{overpic}
     \begin{overpic}[width=0.49\textwidth]{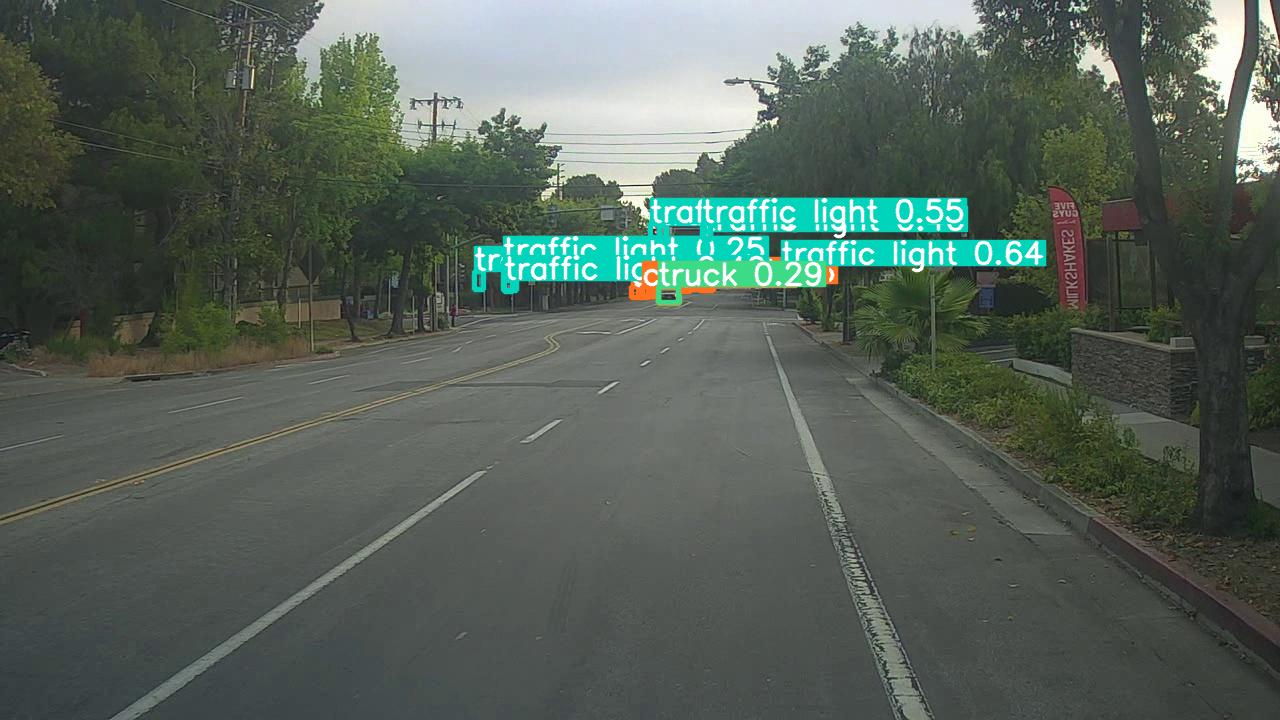}
     \put(5,5){\textcolor{white}{Frame 60, RT-DETR (l)}}
     \end{overpic}
        \caption{Deep learning vision object detection models applied to one example scene in CA.}
        \label{fig:appUS}
\end{figure}

The general impression recognition ability varies strongly across the model, making it very hard for some models to detect small (or far away) objects. The recognition of these objects is not a direct safety issue, but can enhance the track planning by considering potential obstacles with long enough latency (especially at high driving speeds). In the selected example for illustration, the pre-trained YOLOv2 was unable to detect the distant cars and traffic lights (see Fig.~\ref{fig:appUS}. YOLOv3 discovered the driving, one parking car and three out of four traffic lights. YOLOv5 (large) discovered the driving, one parking car and the four traffic lights. In this scene, YOLOv8 (large) detected the two cars also found by the pre-trained YOLOv3 and YOLOv5; the traffic lights remained undetected. RT-DETR (large) found 6 cars, the traffic lights and the road signs, performing at best for this frame. The driving one was classified as truck, having qualitatively little impact on the safety and planning strategy along the journey (with high potential being classified as car as it is approaching). This first qualitative evaluation shows that for pre-trained (and not fine-tuned) YOLOv3/5/8 models cars are likely detected objects, where additional attention must be put on traffic signs detection to be able to comply with traffic rules.

As a next step, the flavors of the YOLOv5 and YOLOv8 model are discussed. For both YOLO versions, the (n) flavor was unable to detect the objects. The YOLOv8 (s) detected only one traffic sign, and the YOLOv8 (m/l/x) detected the driving car and some parking cars (see appendix). Overall, the performance of the pre-trained YOLOv8 is unsatisfactory for this scene. YOLOv5 (s) located three of four traffic lights and the driving car. YOLOv5 (m/l/x) found the main characteristics. With one wrongly detected person near the road in YOLOv5 (x), posing a risk for a potential car trajectory and significantly alters the future trajectory (potentially slowing down the car). Both RT-DETR flavors performed well. RT-DETR (x) detected one bicycle near the road wrong. 
As the objects are approaching the car, the models (YOLOv3, YOLOv5, YOLOv8, RT-DETR) detected the most important objects (car, traffic light, pedestrians, ...). YOLOv2 gradually performed worse compared to the other models. These very general statements will be verified in a detailed follow-up study and should not be taken granted. 
As illustrated in the appendix (see examples in the appendix \ref{sec:append1}, \ref{sec:append2}, \ref{sec:append3}), during the passage of car-objects and traffic light objects YOLOv5, YOLOv8, and RT-DETR performed reasonable on the example scenes. Some object classes being close to each other (in terms of classification), like a car, a bus or a truck, are frequently mixed up by the model depending on the perspective. This issue might be corrected by object tracking or multi-modality easily.
The general qualitative impression was that cars and trucks are well detected (also by the early YOLOv2 and the nano and small flavors of YOLOv5 and YOLOv8) if they occur in the typical size (not too small) regarding the background. 
More difficult (compared to cars) was the correct detection of the traffic signs especially for YOLOv2 and the nano and small flavors of YOLOv5 and YOLOv8, which is an important issue to be considered when fine-tuning such models. 
Another important aspect is the indication of pedestrians on the road and walking paths. In general, there were only two pedestrians in the data leading to a detection rate of one for all YOLO and RT-DETR architectures (since one pedestrian is a tiny feature relatively far away at the next crossing).


\section{Testing - AUT} 
Similar as for the USA-CA dataset, a first qualitative comparison for all models (YOLOv2, YOLOv3, YOLOv5, YOLOv8, RT-DETR) will be performed with the pre-trained models. 

\subsection{Dataset} 
In the dataset, several images are collected by a smartphone camera from a car. The images are taken every few seconds. The scenes (with snowy background) analyzed exemplarily below shows images taken while driving around in winter Austria. The most illustrative representations of the dataset are reported visually. 

\subsection{Results}

\begin{figure}[ht!]
     \centering
     \begin{overpic}[width=0.45\textwidth]{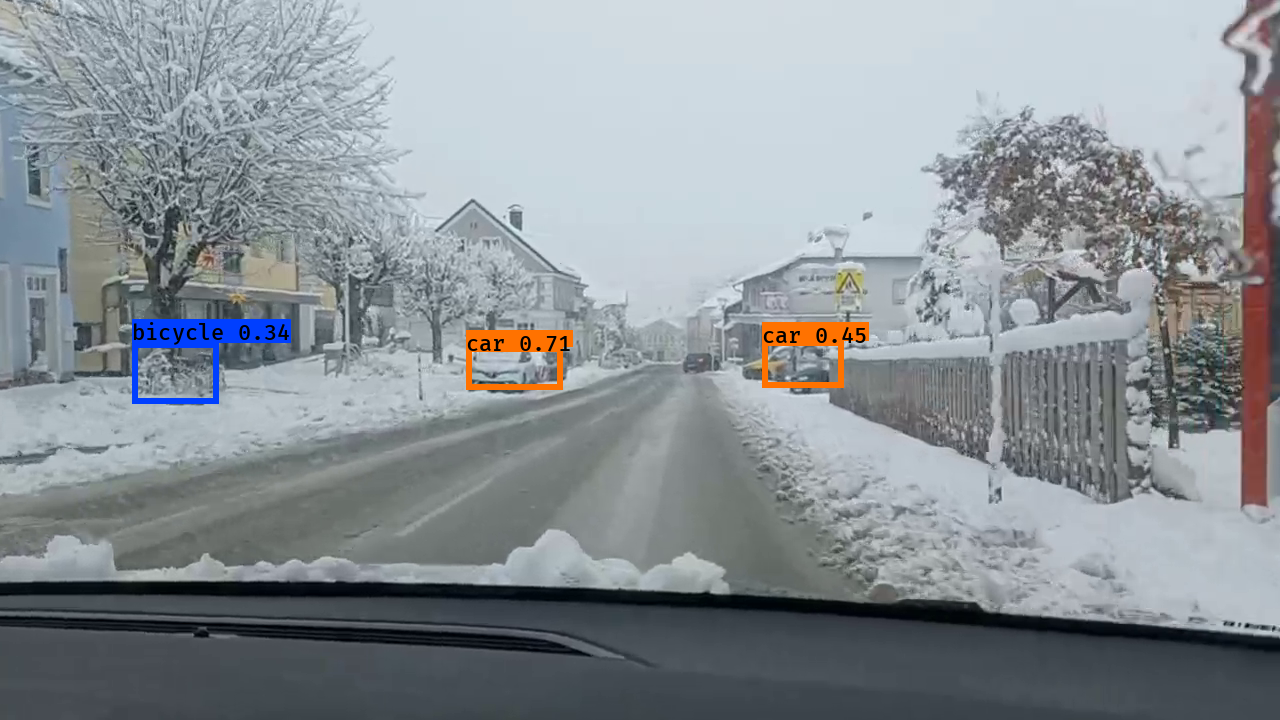}
     \put(5,5){\textcolor{white}{Frame 11, YOLOv2}}
     \end{overpic}
     \begin{overpic}[width=0.45\textwidth]{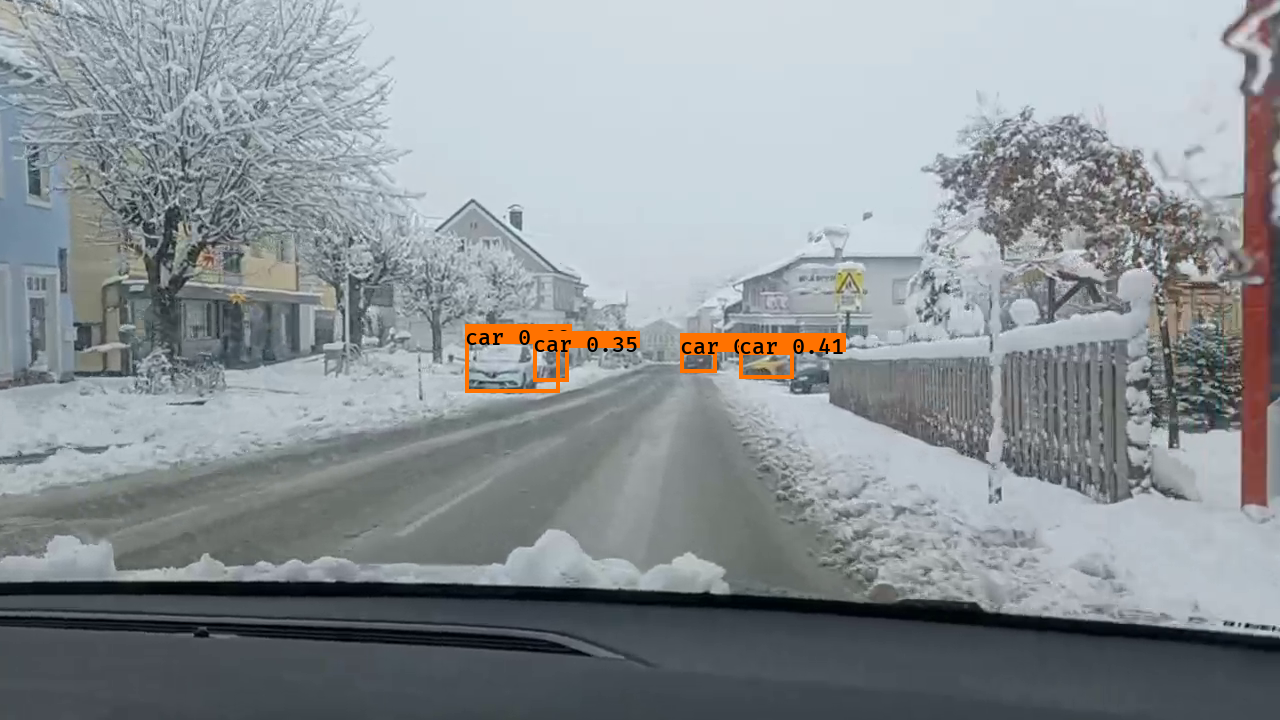}
     \put(5,5){\textcolor{white}{Frame 11, YOLOv3}}
     \end{overpic}
     \begin{overpic}[width=0.45\textwidth]{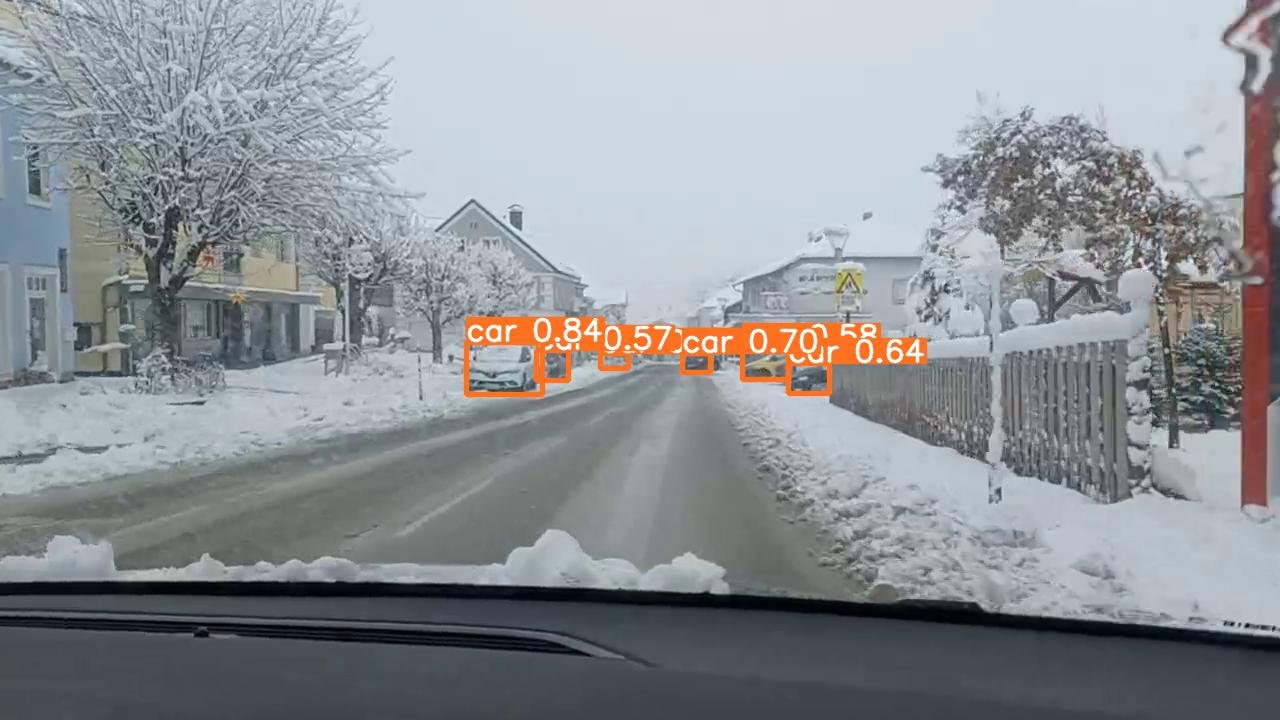}
     \put(5,5){\textcolor{white}{Frame 11, YOLOv5 (l)}}
     \end{overpic}
     \begin{overpic}[width=0.45\textwidth]{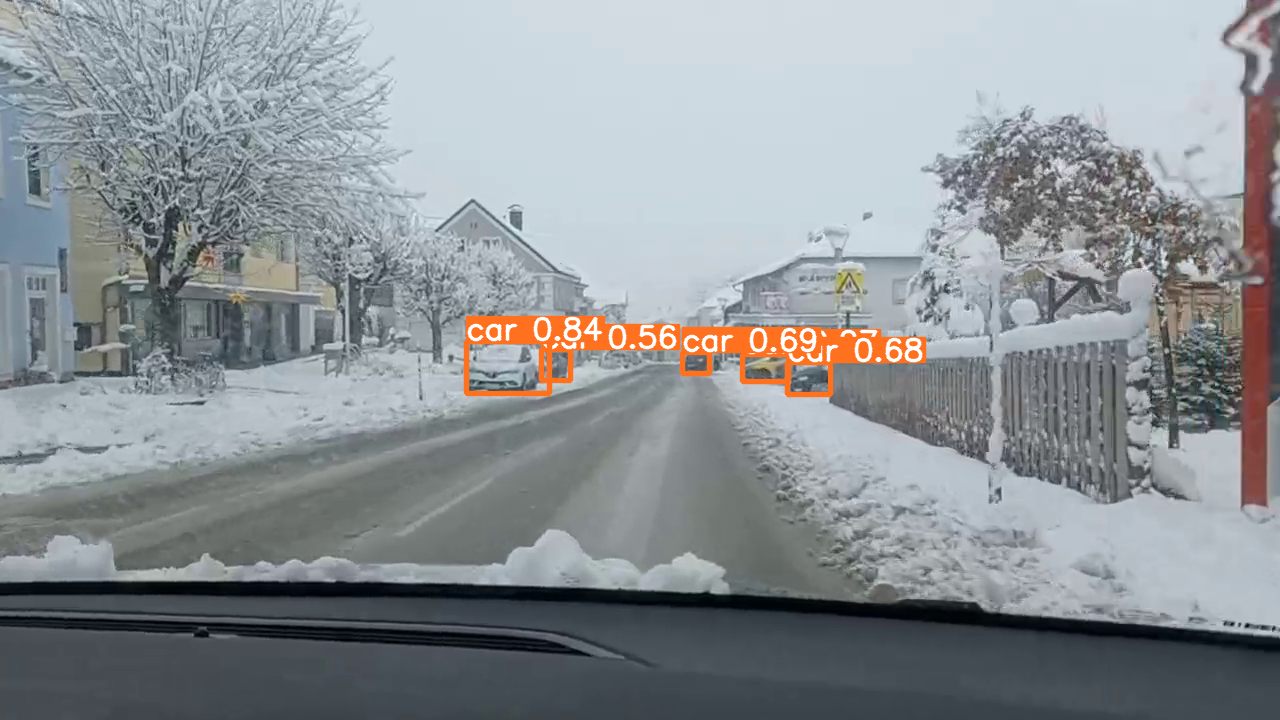}
     \put(5,5){\textcolor{white}{Frame 11, YOLOv8 (l)}}
     \end{overpic}
     \begin{overpic}[width=0.45\textwidth]{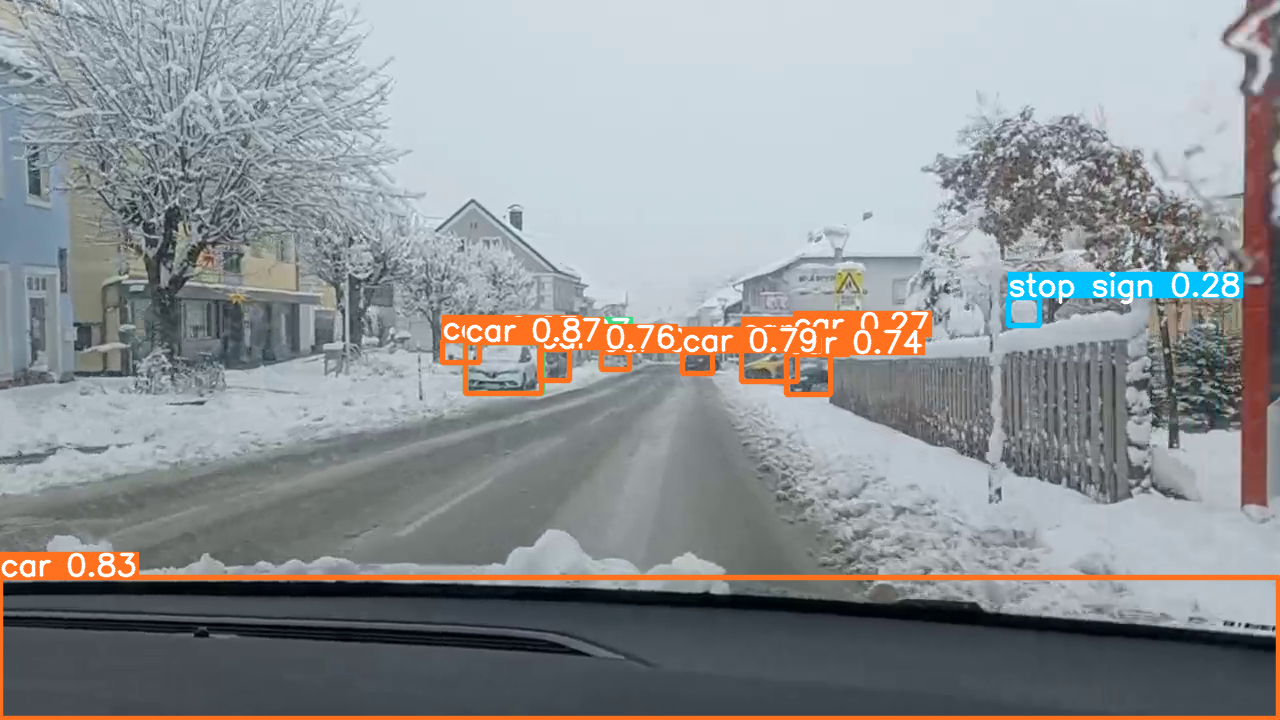}
     \put(5,5){\textcolor{white}{Frame 11, RT-DETR (l)}}
     \end{overpic}
     \begin{overpic}[width=0.45\textwidth]{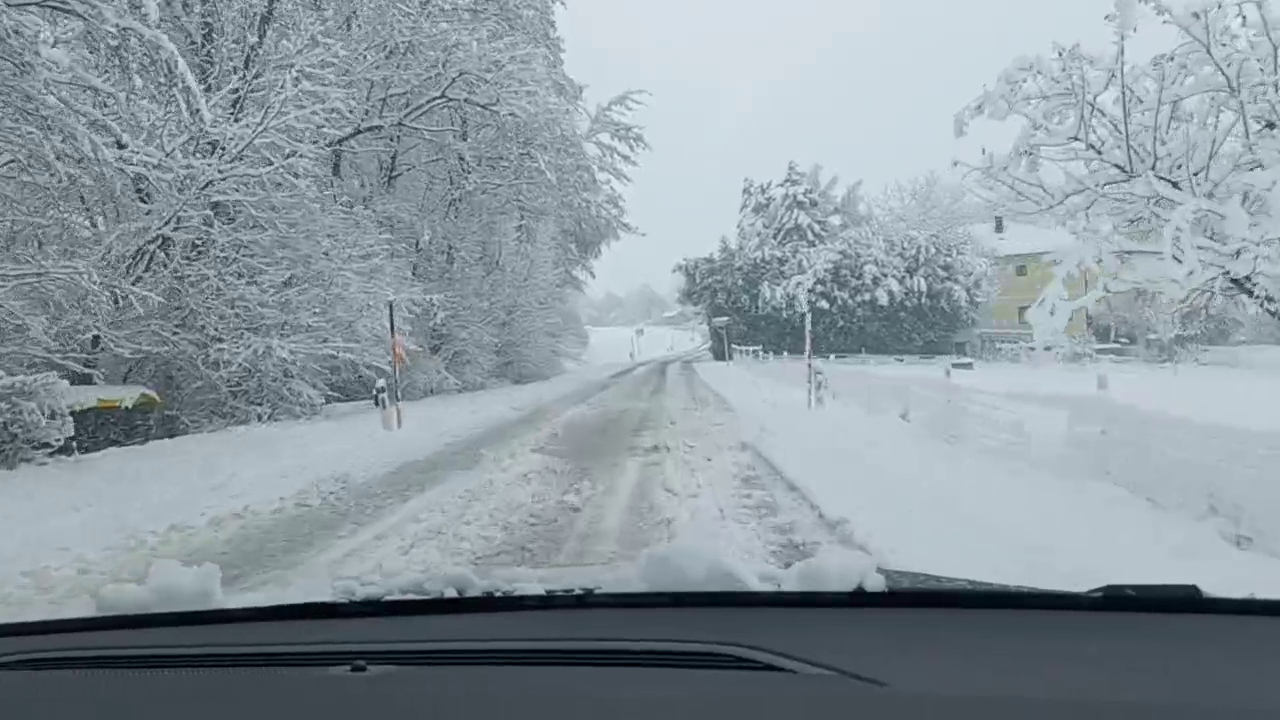}
     \put(5,5){\textcolor{white}{Frame 50, YOLOv2}}
     \end{overpic}
     \begin{overpic}[width=0.45\textwidth]{img/AT/Yv2/0050.png}
     \put(5,5){\textcolor{white}{Frame 50, YOLOv3}}
     \end{overpic}
     \begin{overpic}[width=0.45\textwidth]{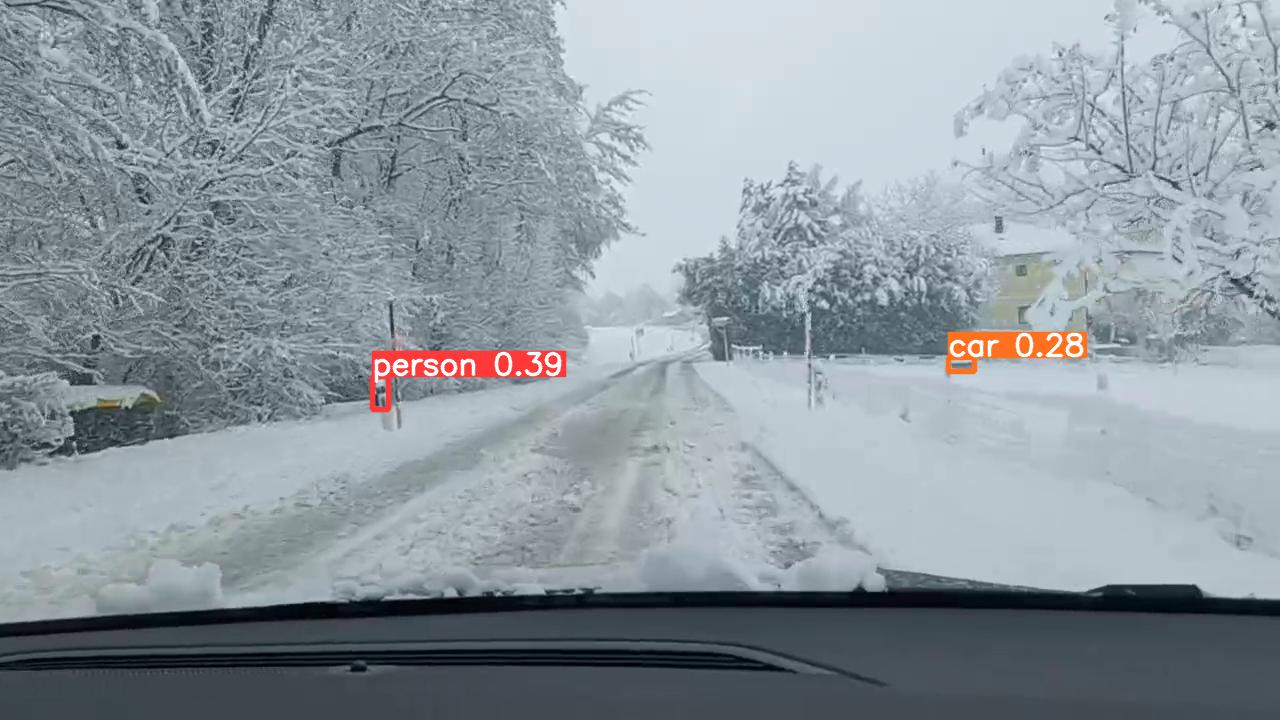}
     \put(5,5){\textcolor{white}{Frame 50, YOLOv5 (l)}}
     \end{overpic}
     \begin{overpic}[width=0.45\textwidth]{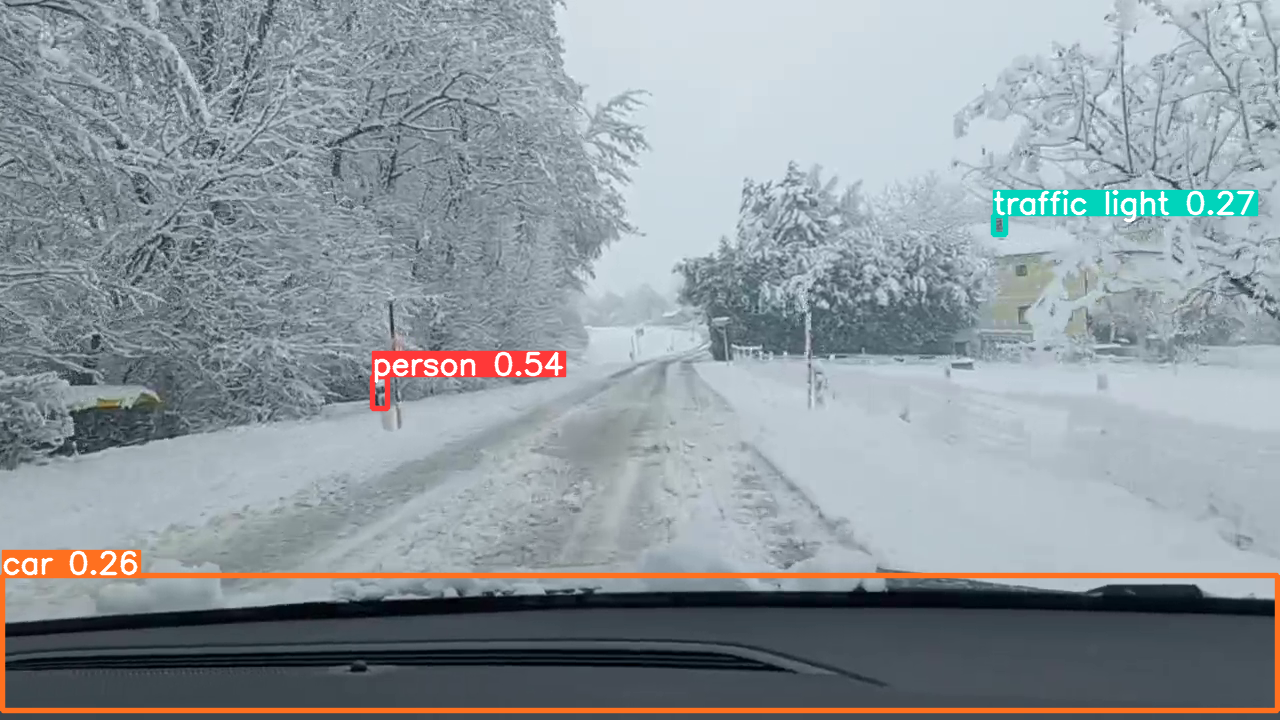}
     \put(5,5){\textcolor{white}{Frame 50, YOLOv8 (l)}}
     \end{overpic}
     \begin{overpic}[width=0.45\textwidth]{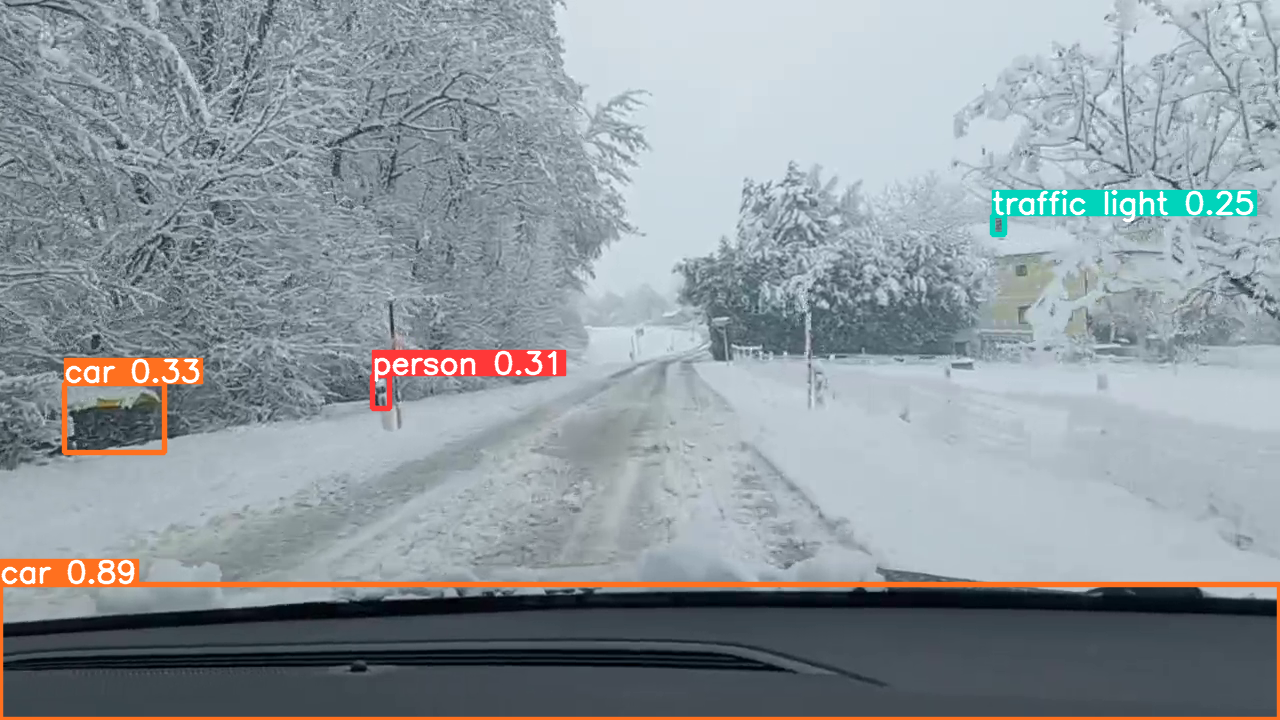}
     \put(5,5){\textcolor{white}{Frame 50, RT-DETR (l)}}
     \end{overpic}
        \caption{Deep learning vision object detection models applied to two example scene in AUT.}
        \label{fig:appAT}
\end{figure}

Figure \ref{fig:appAT} shows a typical scene encountered in Austria (AUT) during winter time. The main characterisitcs and difference to typical landscape is the dominant white background and the shielding of important characteristics of objects by snow and ice, leading to confused recognition of objects. During the object detection, a wide range of objects were seeded into the images with no relation about to reality. Among them were trains, cars, traffic signs, traffic lights, pedestrians, bicycles, a ship, benches, chairs, umbrellas, a kite, trucks actually being a house with snow and so on. Interestingly, the ice in front of the camera was sometimes identified as a bird or person. The only person in the scene was recognized by every algorithm. A general tendency was observed that pre-trained model with more parameters tend to seed more artifacts into the images. Some of the examples can be seen in the appendix \ref{sec:append4}, \ref{sec:append5}, \ref{sec:append6}, and \ref{sec:append7}.

Despite the strong model confusion, the impression based on the road scenes from the California also applies to Austria. The tendency that a model with more parameters is capable of detecting more relevant objects, but additionally it is also prone to finding many false positives were recognized. On the one hand, recognizing more objects can lead to better safety and a better planning of the route. On the other hand, the high number of false positives will significantly influence the route planning of an algorithm and can even cause unfavorable maneuvers.  
Overall, the car detection capability was good, whenever the camera was in the upright position. Whenever it tilted, the detection rate was weaker. The detection rate was also weaker for very close objects (nearly passed car). Moreover, the only pedestrian in the dataset was recognized by all algorithms.
During the whole dataset, it was impossible for the algorithm to detect traffic signs and road signs. In the vast majority, this was attributed to the different layout of the signs and it seemed that the snow and ice coverage of signs significantly worsened the performance. To conclude, pre-trained deep learning vision models have to be fine-tuned with specific data for Austria and especially under alpine circumstances.

\section{Conclusion}

This first qualitative study showed the capabilities of fast deep learning vision models applied to various road traffic scenes in the US and Austria. Generally, the algorithms are relatively well in detecting cars, also in winter scenes. Early YOLO models and YOLO models with relatively few parameters were found to perform poor if the pre-trained weights are used. Especially in the Austrian road traffic scenes with winter landscape and snow, many erroneous detections occurred during the qualitative test. All models were detecting objects at location where they are not present (trains, people, ships, ...). Most likely, these faulty detections were based on the unusual background of white snow for most objects. This can lead to big troubles in automated driving and fusing this information into the model. This first qualitative study shows that especially the alpine landscape is a potential issue for fast object detection algorithms (even for the state-of-art RT-DETR).

\section{Outlook}

Possible improvement to this first qualitative comparison of pretrained models are manifold, and include the algorithm selection 
(like comparison to a accurate Single Shot Detector SSD512 and a fast Single Shot Detector SSD300) 
and optimization. Fine-tuning the algorithms to adapt to Austrian road characteristics and seasonality. Analyzing the algorithm's capability to recognize and interpret Austrian traffic signs and adhere to local traffic regulations, contributing to safe and compliant driving behavior. Followed by rigorous testing and evaluation of the object detection system under different scenarios encountered on Austrian roads. Metrics such precision, recall, and F1 score will be used to assess the algorithm's accuracy based on labeled data (e.g. labeled with CVAT \cite{CVAT_ai_Corporation_Computer_Vision_Annotation_2023}). Furthermore, assessing the feasibility of real-time implementation of the object detection system into the vehicular environment (addressing computational constraints and latency issues).

\subsection{Deep learning models and multi-modality}

A big step, towards a robust system is to include the multi-modal behavior of the sensor streams by sensor fusion using multi-modal models \cite{pham2023nvautonet}. There, images, video streams, audio, LiDAR, RADAR 
and re-projecting the bird's-eye view (BEV) of a camera into real-physical coordinates are combined to classification of hazards, road segmentation
(e.g. with improved \cite{teichmann2018multinet,zhao2023fast}), 2D-object detection, 3D-object detection
, distance estimation, positioning and parking slot recognition. Images, video streams, audio were fused to do multi-modal dept recognition from the KITTY dataset in \cite{srivastava2023omnivec} with a transformer. Additionally, combining image, depth information, 3D point clouds, videos, audio \cite{gong2022ssast}, text description \cite{devlin2018bert} (of planned route as text input), and GPS data would enhance the safety (e.g. openpilot project\footnote{https://github.com/commaai/openpilot}) and the comfort of passengers.

\subsection{Datasets}

A short review of available datasets for future quantitative comparison and sensor fusion is provided here. 
The \href{https://www.cvlibs.net/datasets/kitti/}{KITTI vision benchmark suite dataset} \cite{Geiger2013IJRR} and \href{https://www.cvlibs.net/datasets/kitti-360/}{ KITTI-360} \cite{Liao2022PAMI} were introduced with the explicit purpose of propelling research in autonomous driving through a pioneering collection of real-world computer vision benchmarks. As one of the earliest datasets in the field of autonomous driving, KITTI has garnered over 8200 academic citations and continues to be widely utilized.
Noteworthy in this context are a number of other datasets, like the \href{https://www.a2d2.audi/a2d2/en/dataset.html}{Audi Autonomous Driving Dataset (A2D2)} features over 41,000 frames labeled with 38 features and additional 390,000 frames unlabeled. More than 2.3 TB in total, containing 2D semantic segmentation, 3D point clouds, 3D bounding boxes, and vehicle bus data. The type of landscape is comparable to the urban and rural one in Austria and Germany. 
\href{https://www.cityscapes-dataset.com}{CityScapes} stands out as a substantial dataset with a primary focus on semantic comprehension of urban street scenes in 50 German and Swiss cities \cite{Cordts2016Cityscapes}. This comprehensive resource includes semantic, instance-wise, and dense pixel annotations for 30 classes categorized into 8 groups. The dataset comprises a total of 5,000 annotated images with meticulous fine annotations, supplemented by an additional 20,000 annotated images featuring broader, coarse annotations. 
\href{https://apolloscape.auto}{ApolloScape} is a dynamic research endeavor focused on propelling innovation throughout the spectrum of autonomous driving, spanning from perception to navigation and control. Through their website, users have the opportunity to delve into an array of simulation tools, along with access to an extensive collection of resources, including over 100,000 street view frames, an 80,000-point lidar cloud, and 1000 kilometers of trajectories tailored for urban traffic scenarios \cite{wang2019apolloscape}. The \href{https://www.argoverse.org/index.html}{Agroverse} project consists of two datasets. In the Argoverse 1 dataset 3D tracking annotations for 113 scenes and over 324,000 unique vehicle trajectories for motion forecasting are included \cite{Argoverse}. The Agroverse 2 dataset significantly extended the scope of the data \cite{Argoverse2}. 
\href{https://bdd-data.berkeley.edu}{Berkeley DeepDrive Dataset (BDD 100K)}, provides users with access to a vast resource comprising 100,000 annotated videos and 10 tasks specifically designed for evaluating image recognition algorithms in the context of autonomous driving \cite{yu2020bdd100k}. This dataset encapsulates over 1000 hours of driving experience, encompassing a staggering 100 million frames. Additionally, it incorporates valuable information pertaining to geographic locations, environmental conditions, and weather diversity.
The Comma2k19 Dataset encompasses 33 hours of commuting footage recorded on Highway 280 in California \cite{1812.05752}. Each 1-minute scene within this dataset was captured along a 20km stretch of highway between San Jose and San Francisco. The data collection utilized comma EONs, equipped with a road-facing camera, phone GPS, thermometers, and a 9-axis IMU.

The \href{https://leddartech.com/solutions/leddar-pixset-dataset/}{Leddar PixSet} was released in 2021 and publicly available dataset tailored for autonomous driving research and development \cite{deziel2021pixset}. This dataset encompasses a comprehensive set of data from a full autonomous vehicle (AV) sensor suite, incorporating cameras, LiDARs, radar, and IMU. Notably, it includes full-waveform data from the Leddar Pixell, a cutting-edge 3D solid-state flash LiDAR sensor. The Leddar PixSet comprises 29,000 frames organized into 97 sequences, with the added benefit of more than 1.3 million annotated 3D boxes. 
The \href{https://www.nuscenes.org}{nuScenes} dataset stands as one of the most extensive open-source datasets designed for advancing autonomous driving research \cite{nuscenes}. Captured in both Boston and Singapore, the dataset utilizes a comprehensive sensor suite, including a 32-beam LiDAR, 6 360° cameras, and radars. With over 1.44 million camera images, the nuScenes dataset captures a diverse array of traffic scenarios, driving maneuvers, and unexpected behaviors. Notably, the dataset includes examples of images taken under various conditions such as clear weather, nighttime, rain, and construction zones.
The \href{https://robotcar-dataset.robots.ox.ac.uk}{Oxford RobotCar Dataset} comprises over 100 recordings capturing a consistent route through Oxford, UK, recorded over a period spanning more than a year \cite{RobotCarDatasetIJRR}. This dataset offers a comprehensive representation of various environmental conditions, including diverse weather patterns, traffic scenarios, and pedestrian activities. Notably, the dataset encompasses longer-term changes such as construction activities and roadworks.
\href{https://public.roboflow.com/object-detection/self-driving-car}{Udacity Self Driving Car Dataset} has generously open-sourced a diverse array of projects geared towards autonomous driving. These resources include neural networks specifically trained for predicting steering angles of the car, camera mounts, and extensive datasets consisting of dozens of hours of real driving data.
The \href{https://waymo.com/open/}{Waymo Open dataset} is a freely accessible multimodal sensor dataset tailored for autonomous driving. Derived from Waymo's self-driving vehicles, the dataset encompasses a diverse range of driving scenarios and environmental conditions. It comprises 1000 types of distinct segments, each capturing 20 seconds of continuous driving. This translates to a total of 200,000 frames, captured at a rate of 10 Hz per sensor.
Based on this variety of open datasets, challenges and benchmarks, detailed studies will be performed in the future.


\section*{Data availability}
The datasets used for this qualitative study are in the current status available under request to the author.

\section*{Source code availability}
The source code used for this qualitative study are in the current status available under request to the author.

\bibliographystyle{abbrv}
\bibliography{references}  

\newpage
\appendix
\section{CA - YOLOv5 flavors} \label{sec:append1}

\begin{figure}[ht!]
     \centering
     \begin{overpic}[width=0.45\textwidth]{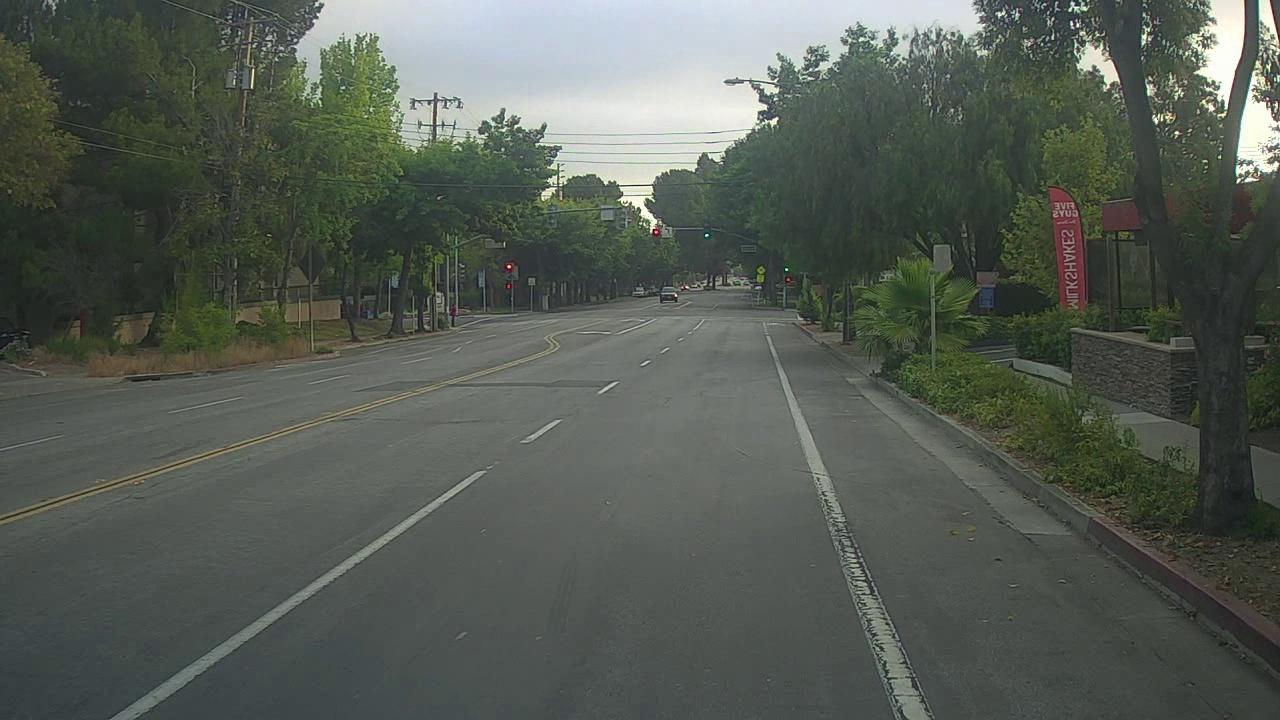}
     \put(5,5){\textcolor{white}{Frame 60, (n)}}
     \end{overpic}
     \begin{overpic}[width=0.45\textwidth]{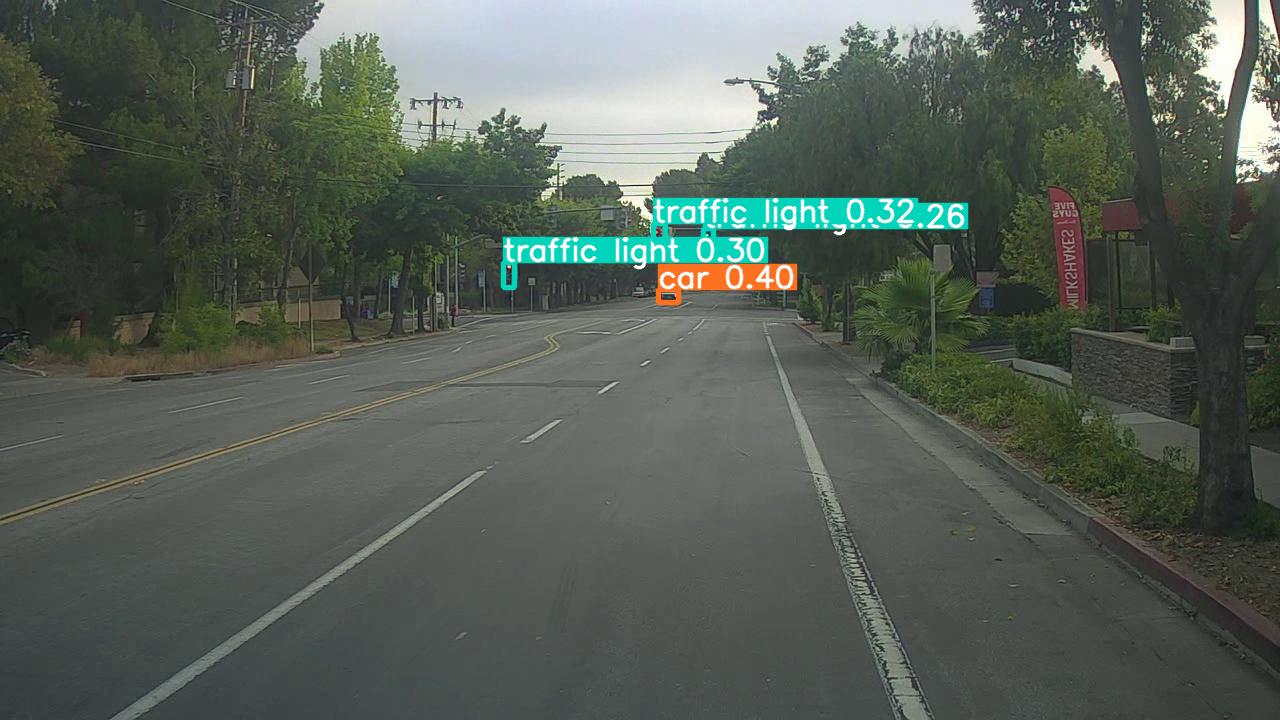}
     \put(5,5){\textcolor{white}{Frame 60, (s)}}
     \end{overpic}
     \begin{overpic}[width=0.45\textwidth]{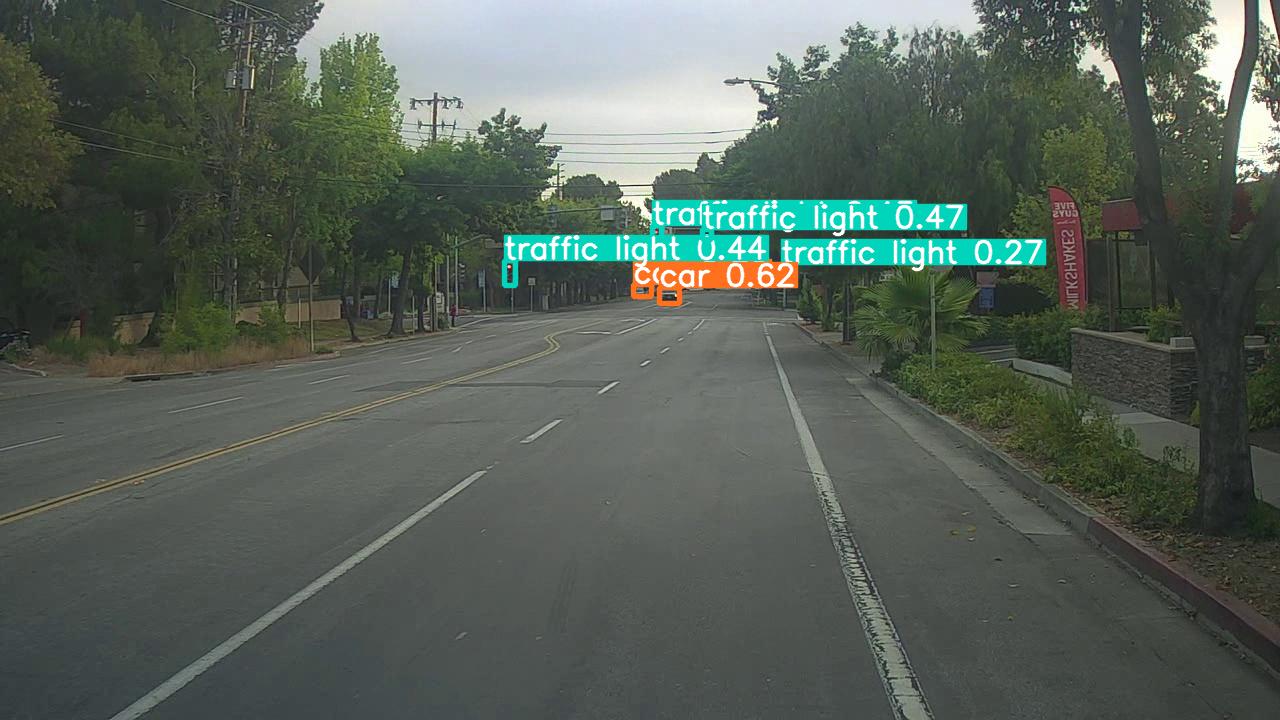}
     \put(5,5){\textcolor{white}{Frame 60, (m)}}
     \end{overpic}
     \begin{overpic}[width=0.45\textwidth]{img/CA/Yv5l/0060.jpg}
     \put(5,5){\textcolor{white}{Frame 60, (l)}}
     \end{overpic}
     \begin{overpic}[width=0.45\textwidth]{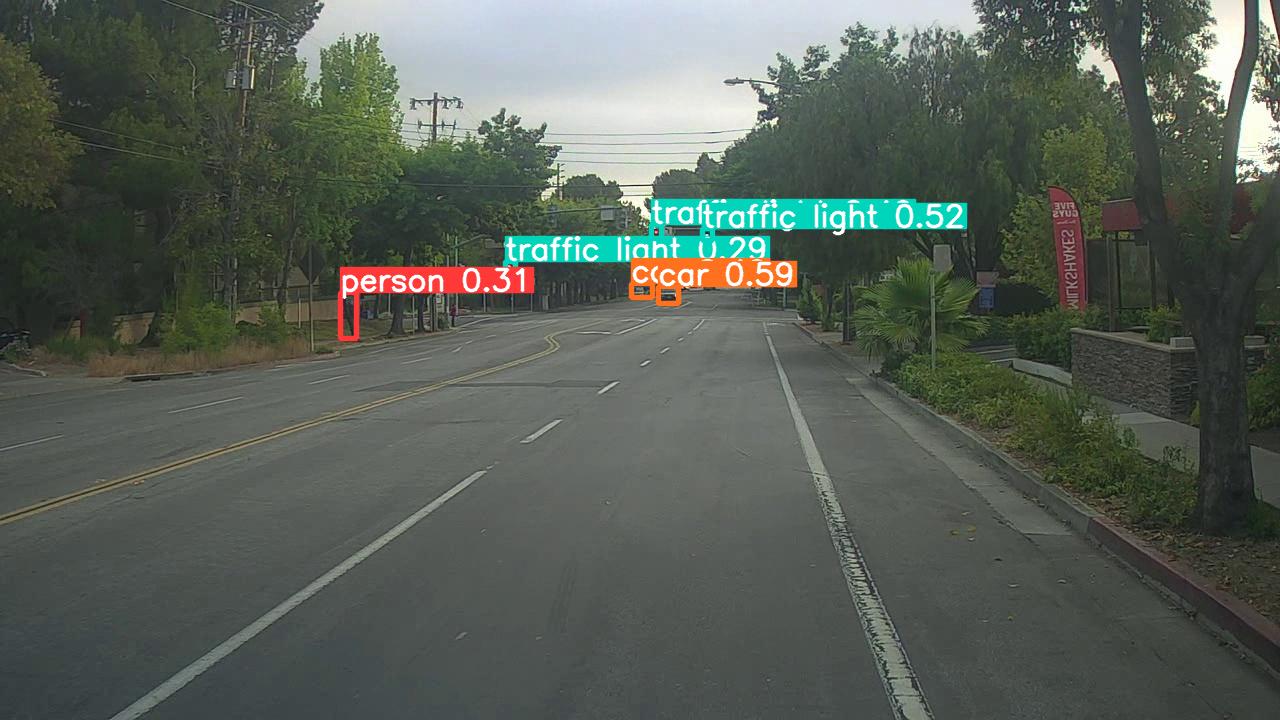}
     \put(5,5){\textcolor{white}{Frame 60, (x)}}
     \end{overpic}
     \begin{overpic}[width=0.45\textwidth]{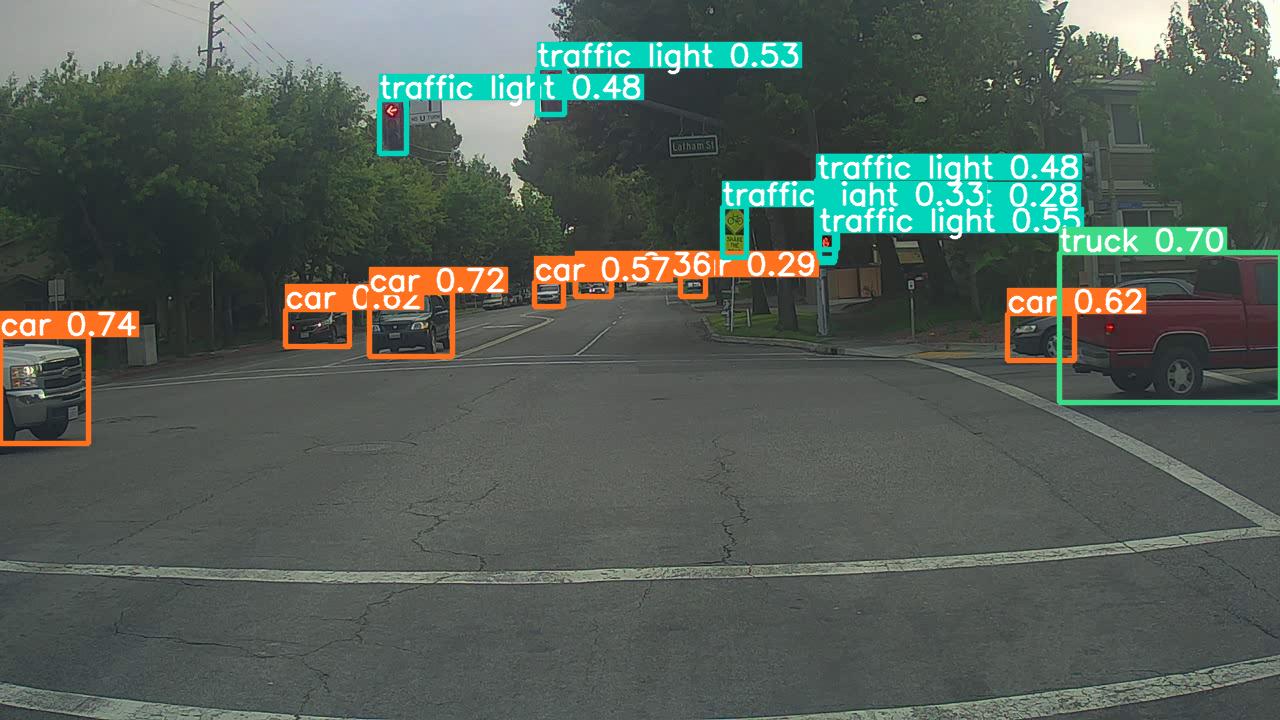}
     \put(5,5){\textcolor{white}{Frame 72, (n)}}
     \end{overpic}
     \begin{overpic}[width=0.45\textwidth]{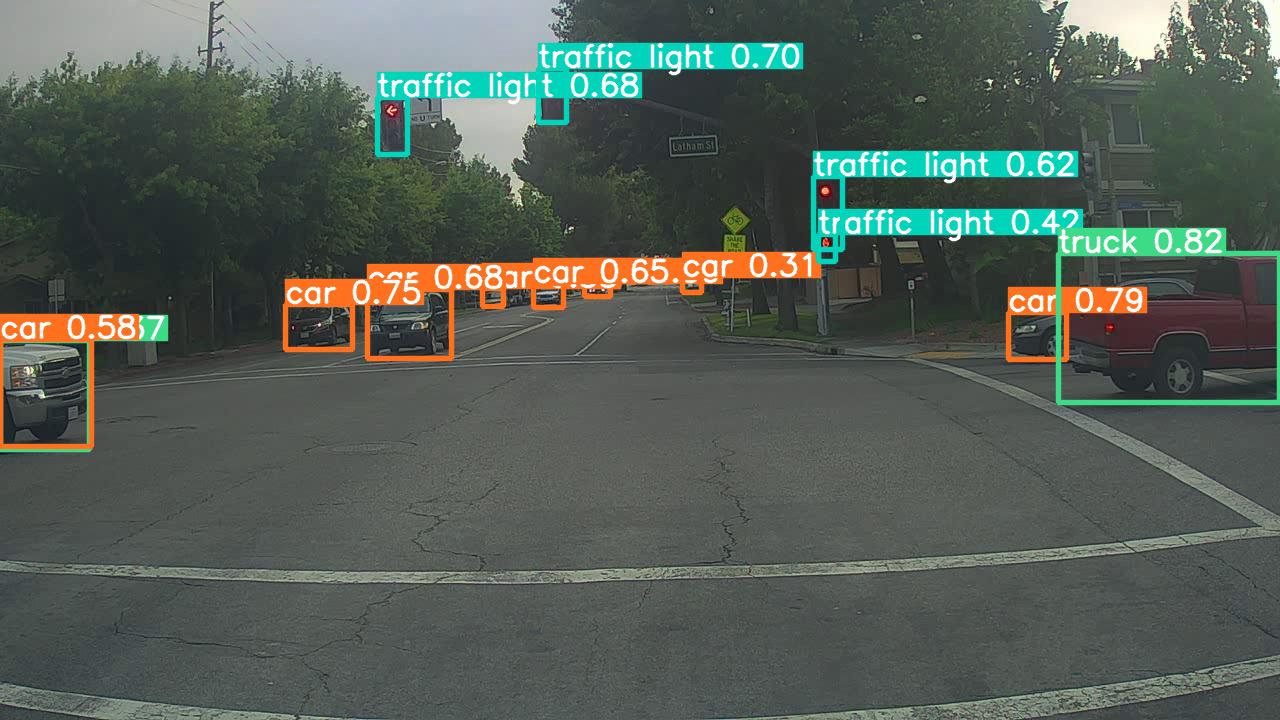}
     \put(5,5){\textcolor{white}{Frame 72, (s)}}
     \end{overpic}
     \begin{overpic}[width=0.45\textwidth]{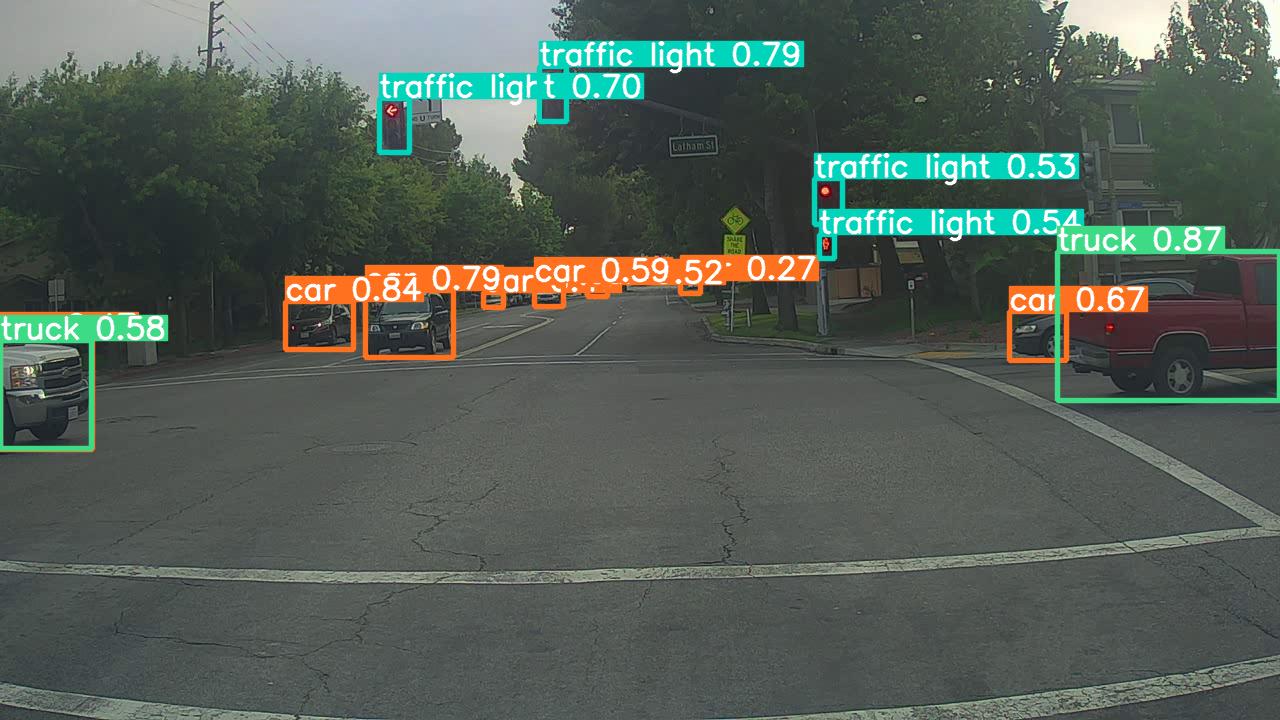}
     \put(5,5){\textcolor{white}{Frame 72, (m)}}
     \end{overpic}
     \begin{overpic}[width=0.45\textwidth]{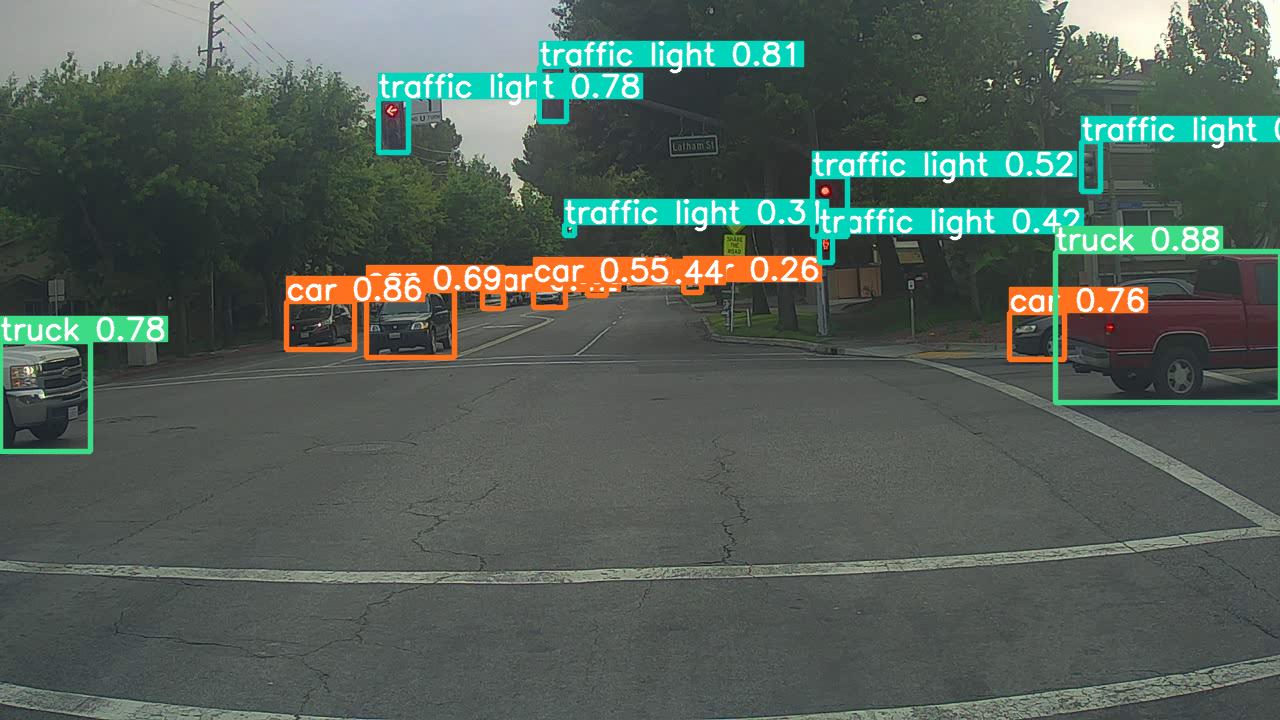}
     \put(5,5){\textcolor{white}{Frame 72, (l)}}
     \end{overpic}
     \begin{overpic}[width=0.45\textwidth]{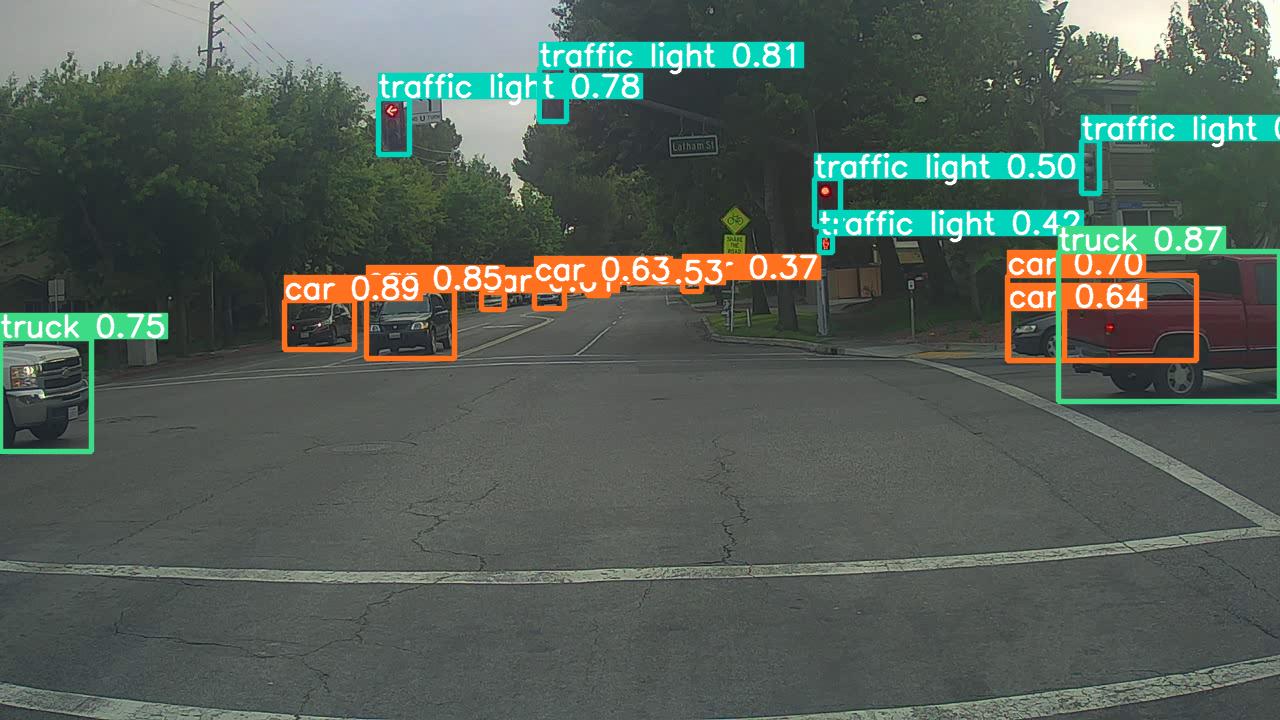}
     \put(5,5){\textcolor{white}{Frame 72, (x)}}
     \end{overpic}
        \caption{YOLOv5 flavors applied to two example scenes in CA.}
        \label{fig:appUS_v5}
\end{figure}

\newpage
\section{CA - YOLOv8 flavors} \label{sec:append2}
\begin{figure}[ht!]
     \centering
     \begin{overpic}[width=0.45\textwidth]{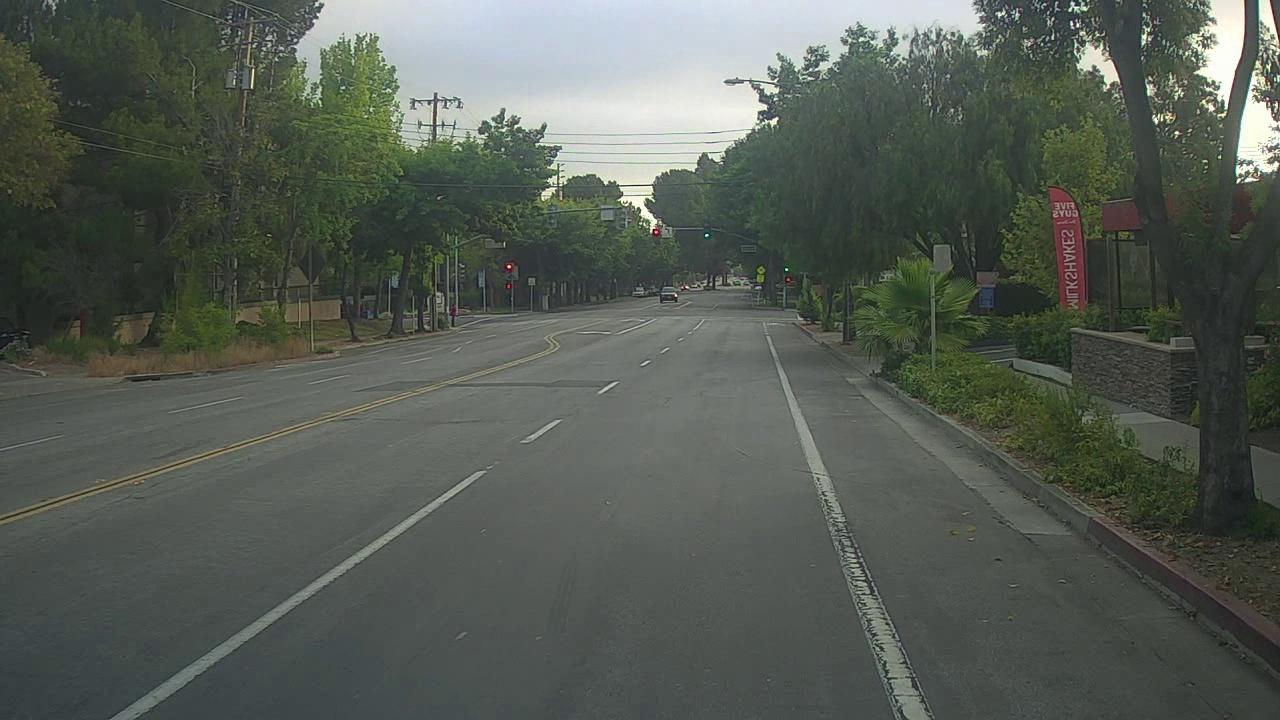}
     \put(5,5){\textcolor{white}{Frame 60, (n)}}
     \end{overpic}
     \begin{overpic}[width=0.45\textwidth]{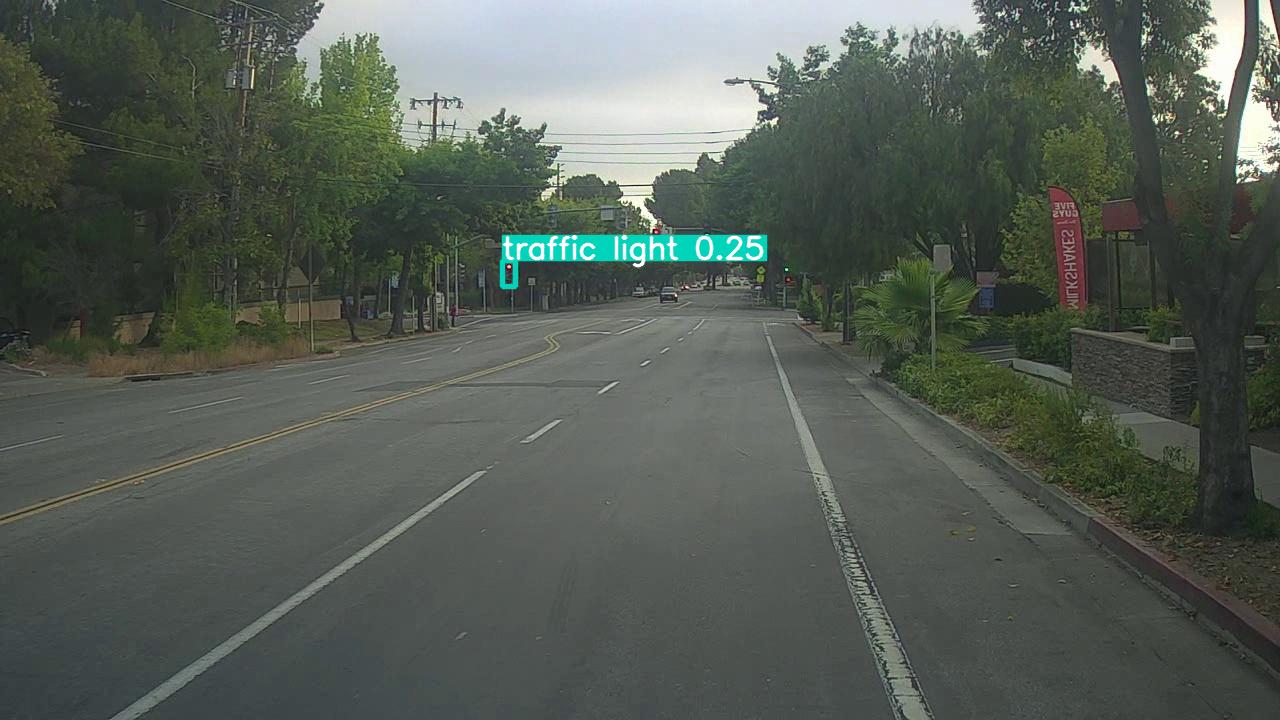}
     \put(5,5){\textcolor{white}{Frame 60, (s)}}
     \end{overpic}
     \begin{overpic}[width=0.45\textwidth]{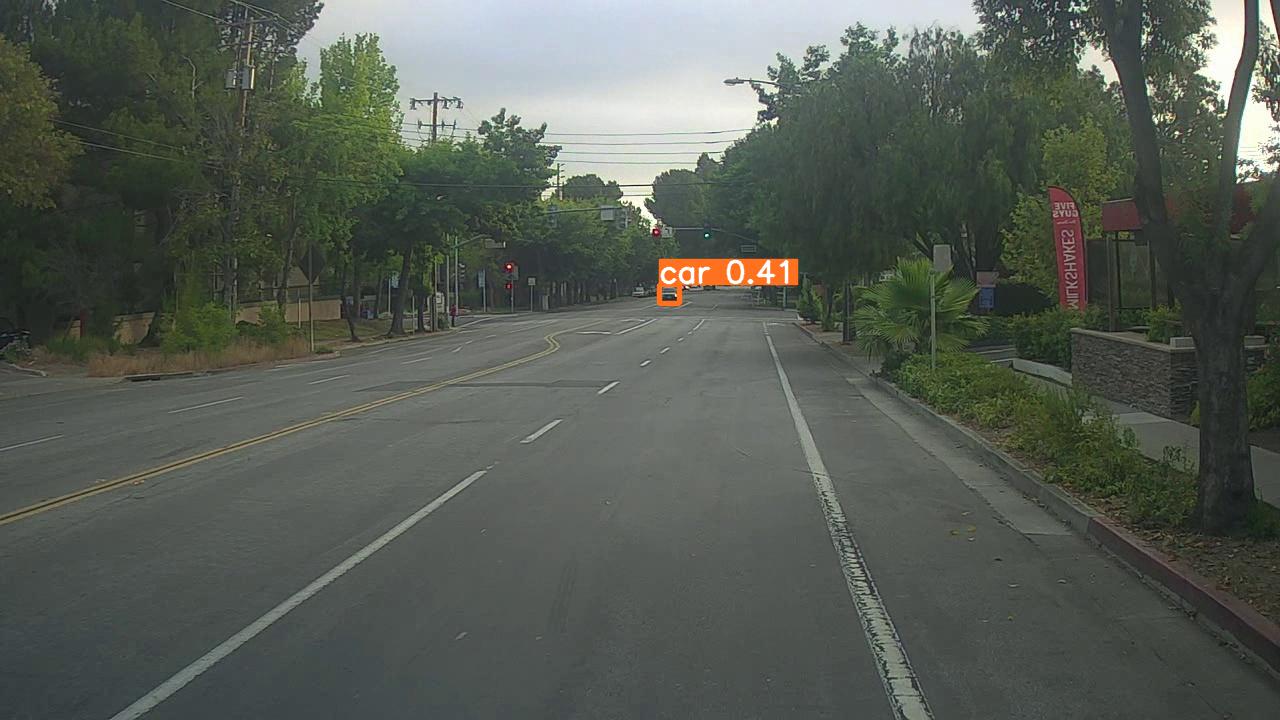}
     \put(5,5){\textcolor{white}{Frame 60, (m)}}
     \end{overpic}
     \begin{overpic}[width=0.45\textwidth]{img/CA/Yv8l/0060.jpg}
     \put(5,5){\textcolor{white}{Frame 60, (l)}}
     \end{overpic}
     \begin{overpic}[width=0.45\textwidth]{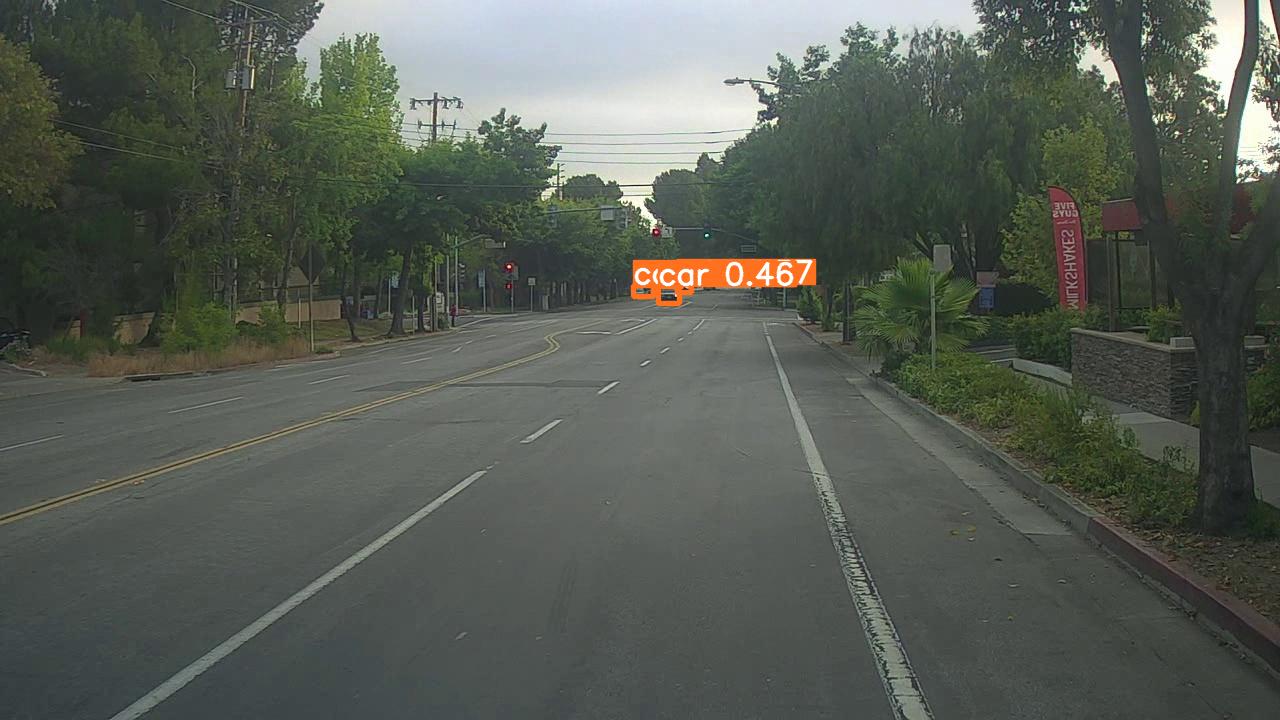}
     \put(5,5){\textcolor{white}{Frame 60, (x)}}
     \end{overpic}
     \begin{overpic}[width=0.45\textwidth]{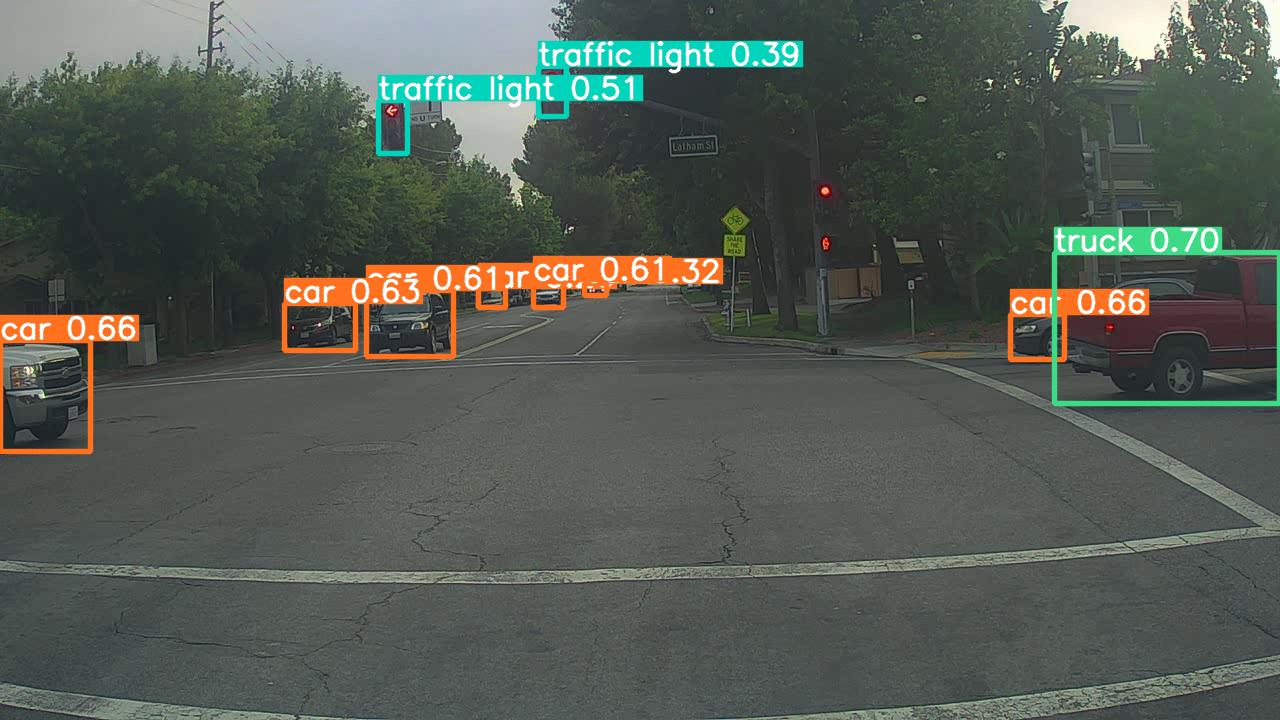}
     \put(5,5){\textcolor{white}{Frame 72, (n)}}
     \end{overpic}
     \begin{overpic}[width=0.45\textwidth]{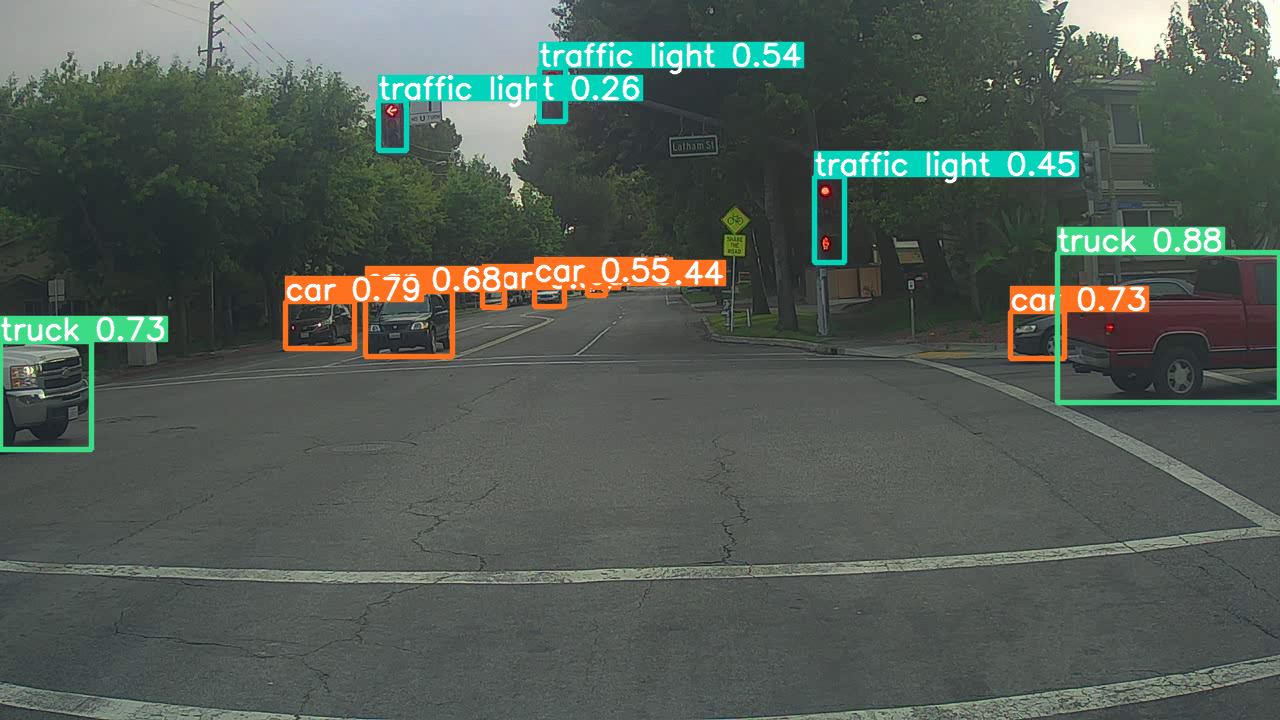}
     \put(5,5){\textcolor{white}{Frame 72, (s)}}
     \end{overpic}
     \begin{overpic}[width=0.45\textwidth]{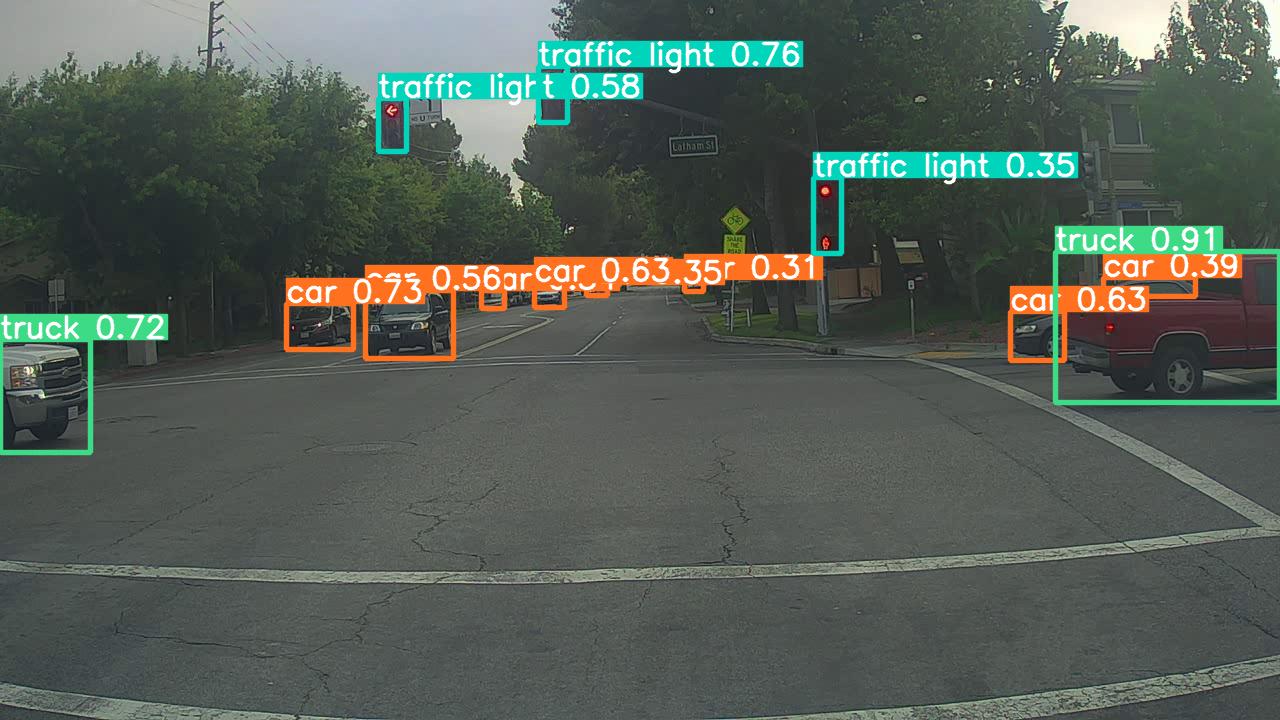}
     \put(5,5){\textcolor{white}{Frame 72, (m)}}
     \end{overpic}
     \begin{overpic}[width=0.45\textwidth]{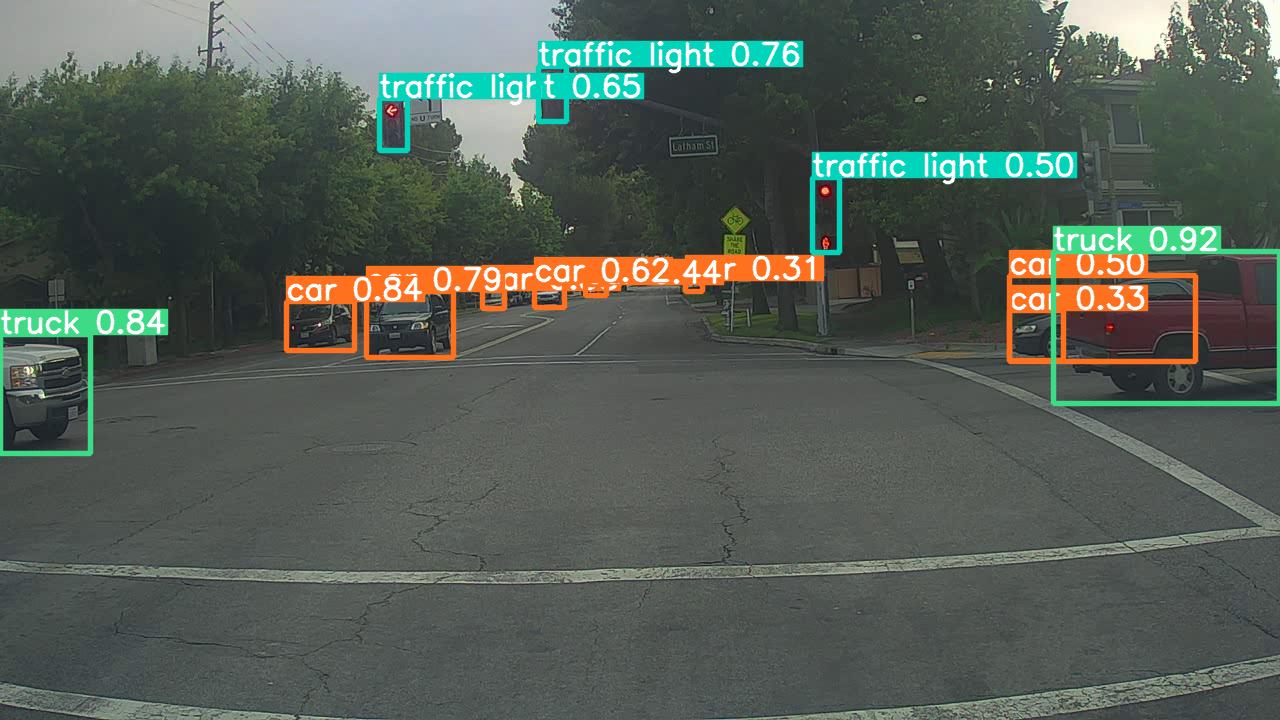}
     \put(5,5){\textcolor{white}{Frame 72, (l)}}
     \end{overpic}
     \begin{overpic}[width=0.45\textwidth]{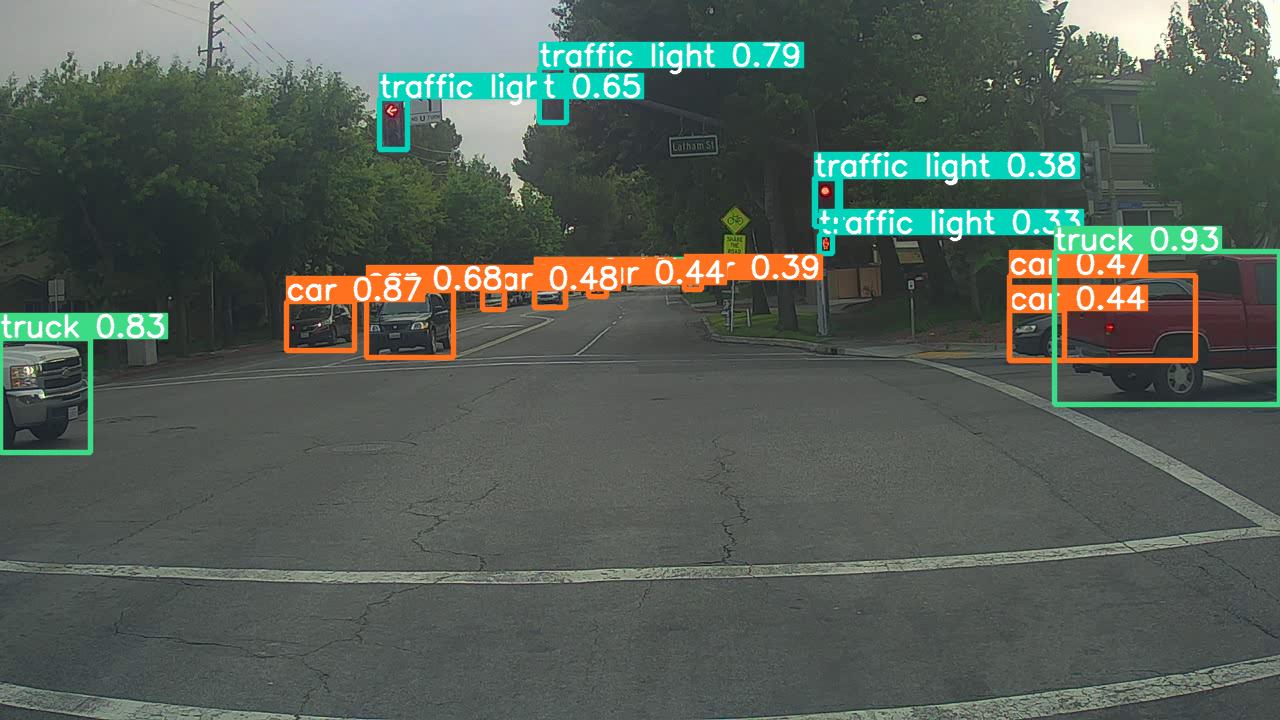}
     \put(5,5){\textcolor{white}{Frame 72, (m)}}
     \end{overpic}
        \caption{YOLOv8 flavors applied to two example scenes in CA.}
        \label{fig:appUS_v8}
\end{figure}

\section{CA - RT-DETR flavors} \label{sec:append3}
\begin{figure}[ht!]
     \centering
     \begin{overpic}[width=0.49\textwidth]{img/CA/Dl/0060.jpg}
     \put(5,5){\textcolor{white}{Frame 60, (l)}}
     \end{overpic}
     \begin{overpic}[width=0.49\textwidth]{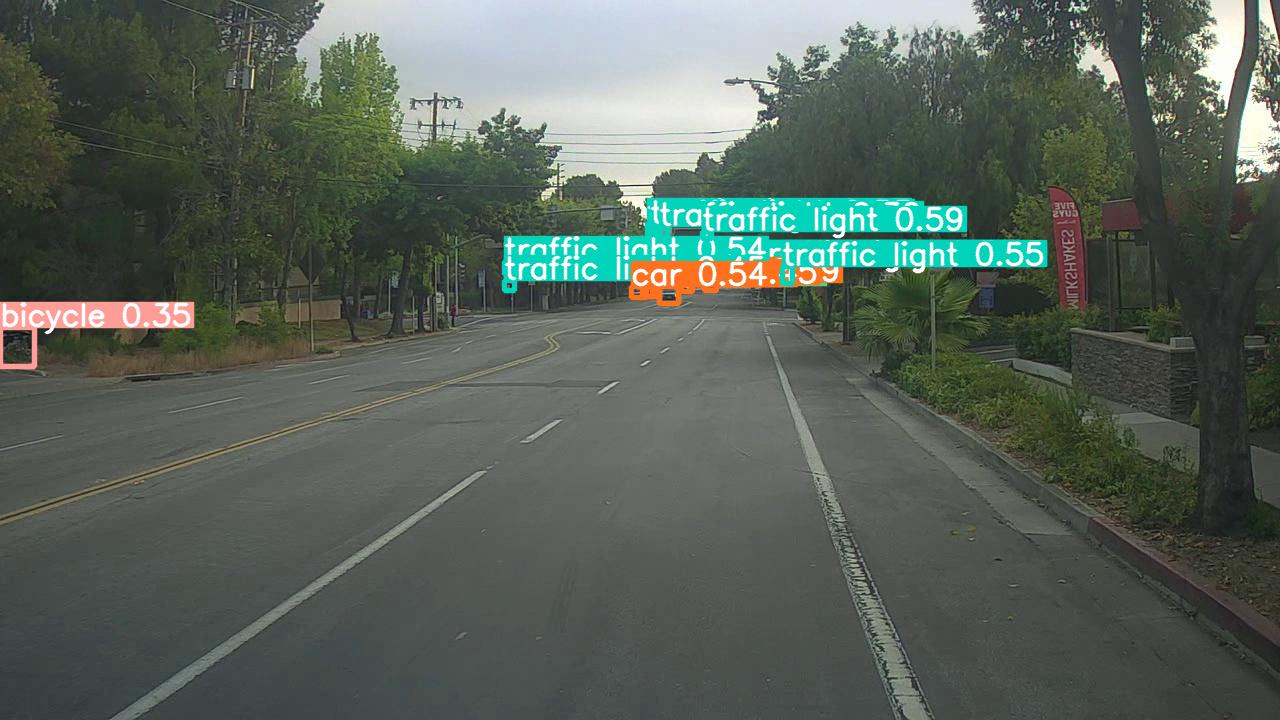}
     \put(5,5){\textcolor{white}{Frame 60, (x)}}
     \end{overpic}
     \begin{overpic}[width=0.49\textwidth]{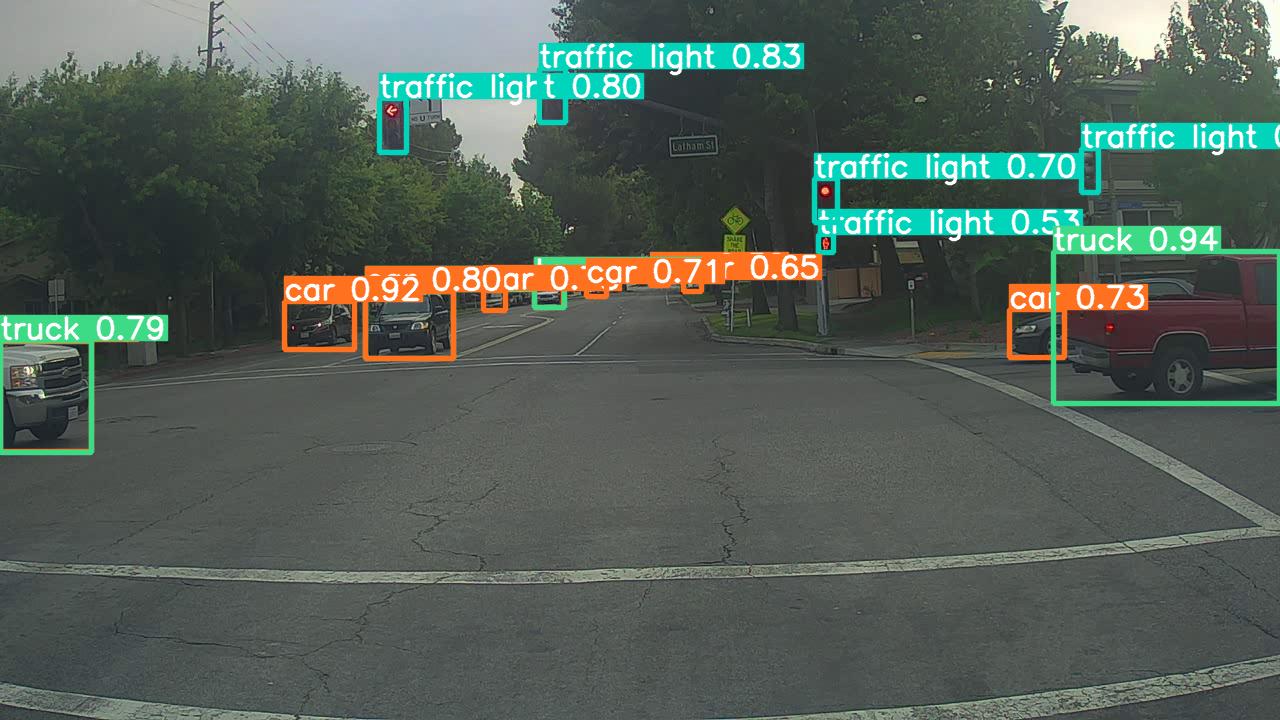}
     \put(5,5){\textcolor{white}{Frame 72, (l)}}
     \end{overpic}
     \begin{overpic}[width=0.49\textwidth]{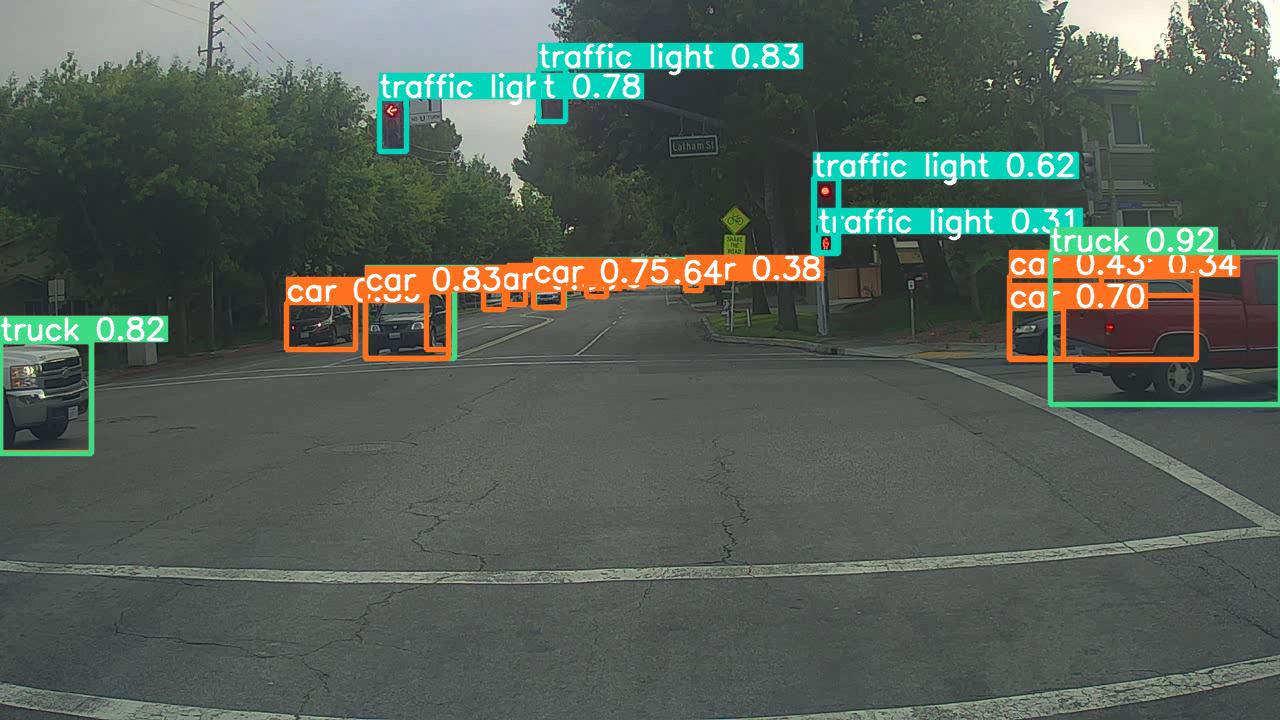}
     \put(5,5){\textcolor{white}{Frame 72, (x)}}
     \end{overpic}
        \caption{RT-DETR flavors applied to two example scenes in CA.}
        \label{fig:appUS_DETR}
\end{figure}

\newpage
\section{AUT - YOLOv5 flavors} \label{sec:append4}
\begin{figure}[ht!]
     \centering
     \begin{overpic}[width=0.45\textwidth]{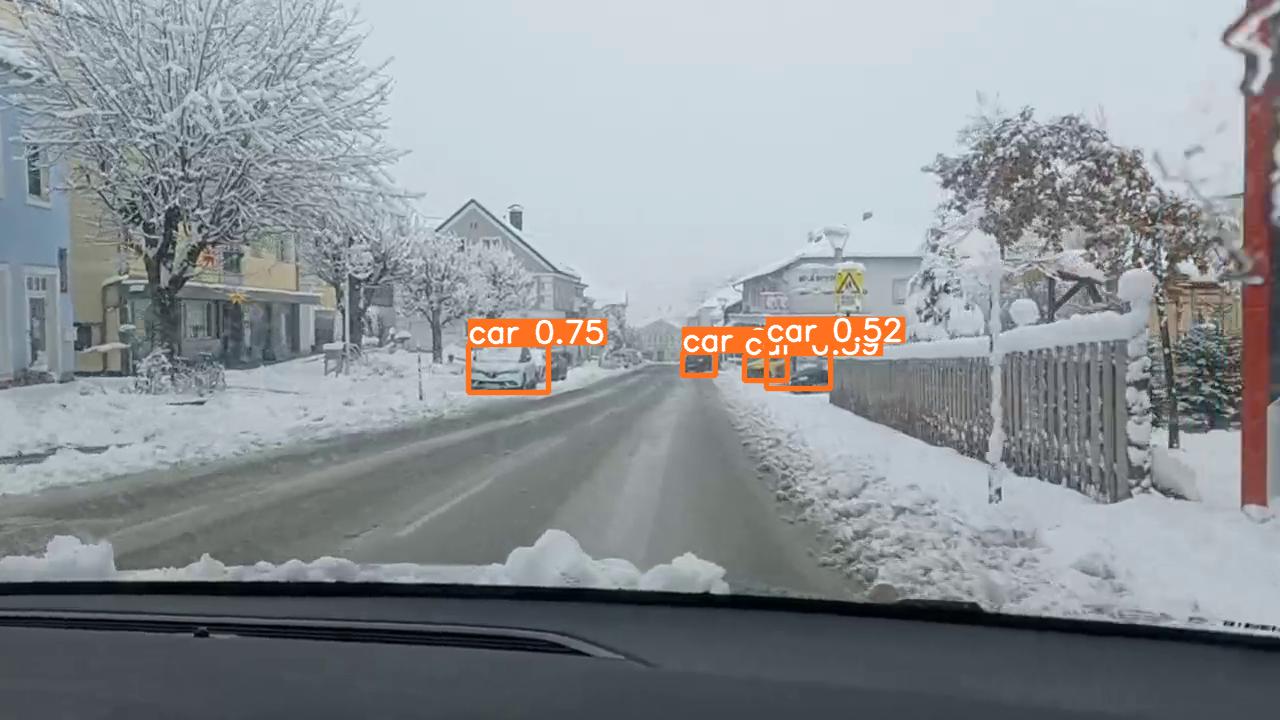}
     \put(5,5){\textcolor{white}{Frame 11, (n)}}
     \end{overpic}
     \begin{overpic}[width=0.45\textwidth]{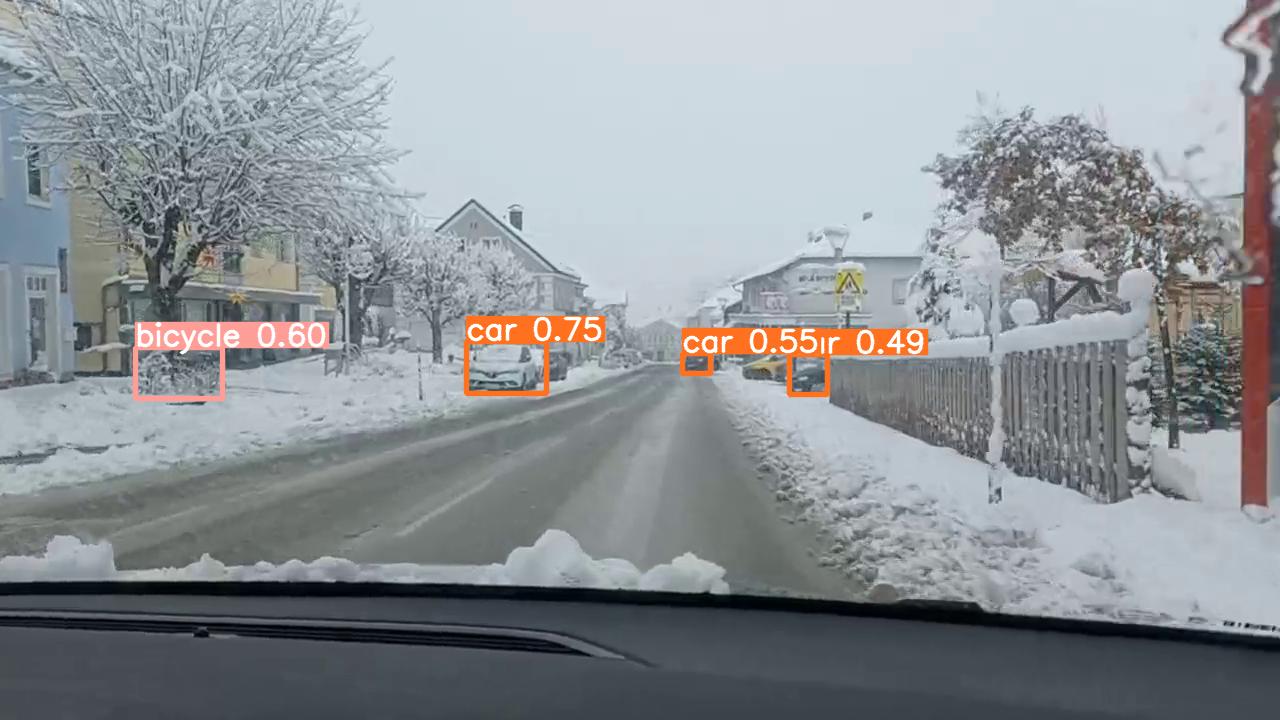}
     \put(5,5){\textcolor{white}{Frame 11, (s)}}
     \end{overpic}
     \begin{overpic}[width=0.45\textwidth]{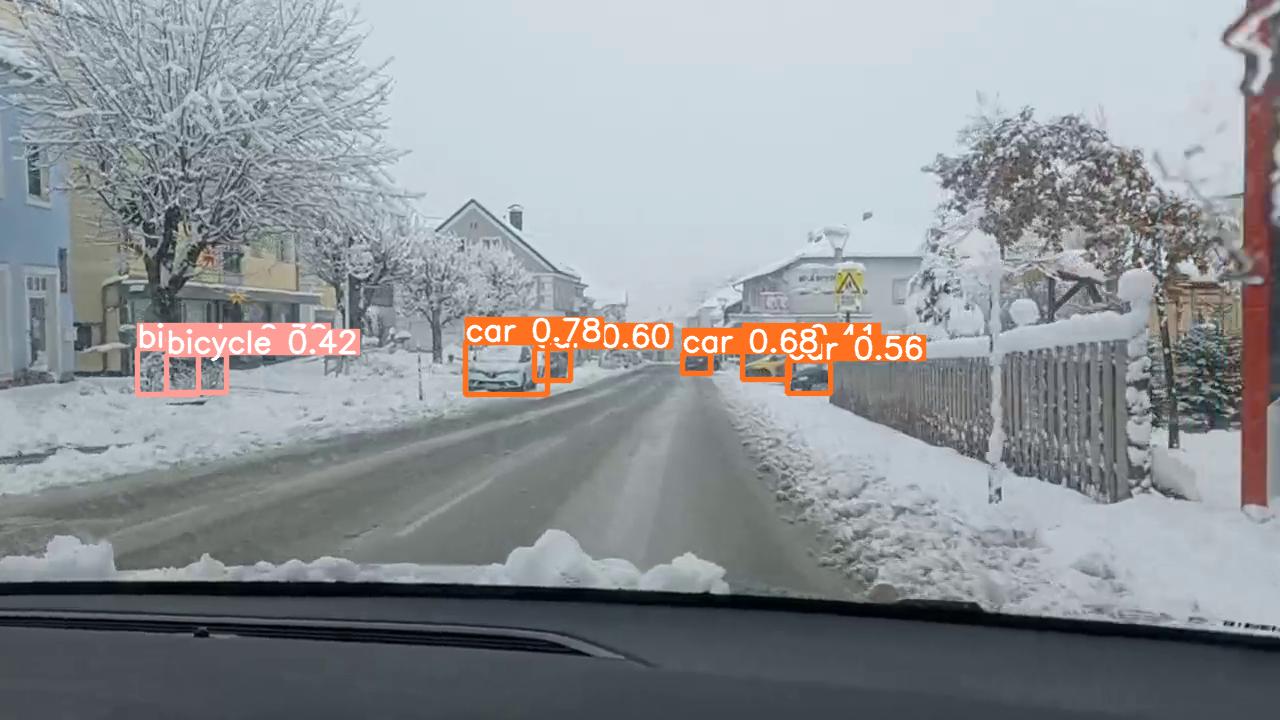}
     \put(5,5){\textcolor{white}{Frame 11, (m)}}
     \end{overpic}
     \begin{overpic}[width=0.45\textwidth]{img/AT/Yv5l/0011.png}
     \put(5,5){\textcolor{white}{Frame 11, (l)}}
     \end{overpic}
     \begin{overpic}[width=0.45\textwidth]{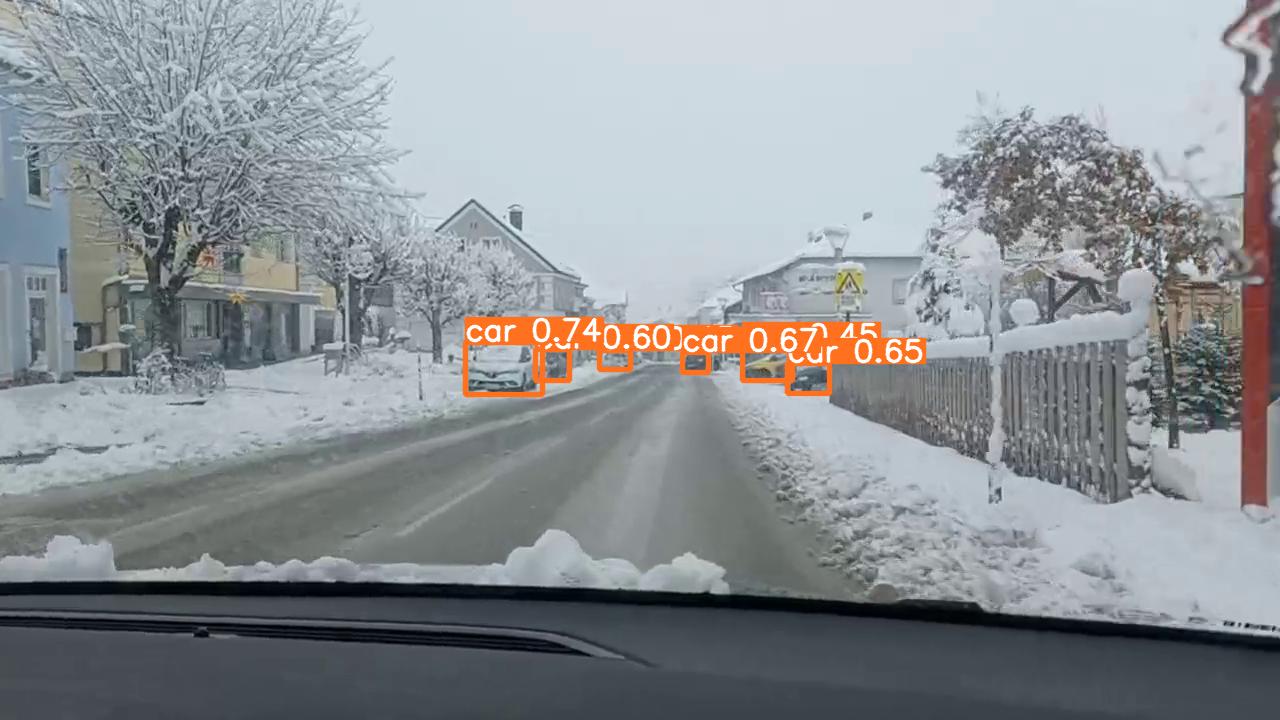}
     \put(5,5){\textcolor{white}{Frame 11, (x)}}
     \end{overpic}
     \begin{overpic}[width=0.45\textwidth]{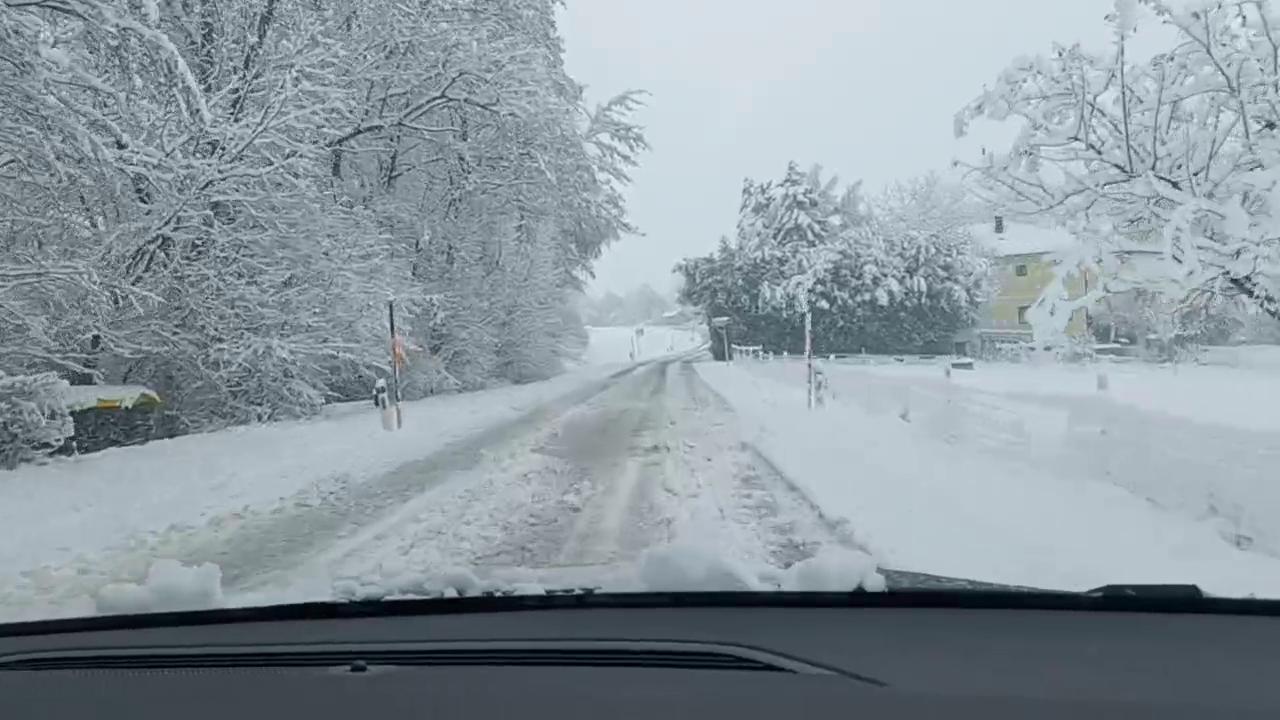}
     \put(5,5){\textcolor{white}{Frame 50, (n)}}
     \end{overpic}
     \begin{overpic}[width=0.45\textwidth]{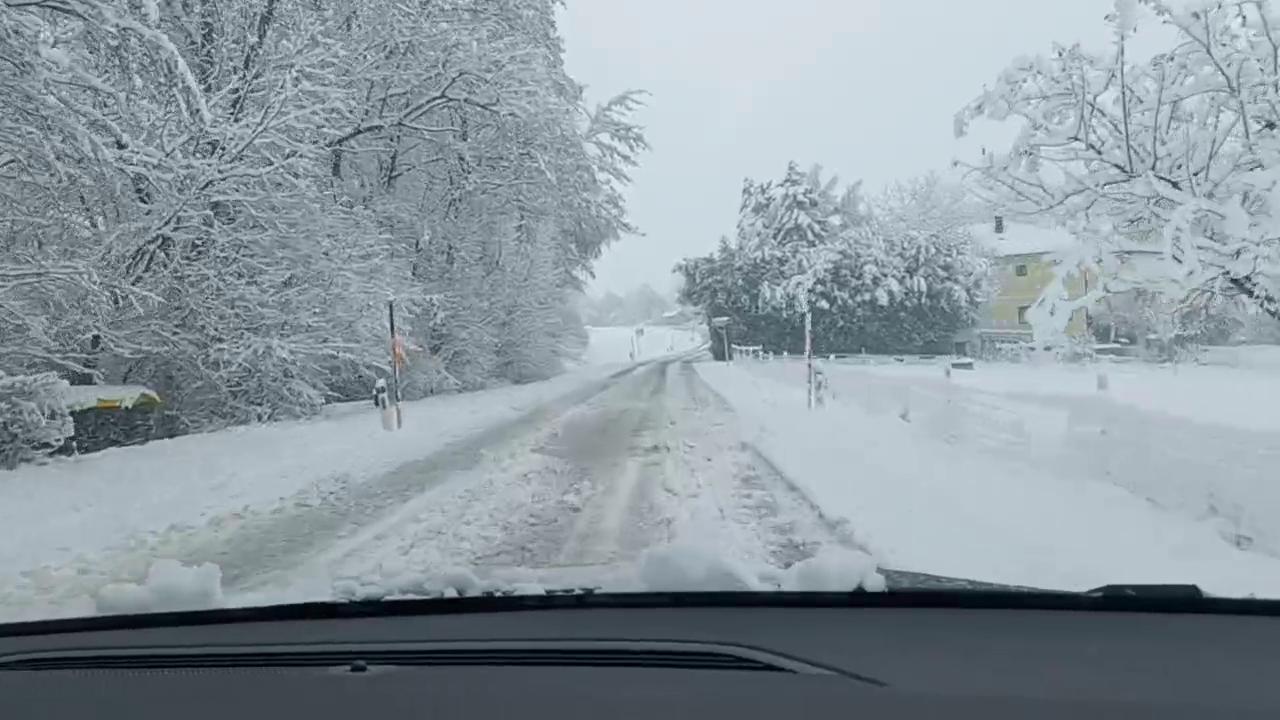}
     \put(5,5){\textcolor{white}{Frame 50, (s)}}
     \end{overpic}
     \begin{overpic}[width=0.45\textwidth]{img/AT/Yv5n/0050.png}
     \put(5,5){\textcolor{white}{Frame 50, (m)}}
     \end{overpic}
     \begin{overpic}[width=0.45\textwidth]{img/AT/Yv5l/0050.png}
     \put(5,5){\textcolor{white}{Frame 50, (l)}}
     \end{overpic}
     \begin{overpic}[width=0.45\textwidth]{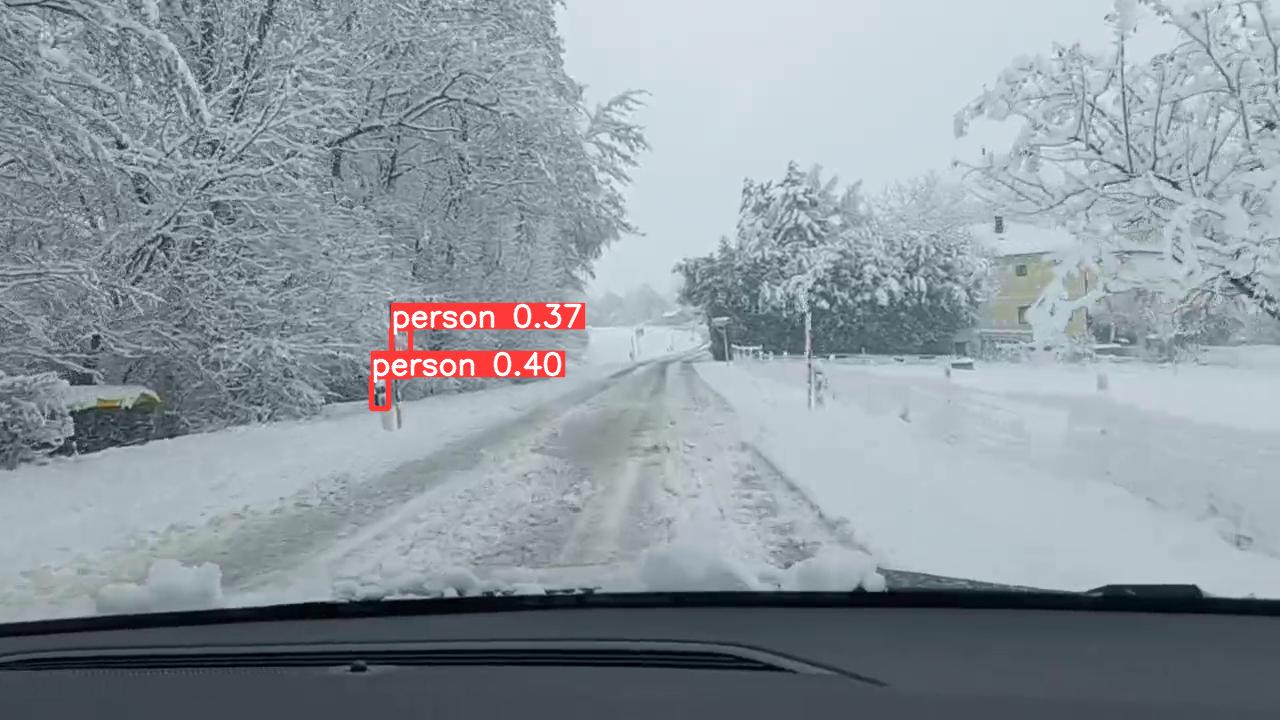}
     \put(5,5){\textcolor{white}{Frame 50, (x)}}
     \end{overpic}
        \caption{YOLOv5 flavors applied to two example scenes in AUT.}
        \label{fig:appAT_v5}
\end{figure}

\newpage
\section{AUT - YOLOv8 flavors} \label{sec:append5}
\begin{figure}[ht!]
     \centering
     \begin{overpic}[width=0.45\textwidth]{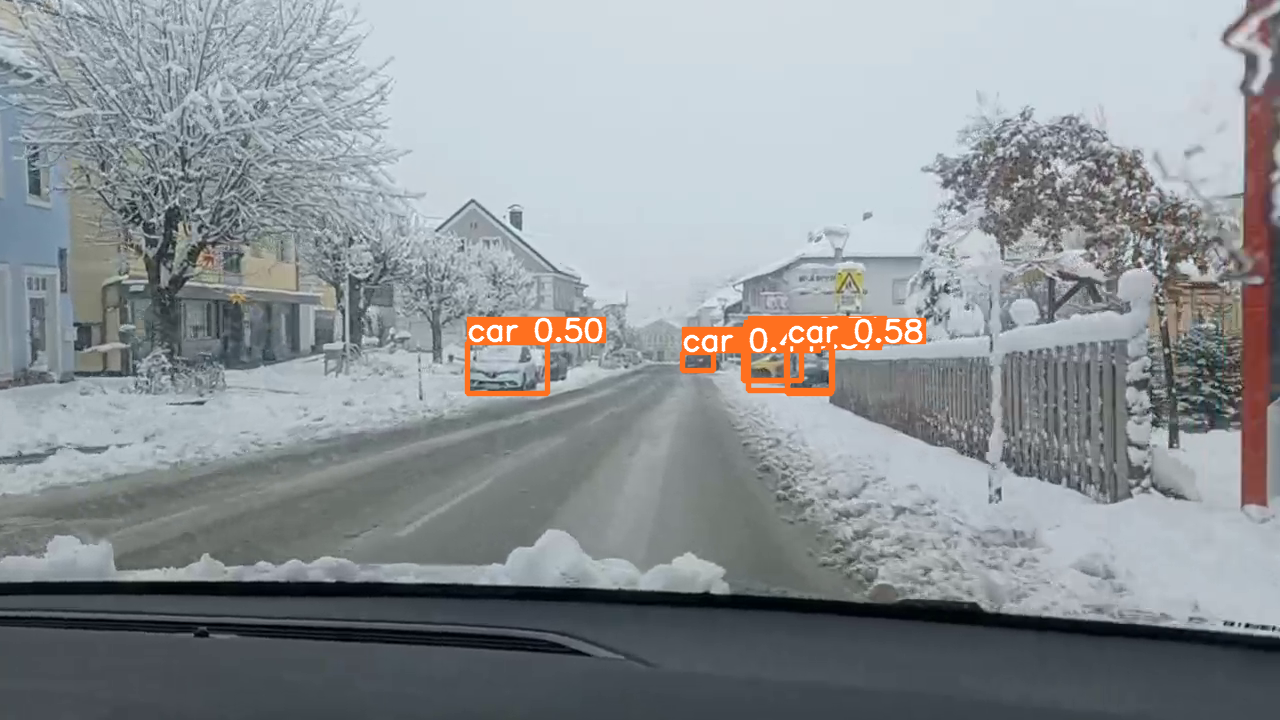}
     \put(5,5){\textcolor{white}{Frame 11, (n)}}
     \end{overpic}
     \begin{overpic}[width=0.45\textwidth]{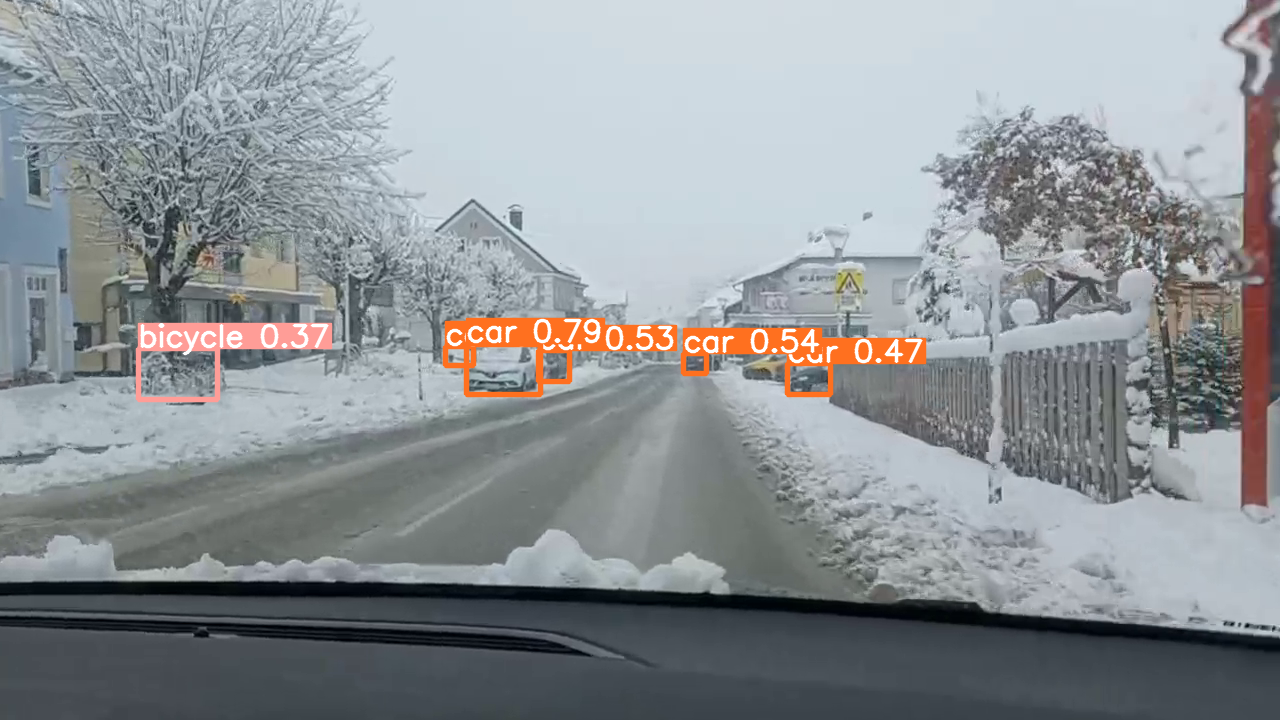}
     \put(5,5){\textcolor{white}{Frame 11, (s)}}
     \end{overpic}
     \begin{overpic}[width=0.45\textwidth]{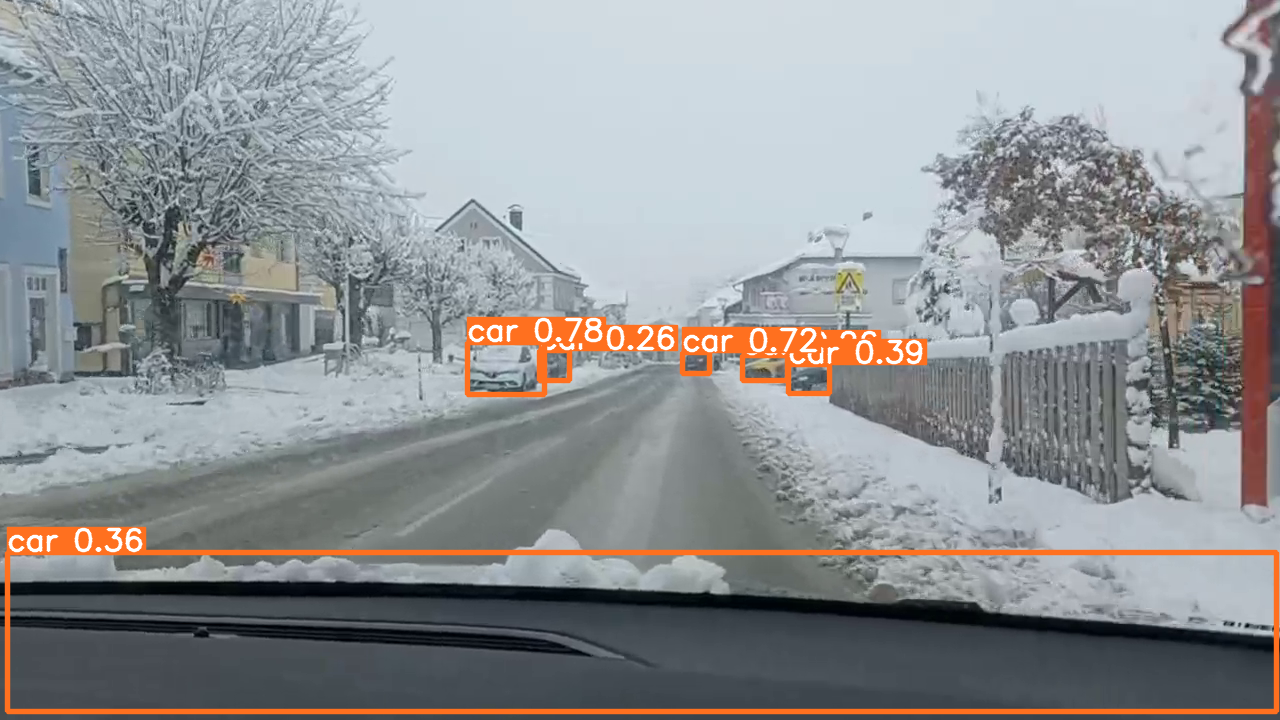}
     \put(5,5){\textcolor{white}{Frame 11, (m)}}
     \end{overpic}
     \begin{overpic}[width=0.45\textwidth]{img/AT/Yv8l/0011.png}
     \put(5,5){\textcolor{white}{Frame 11, (l)}}
     \end{overpic}
     \begin{overpic}[width=0.45\textwidth]{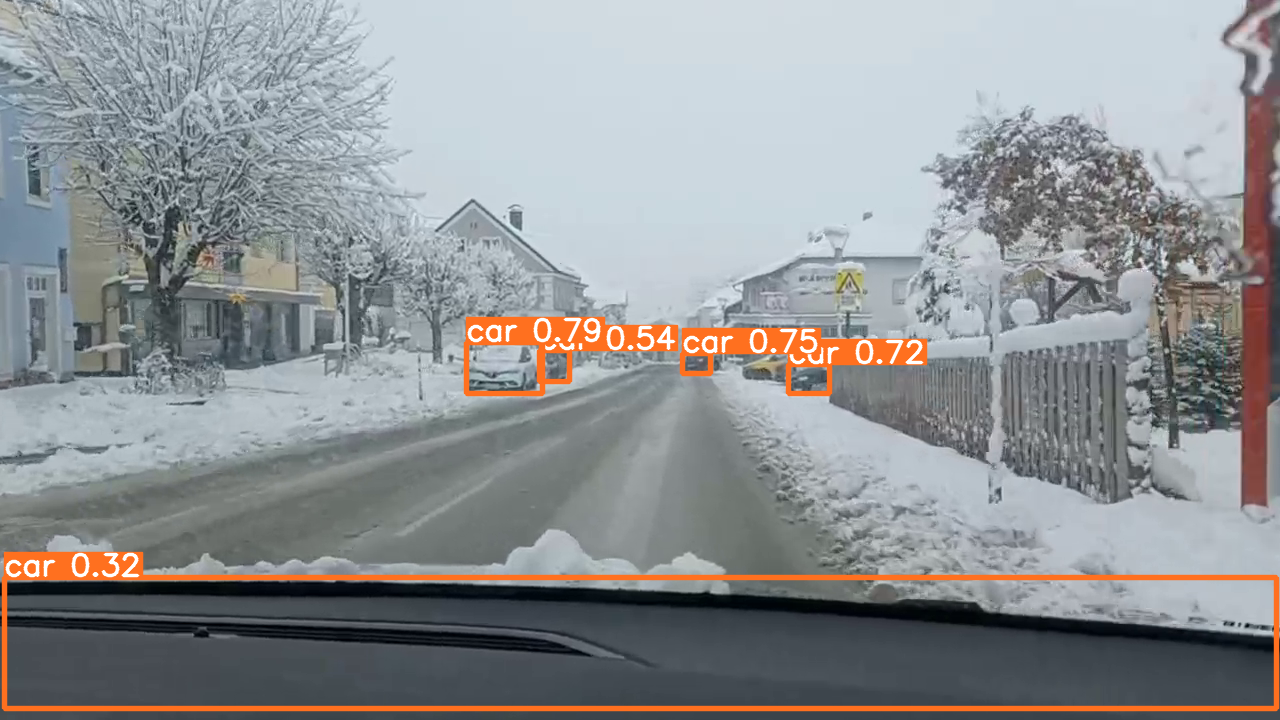}
     \put(5,5){\textcolor{white}{Frame 11, (x)}}
     \end{overpic}
     \begin{overpic}[width=0.45\textwidth]{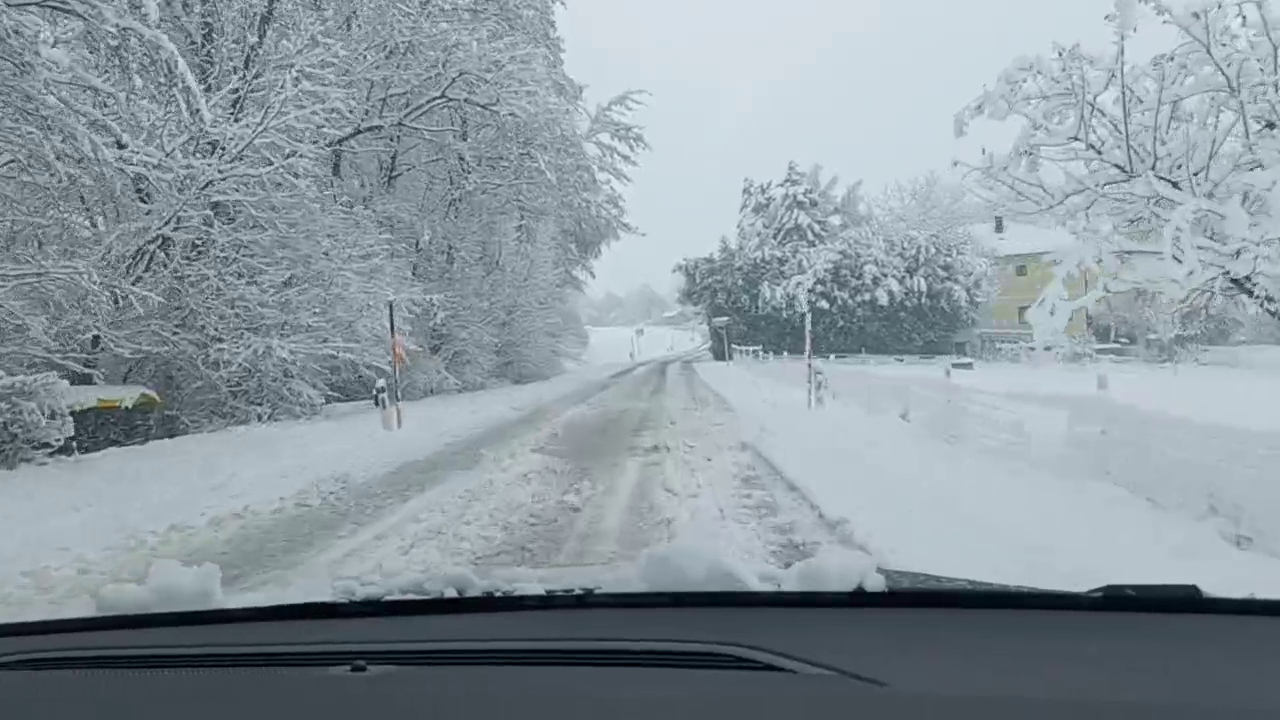}
     \put(5,5){\textcolor{white}{Frame 50, (n)}}
     \end{overpic}
     \begin{overpic}[width=0.45\textwidth]{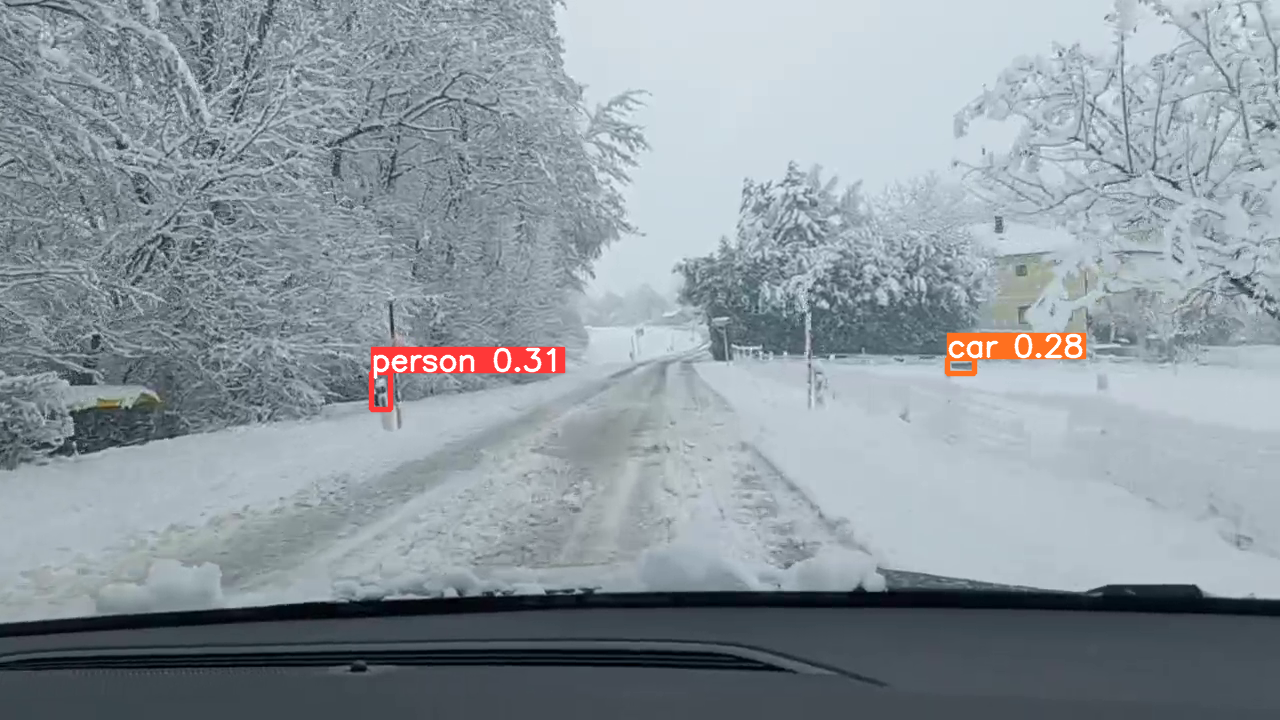}
     \put(5,5){\textcolor{white}{Frame 50, (s)}}
     \end{overpic}
     \begin{overpic}[width=0.45\textwidth]{img/AT/Yv8n/0050.png}
     \put(5,5){\textcolor{white}{Frame 50, (m)}}
     \end{overpic}
     \begin{overpic}[width=0.45\textwidth]{img/AT/Yv8l/0050.png}
     \put(5,5){\textcolor{white}{Frame 50, (l)}}
     \end{overpic}
     \begin{overpic}[width=0.45\textwidth]{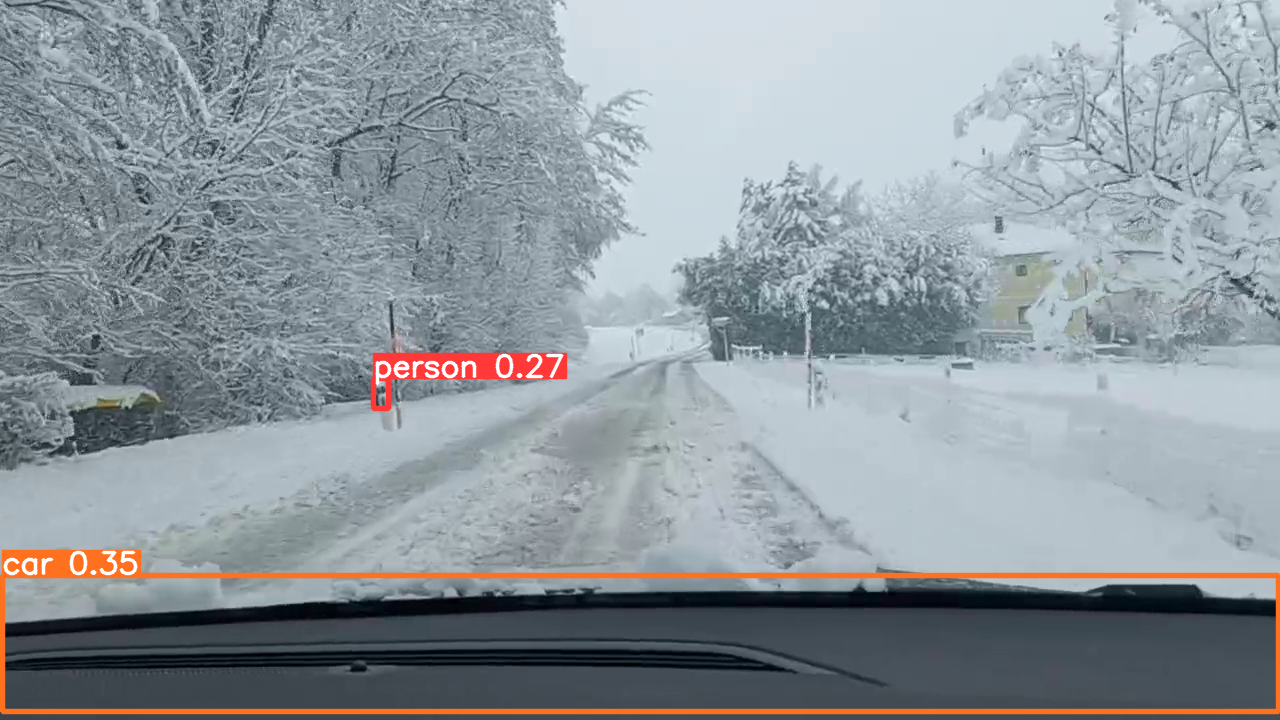}
     \put(5,5){\textcolor{white}{Frame 50, (x)}}
     \end{overpic}
        \caption{YOLOv8 flavors applied to two example scenes in AUT.}
        \label{fig:appAT_v8}
\end{figure}

\section{AUT - RT-DETR flavors} \label{sec:append6}
\begin{figure}[ht!]
     \centering
     \begin{overpic}[width=0.49\textwidth]{img/AT/Dl/0011.png}
     \put(5,5){\textcolor{white}{Frame 11, (l)}}
     \end{overpic}
     \begin{overpic}[width=0.49\textwidth]{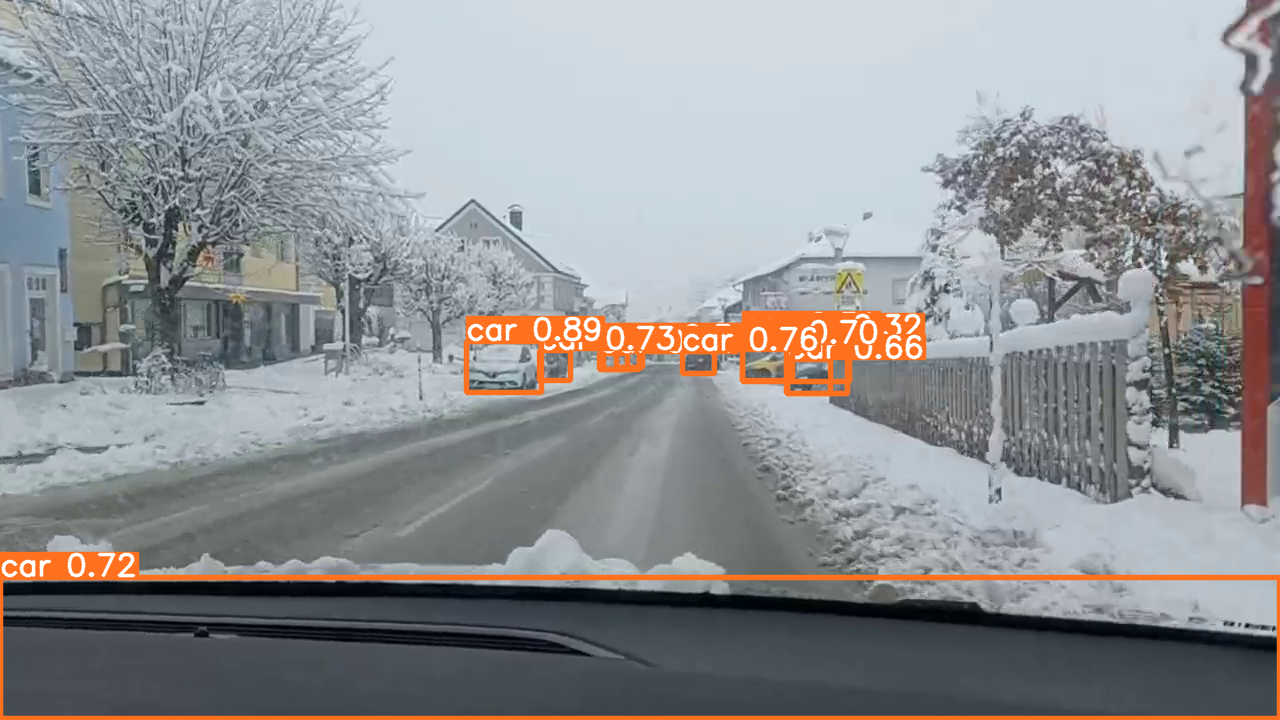}
     \put(5,5){\textcolor{white}{Frame 11, (x)}}
     \end{overpic}
     \begin{overpic}[width=0.49\textwidth]{img/AT/Dl/0050.png}
     \put(5,5){\textcolor{white}{Frame 50, (l)}}
     \end{overpic}
     \begin{overpic}[width=0.49\textwidth]{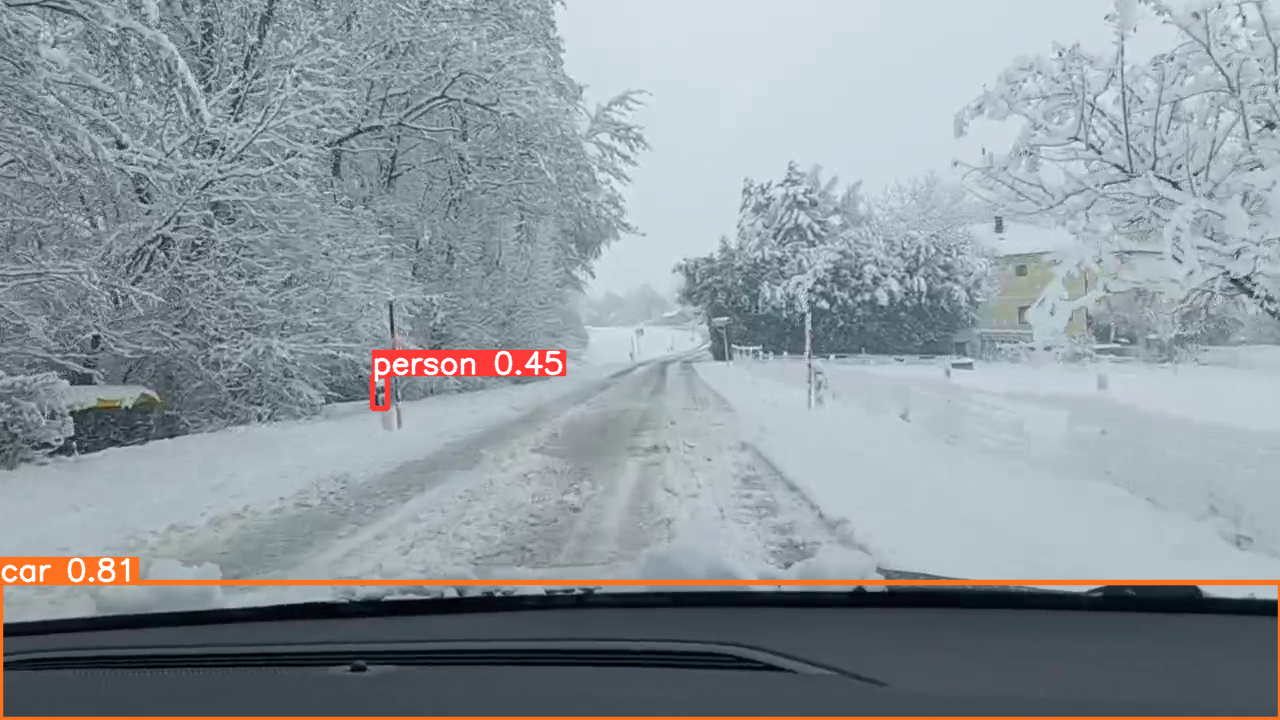}
     \put(5,5){\textcolor{white}{Frame 50, (x)}}
     \end{overpic}
     \begin{overpic}[width=0.49\textwidth]{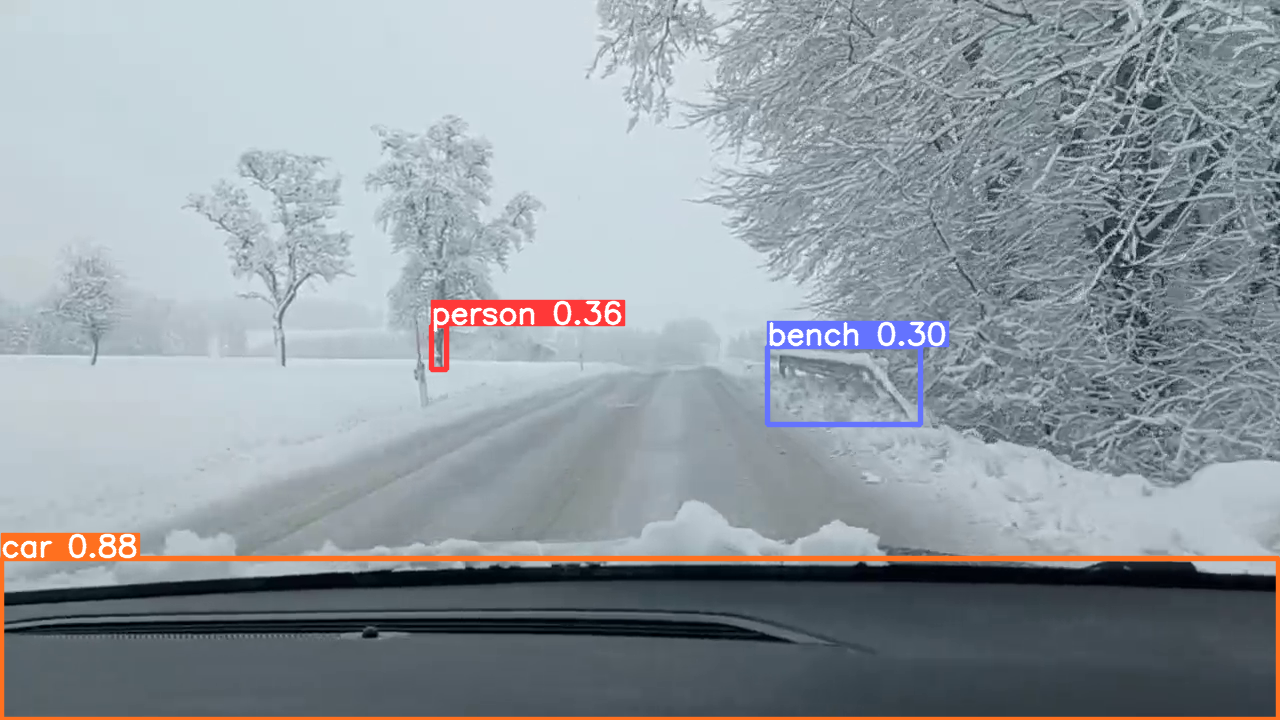}
     \put(5,5){\textcolor{white}{Frame 77, (l)}}
     \end{overpic}
     \begin{overpic}[width=0.49\textwidth]{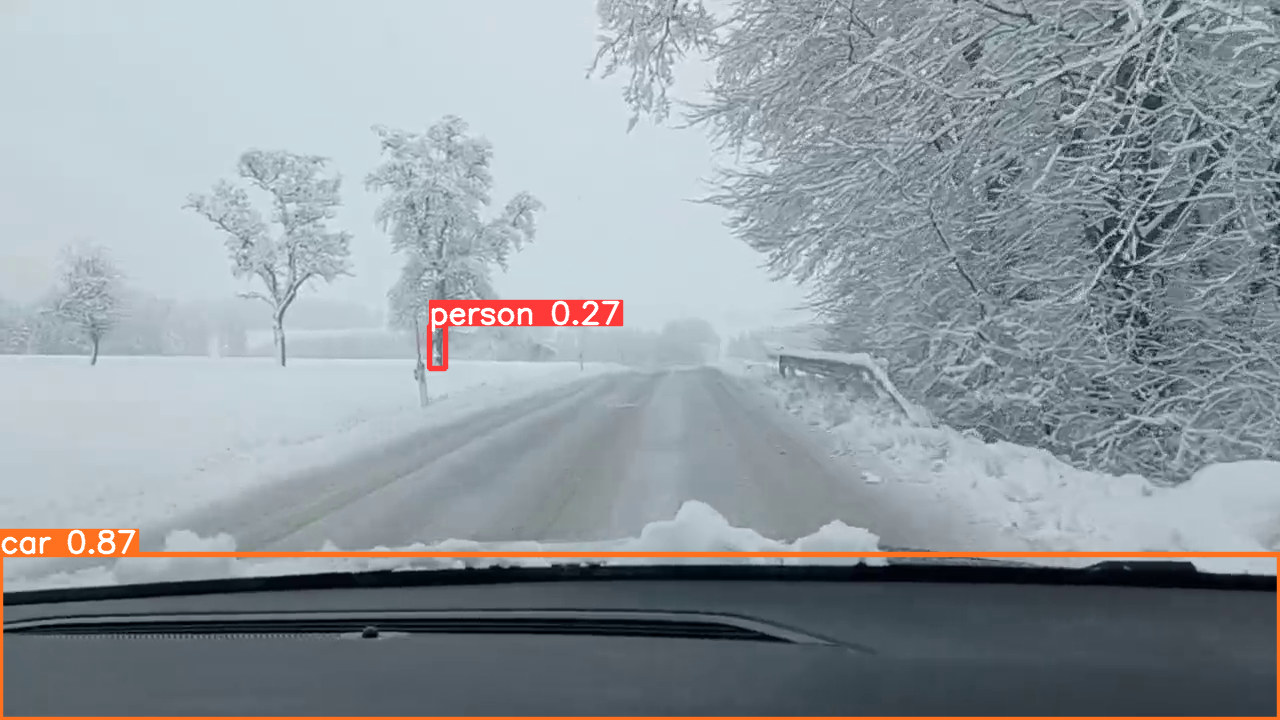}
     \put(5,5){\textcolor{white}{Frame 77, (x)}}
     \end{overpic}
        \caption{RT-DETR flavors applied to three example scenes in AUT.}
        \label{fig:appAT_DETR}
\end{figure}

\newpage
\section{AUT - Comparison YOLOv5, YOLOv8 and RT-DETR flavors} \label{sec:append7}
\begin{figure}[ht!]
     \centering
     \begin{overpic}[width=0.45\textwidth]{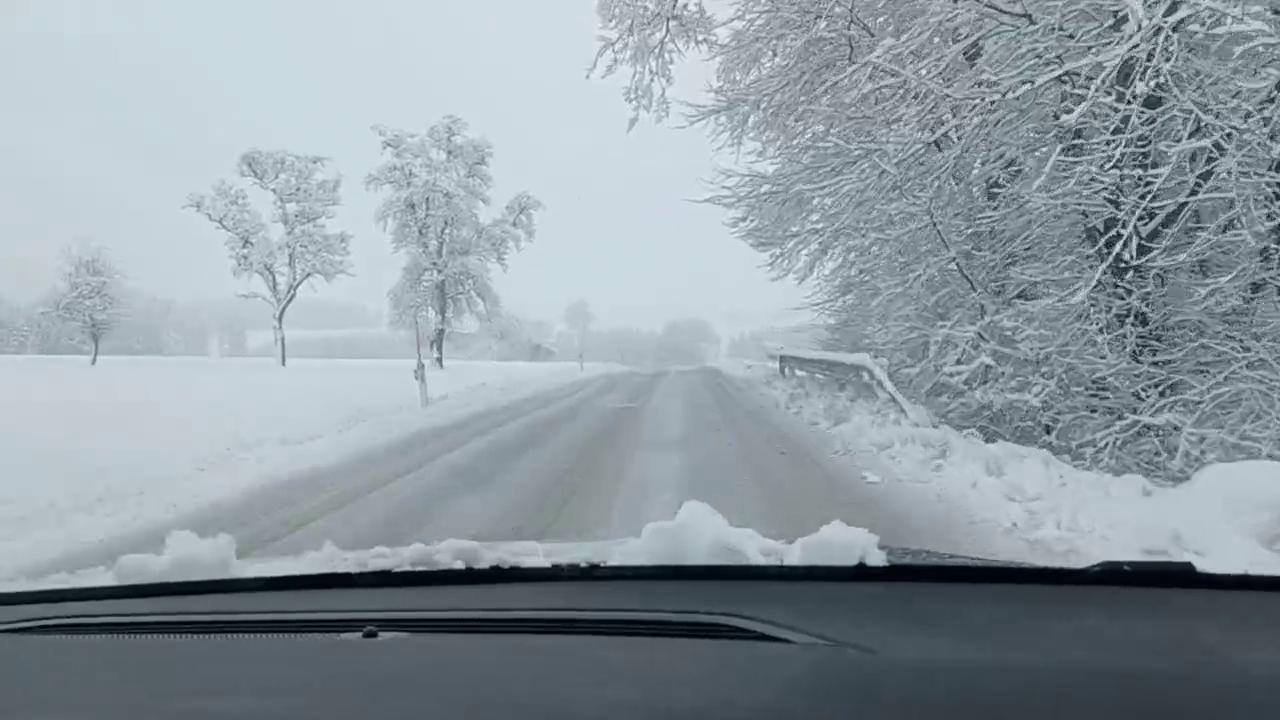}
     \put(5,5){\textcolor{white}{Frame 77, YOLOv5 (s)}}
     \end{overpic}
     \begin{overpic}[width=0.45\textwidth]{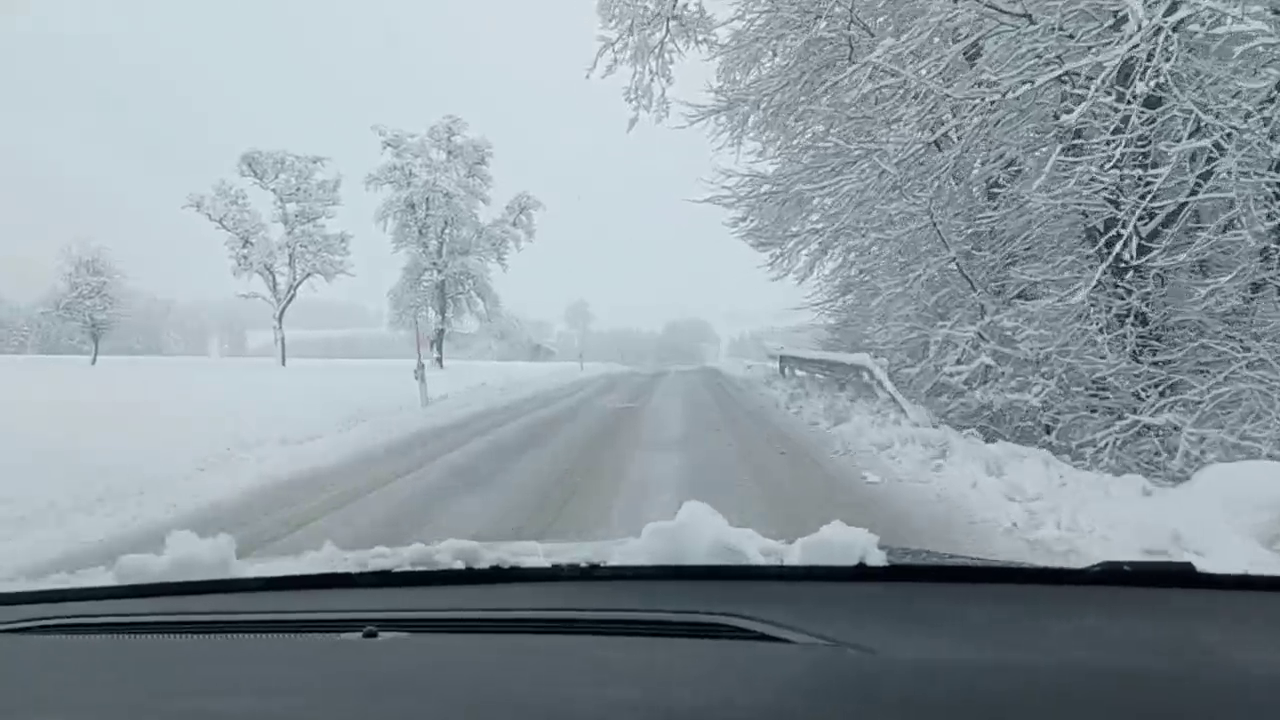}
     \put(5,5){\textcolor{white}{Frame 77, YOLOv8 (s)}}
     \end{overpic}
     \begin{overpic}[width=0.45\textwidth]{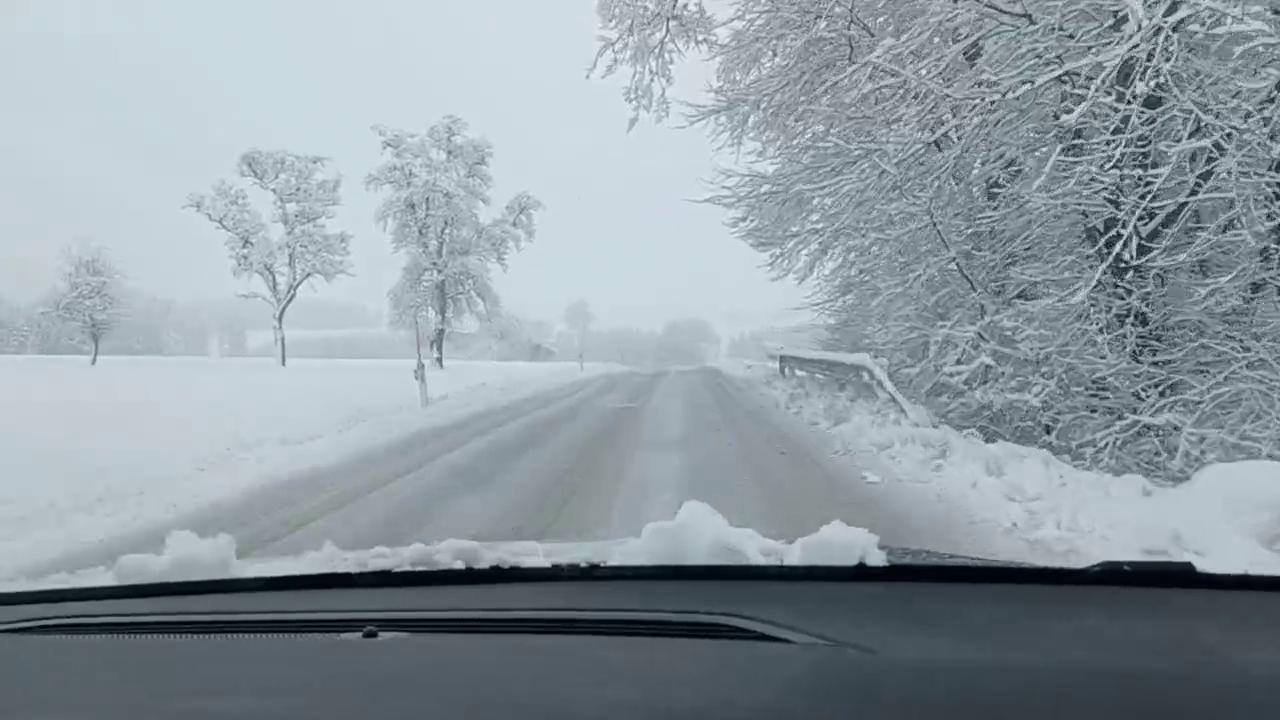}
     \put(5,5){\textcolor{white}{Frame 77, YOLOv5 (m)}}
     \end{overpic}
     \begin{overpic}[width=0.45\textwidth]{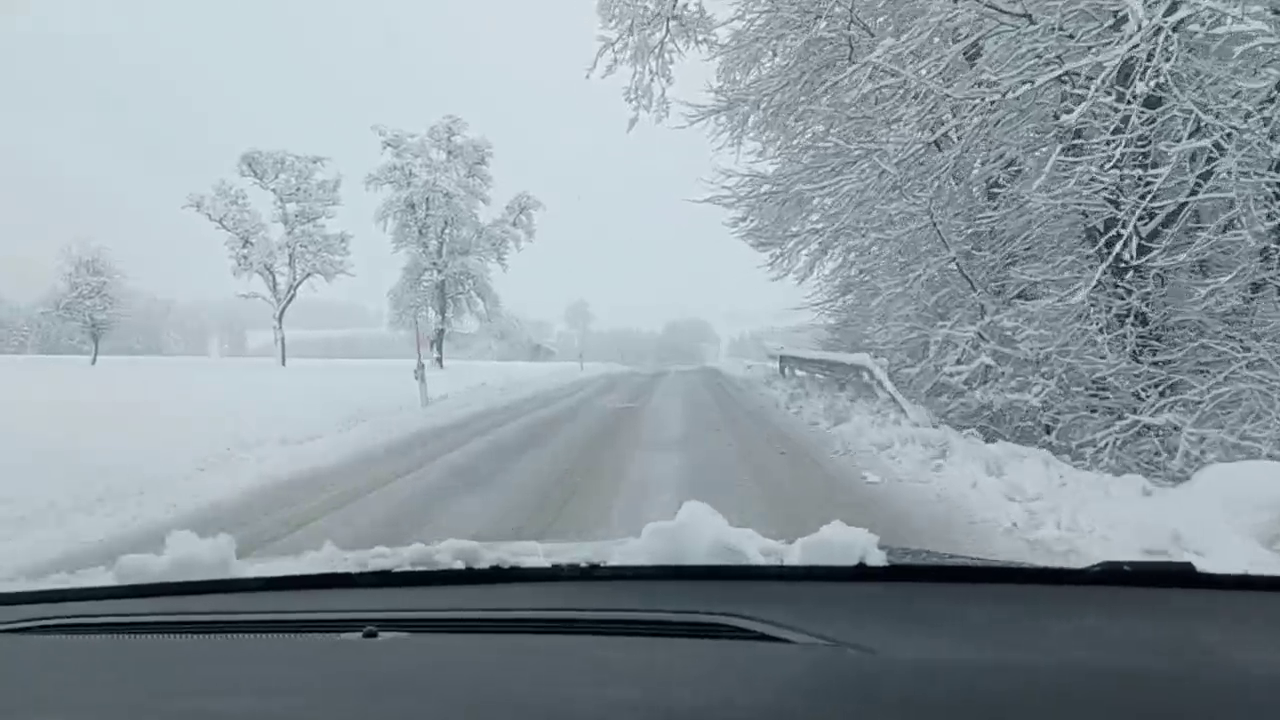}
     \put(5,5){\textcolor{white}{Frame 77, YOLOv8 (m)}}
     \end{overpic}
     \begin{overpic}[width=0.45\textwidth]{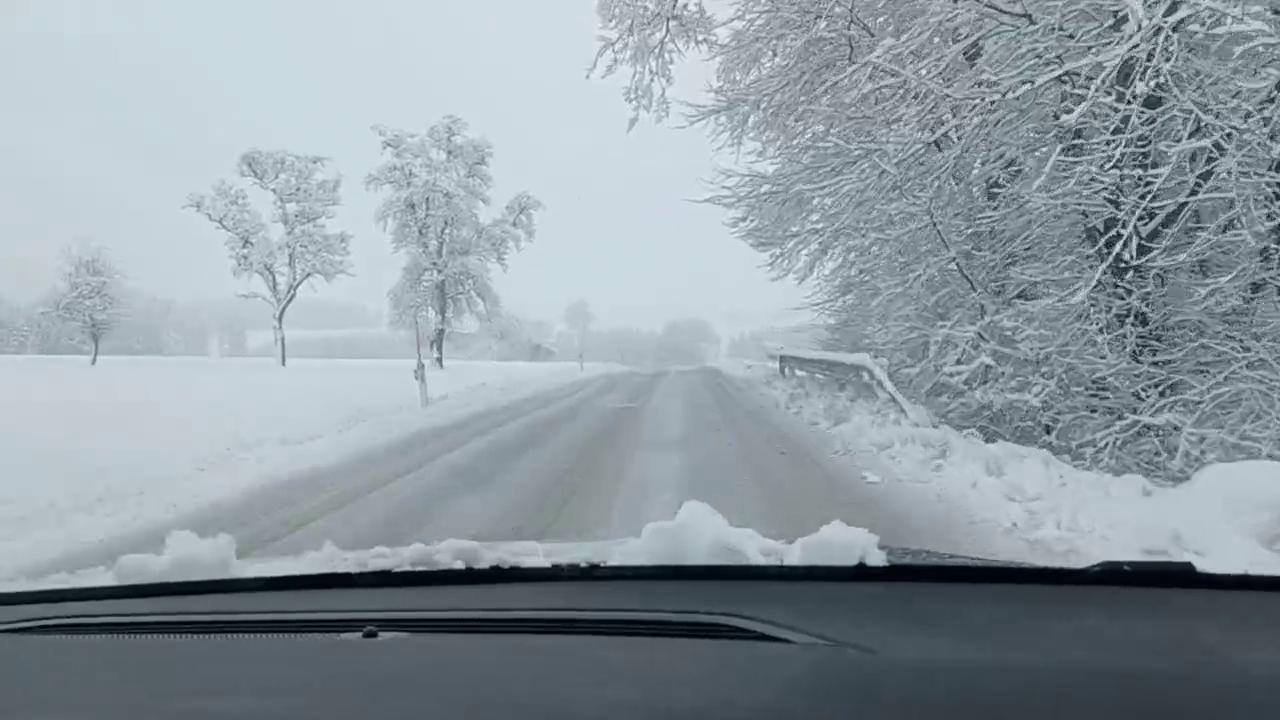}
     \put(5,5){\textcolor{white}{Frame 77, YOLOv5 (l)}}
     \end{overpic}
     \begin{overpic}[width=0.45\textwidth]{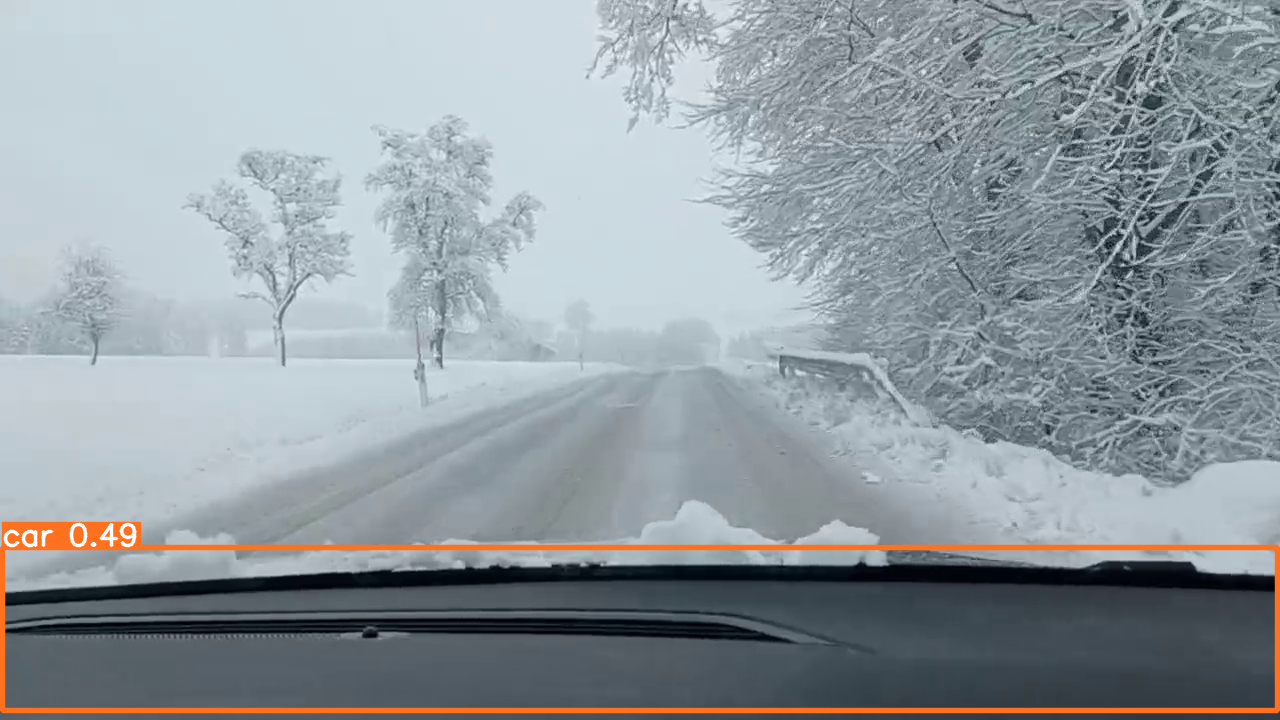}
     \put(5,5){\textcolor{white}{Frame 77, YOLOv8 (l)}}
     \end{overpic}
     \begin{overpic}[width=0.45\textwidth]{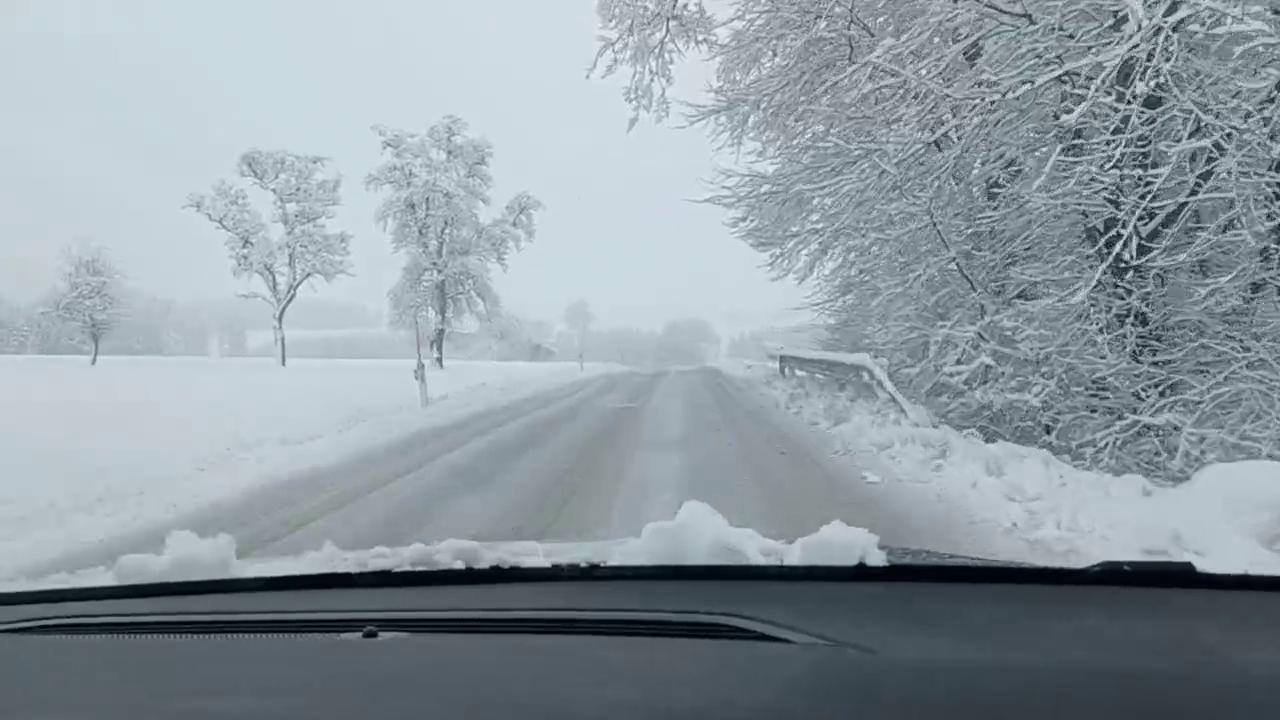}
     \put(5,5){\textcolor{white}{Frame 77, YOLOv5 (x)}}
     \end{overpic}
     \begin{overpic}[width=0.45\textwidth]{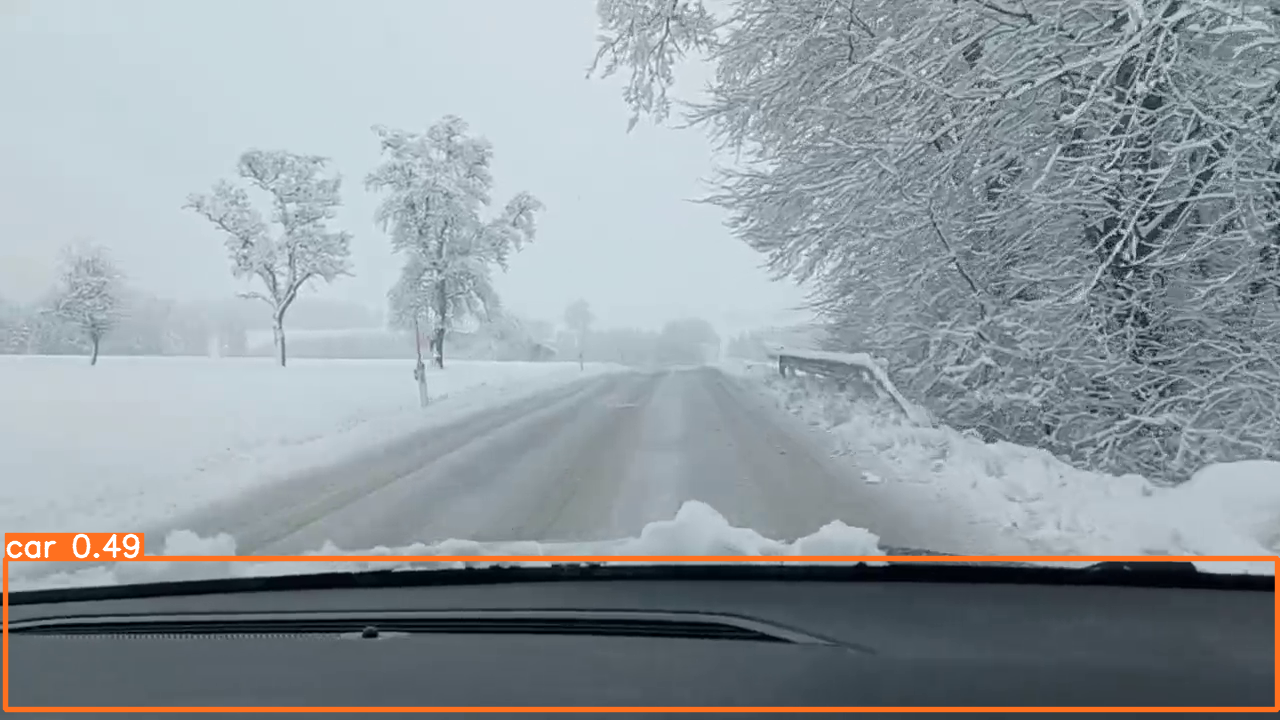}
     \put(5,5){\textcolor{white}{Frame 77, YOLOv8 (x)}}
     \end{overpic}
     \begin{overpic}[width=0.45\textwidth]{img/AT/Dl/0077.png}
     \put(5,5){\textcolor{white}{Frame 77, RT-DETR (l)}}
     \end{overpic}
     \begin{overpic}[width=0.45\textwidth]{img/AT/Dx/0077.png}
     \put(5,5){\textcolor{white}{Frame 77, RT-DETR (x)}}
     \end{overpic}
        \caption{YOLOv5 (left), YOLOv8 (right), RT-DETR (last row) flavors applied to one example scenes in AUT.}
        \label{fig:appAT_v58}
\end{figure}

\end{document}